%% file: main.tex
\documentclass[10pt,twocolumn,letterpaper]{article}

\usepackage{cvpr}

\definecolor{cvprblue}{rgb}{0.21,0.49,0.74}
\usepackage[pagebackref,breaklinks,colorlinks,allcolors=cvprblue]{hyperref}
\usepackage{svg}
\usepackage{graphicx}
\usepackage{import}
\usepackage{multicol}
\usepackage{multirow}
\usepackage{layouts}
\usepackage[export]{adjustbox}
\usepackage{pgfplots}
\usepackage[accsupp]{axessibility}

\def\paperID{17168}
\def\confName{CVPR}
\def\confYear{2025}

\title{Segment This Thing: Foveated Tokenization for Efficient Point-Prompted Segmentation}

\author{Tanner Schmidt\\
Meta Reality Labs\\
{\tt\small tanner.schmidt@meta.com}
\and
Richard Newcombe\\
Meta Reality Labs\\
{\tt\small newcombe@meta.com}
}

\begin{document}
\maketitle
\input{sec_0_abstract}    
\input{sec_1_intro}
\input{sec_2_related}
\input{sec_3_model}
\input{sec_4_experiments}
\input{sec_5_conclusion}
{
    \small
    \bibliographystyle{ieeenat_fullname}
    \bibliography{main}
}

\input{sec_X_suppl}

\typeout{get arXiv to do 4 passes: Label(s) may have changed. Rerun}
\end{document}

%% file: sec_0_abstract.tex
\begin{abstract}
This paper presents Segment This Thing (STT), a new efficient image segmentation model designed to produce a single segment given a single point prompt. Instead of following prior work and increasing efficiency by decreasing model size, we gain efficiency by \textit{foveating} input images.
Given an image and a point prompt, we extract a crop centered on the prompt and apply a novel variable-resolution patch tokenization in which patches are downsampled at a rate that increases with increased distance from the prompt. 
This approach yields far fewer image tokens than uniform patch tokenization. 
As a result we can drastically reduce the computational cost of segmentation without reducing model size. 
Furthermore, the foveation focuses the model on the region of interest, a potentially useful inductive bias. 
We show that our Segment This Thing model is more efficient than prior work while remaining competitive on segmentation benchmarks. 
It can easily run at interactive frame rates on consumer hardware and is thus a promising tool for augmented reality or robotics applications.
\end{abstract}

%% file: sec_1_intro.tex
\section{Introduction}
\label{sec:intro}

Since its introduction in 2023, the Segment Anything Model (SAM) \cite{Kirillov:etal:CVPR23} has brought new focus to the image segmentation task and enabled numerous downstream applications with its flexible design and impressive performance.

However, SAM's capabilities come at a cost: the model leverages a heavyweight image encoder (ViT-H in the standard model).

In this paper we focus on segmentation for streaming use cases, for example robotics or augmented/virtual reality applications.
With many frames arriving every second, it is likely that only one or a few entities need to be segmented in each image.
Latency is critical, as frames may be separated in time by only tens of milliseconds.
Depending on the relative locations of and communication channels between sensors and compute, bandwidth may also be a concern.

\begin{figure}
\centering
\begin{tabular}{c c}
    Patch Tokenization & Foveated Tokenization \\
    \includegraphics[width=0.47\columnwidth]{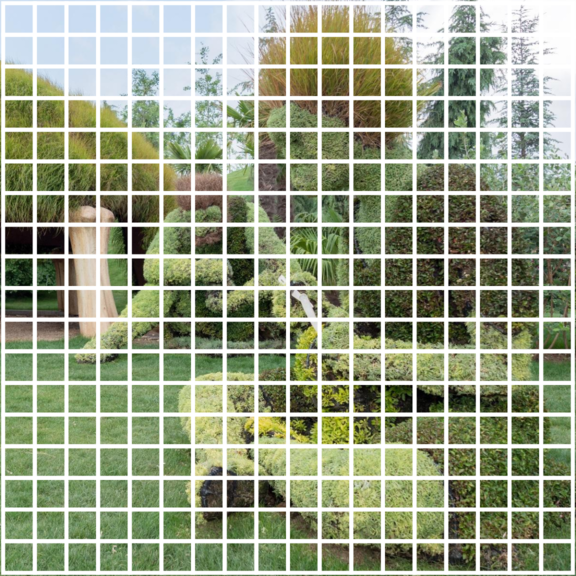} &
    \includegraphics[width=0.47\columnwidth]{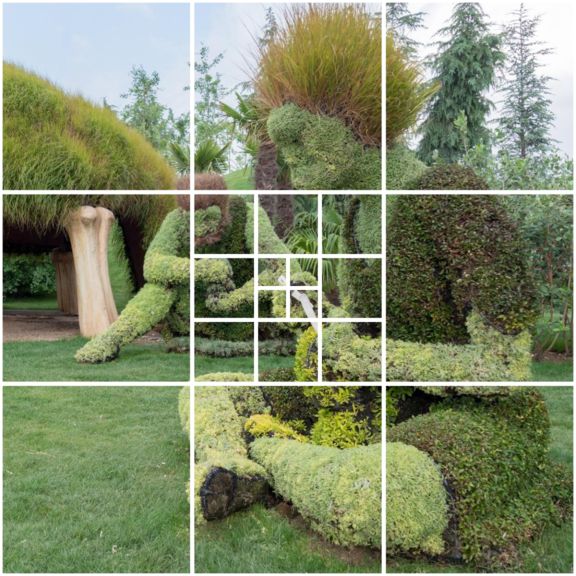}
\end{tabular}
\caption{\label{fig:toy}A visualization of the foveated tokenization employed by Segment This Thing.
Pixels are grouped into small patches in the center and progressively larger patches towards the edges.
}
\end{figure}

We introduce a new efficient segmentation model called ``Segment This Thing'' (STT) that was designed for streaming video applications.
There have been many other follow-up works aimed at making SAM more efficient, but they have all done so by reducing the model size.
We take a different approach, and instead apply a novel biologically-inspired tokenization scheme which results in a drastic reduction in token count and a concomitant increase in efficiency of the Transformer-based image encoder.
This scheme, which we refer to as ``foveated tokenization,'' is modeled after the variability of the visual acuity of human eye, which is highest in the fovea and decreases with increased eccentricity \cite{Carrasco:etal:PP95}.
Instead of dividing the image into evenly sized patches on a uniform grid, we resample patches with variable resolution on an irregular grid centered on a single point prompt (see Figure \ref{fig:toy}). 
The further a patch is from the center, the lower its resolution. 

In the analogy with the human eye, it is as if the model directs its visual attention towards the prompt point and estimates the extent of the thing it is ``looking at'' (hence the name of our model).
If the object is small or far away such that it subtends a small solid angle, it will be contained  within the highest resolution patches and thus will not be affected by the downsampling (so long as there is sufficient context from the surrounding regions).
If the object is large, it will extend into the periphery of the image and some information will be lost.
However, we argue that for larger segments, high resolution is not as important.
Our experiments show that our resampling of the image results in only moderate decreases in segmentation accuracy across a wide range of segment sizes.
Importantly, there is no need to know the size of an object before attempting to segment it.

Our foveated tokenization reduces the token count so significantly, we can actually use \textit{larger models} relative to other efficient SAM variants while still requiring \textit{fewer  FLOPs} and thus lower latency, due to the quadratic scaling of self-attention with respect to the token count.
This gives Segment This Thing more modeling flexibility and suggests that this tokenization and image encoding strategy could be applied gainfully towards other tasks besides segmentation.

In addition to reducing latency, our foveated tokenization has the benefit of reducing the bandwidth required to get pixels from the sensor to the model.
The tokenization results in a significant compression of the image, and is orthogonal to other compression schemes that could be applied to the patches (e.g. JPEG compression) for further bandwidth reduction.
Furthermore, the resampling is computationally cheap and would be relatively easy to implement efficiently in hardware.
Altogether, this makes the Segment This Thing model a promising option for power-constrained edge devices that need to stream data off-device to be segmented elsewhere.

With foveated tokenization, each prompt leads to a unique foveation and thus requires running both the image encoder and the mask decoder, whereas models like SAM can amortize the cost of the image encoder across multiple prompts.
However, our model is so much faster than the baselines that it would require a number of prompts on the same image before the cost is equivalent.
Futhermore, the need to segment many objects is more common in static image analysis and less relevant to streaming video use cases.

We benchmark Segment This Thing on a number of publicly-available segmentation datasets to quantify the performance degradation that comes with the increased efficiency.
We also show qualitative experiments in which users wear glasses with front-facing cameras and eye-tracking capabilities, applying STT to estimate what objects the user may be looking by using gaze-derived prompts.

%% file: sec_2_related.tex
\section{Related Work}
\label{sec:related}

Our work is built on Segment Anything \cite{Kirillov:etal:CVPR23}, which aims to segment any entity in an image given a prompt in the form of a bounding box or one or more 2D points.
Our model aims to address the high computational cost of running the Segment Anything Model (SAM).

\noindent \textbf{Efficient Prompted Segmentation}
A number of other papers have also introduced techniques to reduce the cost of prompted segmentation.
FastSAM \cite{Zhao:etal:arXiv23} uses an efficient CNN-based object detection and segmentation backbone to produce segmentation candidates and then uses prompts to select from and / or merge these candidates into an output mask.
SqueezeSAM \cite{Varadarajan:etal:arXiv23} employs a UNet architecture with a transformer at the lowest-resolution level.
Their early prompt fusion approach (providing prompts as additional channels in the input to the UNet rather than injecting them in a second stage) is similar to our centering of foveated inputs on the prompt, allowing the network to focus on relevant information from the start.
EfficientViT-SAM \cite{Zhang:etal:CVPR24} substitutes the ViT image encoder used by SAM for an EfficientViT \cite{Liu:etal:CVPR23} encoder.
EfficientSAM \cite{Xiong:etal:CVPR24}, EdgeSAM \cite{Zhou:etal:arXiv23}, and MobileSAM \cite{Zhang:etal:arXiv23} reduce the size of SAM by applying a knowledge distillation approach to transfer knowledge from the SAM image encoder into a lighter-weight encoder, then fine-tuning on the prompted segmentation task.
For a recent review of efficient SAM methods, see Sun \etal \cite{Sun:etal:arXiv254}.

\noindent \textbf{Prompted Video Segmentation}
One of the aims of our work is to reduce the cost of prompted segmentation such that it is reasonable to run on every frame of a streaming video.
Recently, Ravi \etal introduced SAM-2 \cite{Ravi:etal:arXiv24}, an extension of SAM to prompted video object segmentation (VOS).
However, with SAM-2 each prompt recovers a segmentation of only one object across the video, with multiple segmentations requiring multiple prompts. 
Our model can segment a different object each frame depending on where the foveation center is placed.
We leave the extension of foveated tokenization to video input to future work.

\noindent \textbf{Resolution reduction}
In the deep learning era of computer vision it has always been standard practice to rapidly reduce resolution in early network layers.
For example, AlexNet \cite{Krizhevsky:etal:NeurIPS12} begins with a stride-4 convolution followed by a stride-2 pooling, reducing an input image of size $224 \times 224$ to a feature map of size $26 \times 26$ after just these two layers.
This did not change as hardward advanced;
ResNets \cite{He:etal:ICCV16} begin with a convolutional layer and a pooling layer each with a stride of two pixels, resulting in a quartering of the image resolution after progressing through a small fraction of the network depth.
Our approach also reduces input resolution at the beginning of (in fact, before) the network, but with a rate that increases with increased distance from the prompt.

\noindent \textbf{Compression}
Other work reduces data sizes by learning from compressed representations.
For example, Park and Johnson \cite{Park:Johnson:CVPR23} train a Transformer-based image classifier directly on DCT co-efficients as used in JPEG encoding of images.
Horton \etal take this idea even further and train a classifier directly on compressed byte encodings of images \cite{Horton:etal:arXiv23}.
Other methods such as VQ-VAEs \cite{VanDenOord:etal:NeurIPS17} or the more recent Lookup-Free Quantization \cite{Yu:etal:arXiv23} allow learning a compression of input data which can then be leveraged to solve downstream tasks more efficiently.
However, depending on the specific architecture these techniques can incur significant costs to compute the compressed representations and are therefore not as widely applicable to deployment on power-constrained edge devices.
Our foveated tokenization can be seen as a simple compression of the input image, but the compression technique is extremely cheap and largely orthogonal to these other approaches (i.e. one could apply further compression to the foveated patches).

\noindent \textbf{Efficient Tokenization}
Other prior works similarly addressed the issue of large token counts in transformer architectures.
Yu \etal introduced TiTok \cite{Yu:etal:arXiv24}, which generates a concise set of 32 or 64 tokens to represent an image.
Yu \etal also make note of the ability to increase model size when token counts are low, but focused on the downstream task of image generation -- the TiTok tokenizer itself is a Vision Transformer and is thus not particularly efficient.
A large number of different token pruning or merging approaches have been proposed \cite{Bolya:etal:arXiv22, Renggli:etal:arXiv22, Meng:etal:CVPR22, Yin:etal:CVPR22, Kong:etal:ECCV22, Song:etal:arXiv22, Rao:etal:NeurIPS21, Fayyaz:etal:ECCV22, Marin:etal:arXiv21, Chen:etal:ICCV23, Yan:etal:arXiv24}.
These methods typically assume a full input image, starting with a full token set and progressively reducing the token count from there.
They therefore do not offer the same benefits resulting from input size reduction as our approach.

\noindent \textbf{Foveated Vision}
There have been some prior works applying foveation models to deep learning.
Jonnalagadda \etal introduced FoveaTer \cite{Jonnalagadda:etal:arXiv21}, which uses variably-sized pooling windows applied to the output of a convolutional network to produce foveated feature maps. 
These foveated features are then processed by a Transformer, which like our image encoder benefits from the reduction in token count.
However, the convolutional backbone of FoveaTer still requires full resolution images, which places higher constraints on sensors and bandwidth compared to our approach, in which foveation is applied as a first step (and can therefore be implemented on the sensor itself or in nearby accelerators).
Peripheral Vision Transformer \cite{Min:etal:NeurIPS22} also takes inspiration from biological vision systems, but focuses on the inductive bias aspect of foveation by modifying the position encoding of a Transformer-based image classifier.
They do not achieve efficiency gains relative to their baseline.
PeLK \cite{Chen:etal:CVPR24} applies the foveation concept to convolutional neural networks, introducing adaptive weight sharing that allocates more weights to the center of a large convolutional kernel and fewer towards the edges. 
This work actually moves in the opposite direction of ours by decreasing parameter counts; we instead decrease input size and increase parameter counts.
Furthermore, the above works are evaluated on older datasets with small images such as ImageNet \cite{Deng:etal:CVPR09}, ADE20K \cite{Zhou:etal:IJCV19}, or MS-COCO \cite{Lin:etal:ECCV14}.
With our focus on efficiency, we evaluate Segment This Thing on more modern datasets with much larger images, showing the ability to segment objects ranging in size from a handful of pixels to millions of pixels with very low latency.

Konrad \etal also explore foveated imaging with GazeGPT \cite{Konrad:etal:arXiv24}.
GazeGPT also investigates gaze-based prompting, in their case of a multimodal large language model.
This model was pre-trained on uniform resolution images, which constrains the GazeGPT foveation model (it uses three overlapping images of the same size but varying receptive field).

\begin{figure}
    \centering
    \includegraphics[width=0.28\columnwidth]{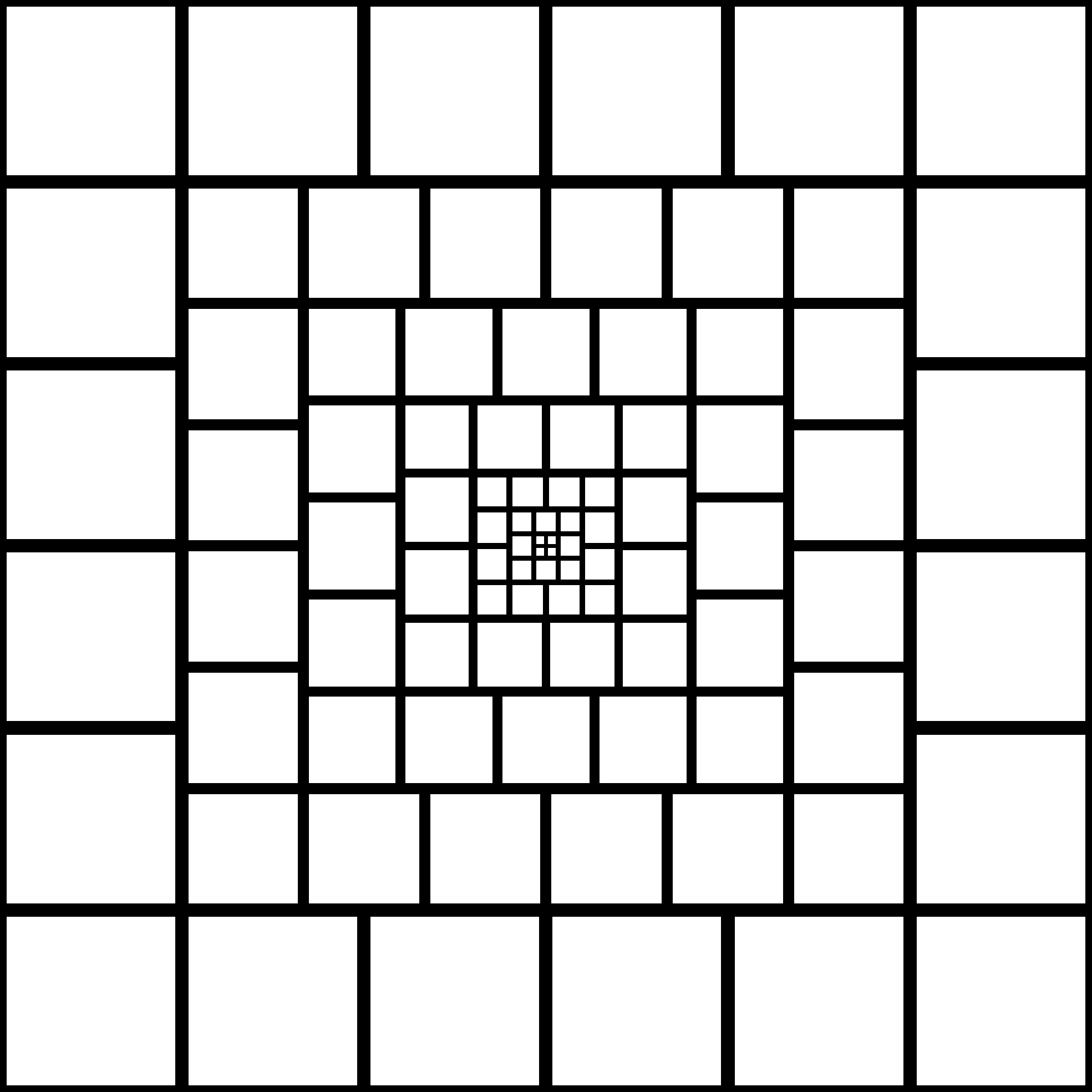}
    \hspace{7mm}
    \includegraphics[width=0.35\columnwidth]{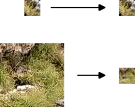}
    \caption{\label{fig:downsampling} (left) The foveation pattern used by our trained models. (right, top) A patch from the center maintains its original resolution. (right, bottom) A patch from the outer ring gets downsampled by a factor of 8. Best viewed digitally.}
\end{figure}

%% file: sec_3_model.tex
\section{The Segment This Thing Model}
\label{sec:model}

In this section we introduce the Segment This Thing model.
First, we describe the foveation model used to map full resolution images to foveated token sets.
Then we describe the architecture that maps these tokens to segmentation maps.

\begin{figure*}
    \centering
    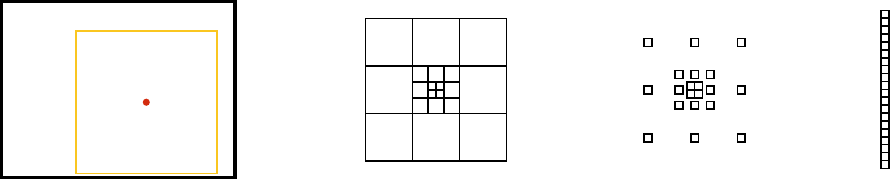
    \caption{\label{fig:foveated_tokenization} The foveated tokenization process. It begins by cropping a square region around the point prompt with a fixed size determined by the foveation pattern. This pattern divides the crop region into a set of variably-sized patches, each of which is resampled such that they are all the same size. The set is then concatenated to form a token tensor.}
\end{figure*}

\subsection{Foveated Tokenization}
\label{sec:foveated_tokenization}

Each instance of the Segment This Thing model is trained with a fixed foveation pattern which divides an image region into a set of patches.
An example of such a pattern is depicted in Figure \ref{fig:downsampling}. 
In the center of the pattern is a dense grid of patches corresponding to the center of the fovea in the analogy with human vision.
Surrounding this block is a series of concentric rings of larger patches. 
The patches in each ring are larger than patches in the preceding ring and have a size that is an integer multiple of the central patch size.
The patch sizes and the width of each ring are selected such that they can be nested without gaps or overlap\footnote{Other formulations that allow for overlap are possible; however, we judged this model to be most similar to the currently predominant uniform patch tokenization as introduced by Vision Transformers \cite{Dosovitskiy:etal:ICLR20} and thus a good starting point to explore foveated tokenization.}.

Given an image and a point prompt, the first step is to extract a crop centered on the prompt and sized according to the pattern.
If the crop region extends beyond the image boundary, we pad the image. 
Next, each patch in the foveation pattern is downsampled to the central patch size.
This can be implemented efficiently using an integral image, which is equivalent to applying a box filter and subsampling with integer strides.
This simplistic filter results in lower image quality than more sophisticated methods, but is hardware-friendly such that it could be deployed on power-constrained edge devices.
The downsampling process is visualized in Figure \ref{fig:downsampling}.
We follow SAM and use $16 \times 16$ patches.

After downsampling, the patches are all the same size, as is the output of the standard patch tokenization used \eg in Vision Transformers \cite{Dosovitskiy:etal:ICLR20}.
We can thus flatten each patch into a vector and stack them to form a token matrix, hence the name foveated tokenization.
The token set is then passed to the Segment This Thing model for processing.
The full tokenization pipeline is depicted in Figure~\ref{fig:foveated_tokenization}.

\subsubsection{Visualization}
\label{sec:visualization}
It is often useful to visualize the foveated input to Segment This Thing.
However, after foveated tokenization the image is represented as a $N \times T \times T$ tensor,  where $N$ is the token count and $T$ is the patch size.
To visualize such a tensor, we reverse the tokenization process: patches are upsampled back to their original size (typically using nearest neighbor interpolation) at which point they can be recomposed into an image.
Note that while this process does not re-introduce any new information and thus provides a faithful view of what the network sees, it does introduce some distortion --- the central regions take up much more space in the network input than in the visualizations thereof.

\begin{figure}[hb!]
    \centering
    \begin{tabular}{c c}
    \multirow{4}{*}{\includegraphics[height=0.45\columnwidth, valign=t]{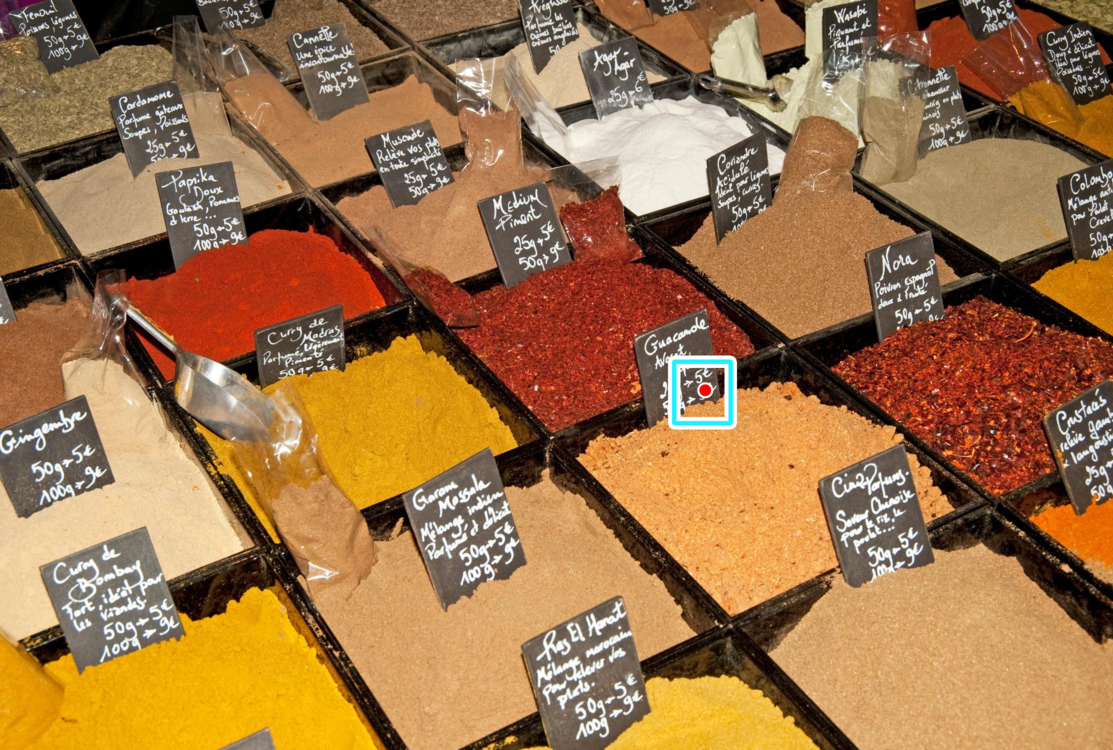}} & SAM \\
    & \includegraphics[height=0.175\columnwidth, valign=t]{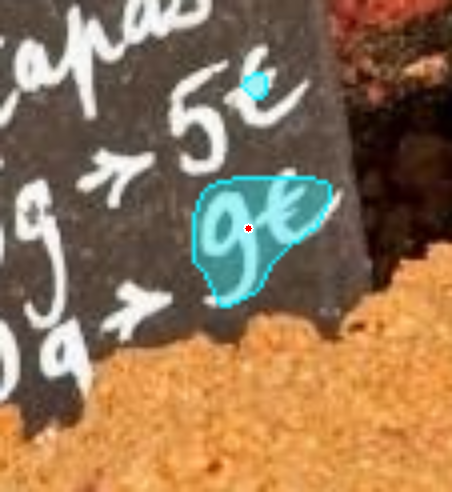} \\
    & STT \\
    & \includegraphics[height=0.175\columnwidth, valign=t]{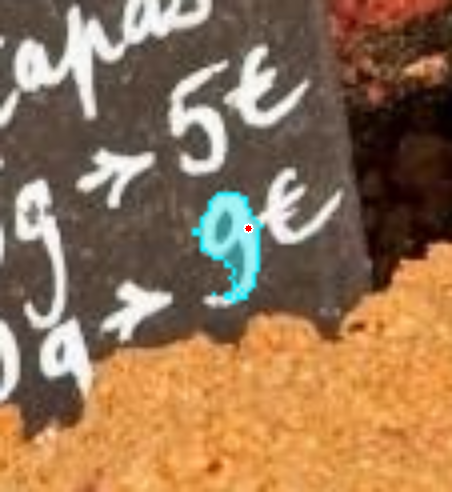}
    \end{tabular}
    \caption{\label{fig:tiny_segment}Segment This Thing has a precision advantage for extremely small targets due to the additional decovolutional layers in the mask decoder. \textbf{Left:} an input image with prompt indicated in red. \textbf{Right:} a visualization of the segmentation output of SAM and STT. The crop corresponds to the region indicated by a rectangle at left. STT is able to output a smaller segment than SAM due to its higher resolution output in the central tokens. Note that this is not the most confident output mask for either model.}
\end{figure}

\begin{figure*}[ht]
    \centering
    \includegraphics[height=0.25\columnwidth]{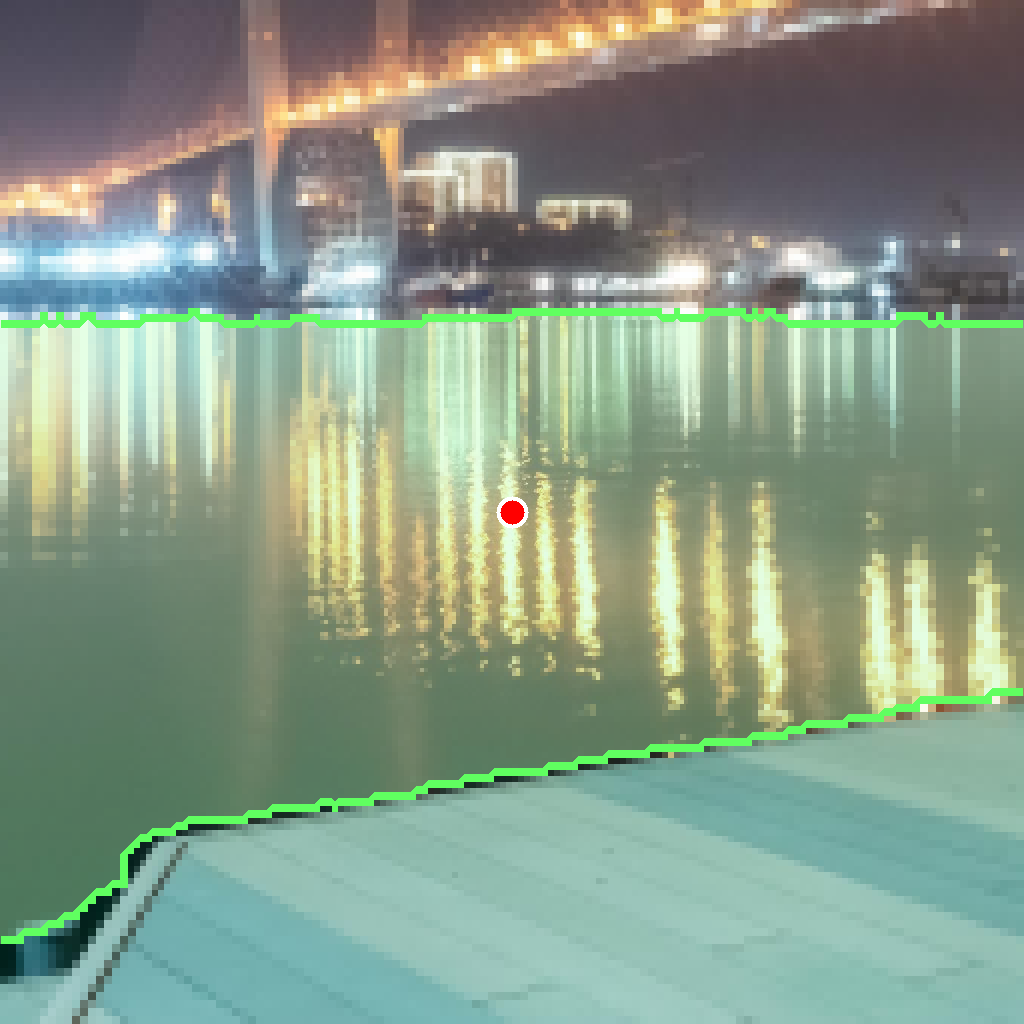}
    \includegraphics[height=0.25\columnwidth]{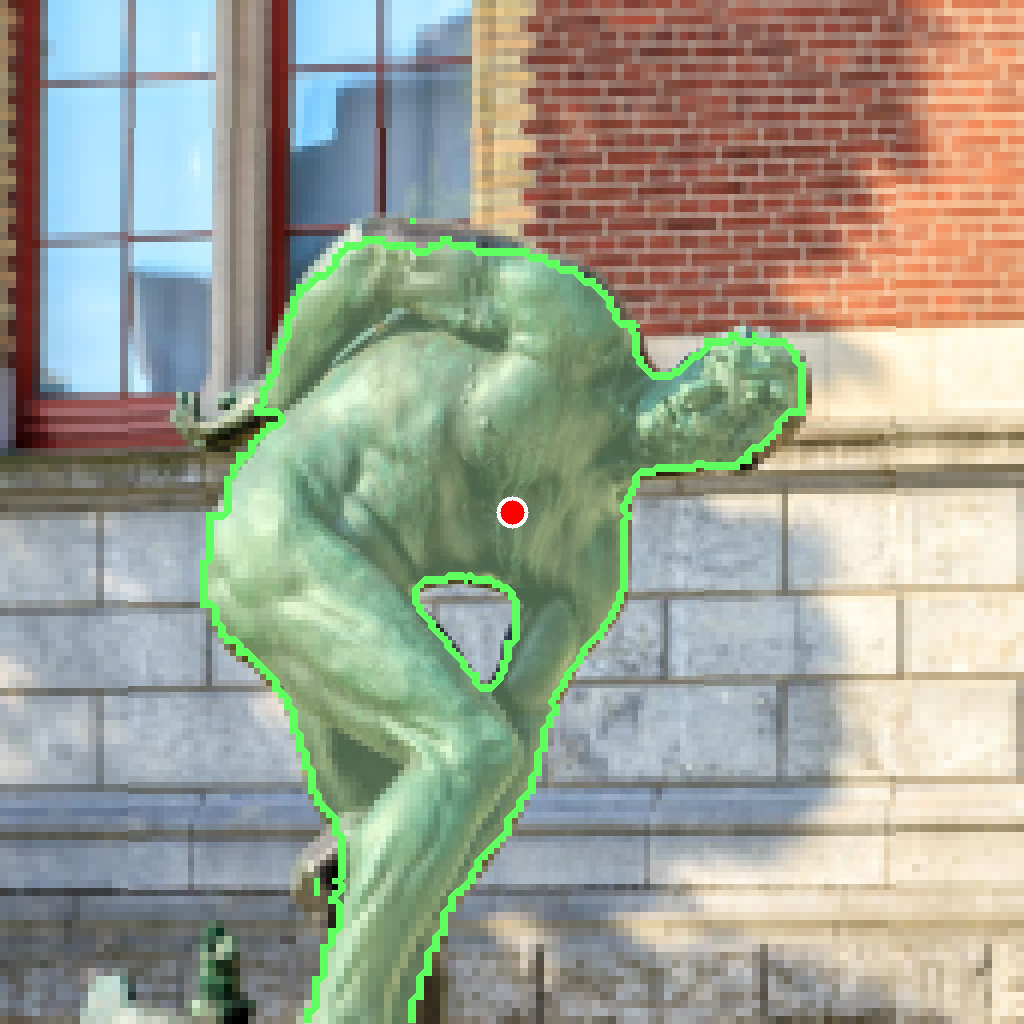}
    \includegraphics[height=0.25\columnwidth]{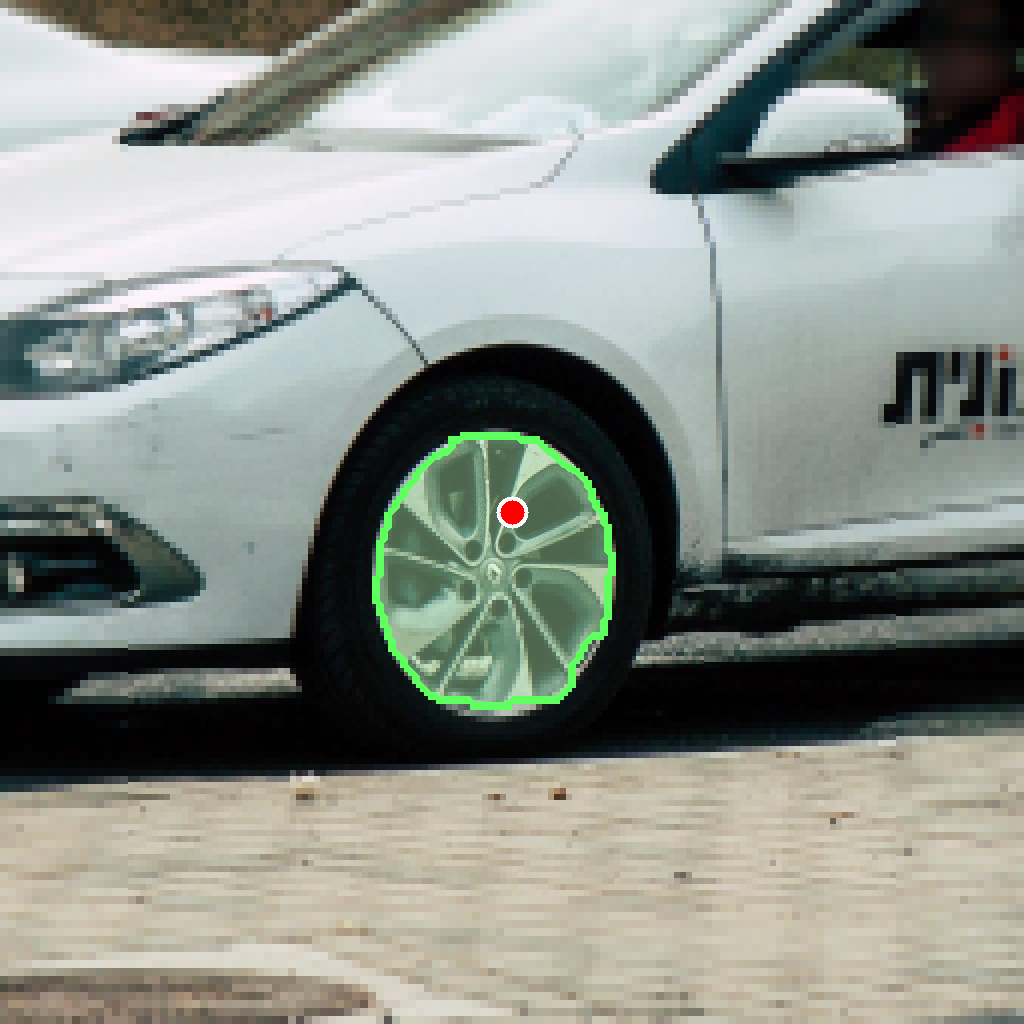}
    \includegraphics[height=0.25\columnwidth]{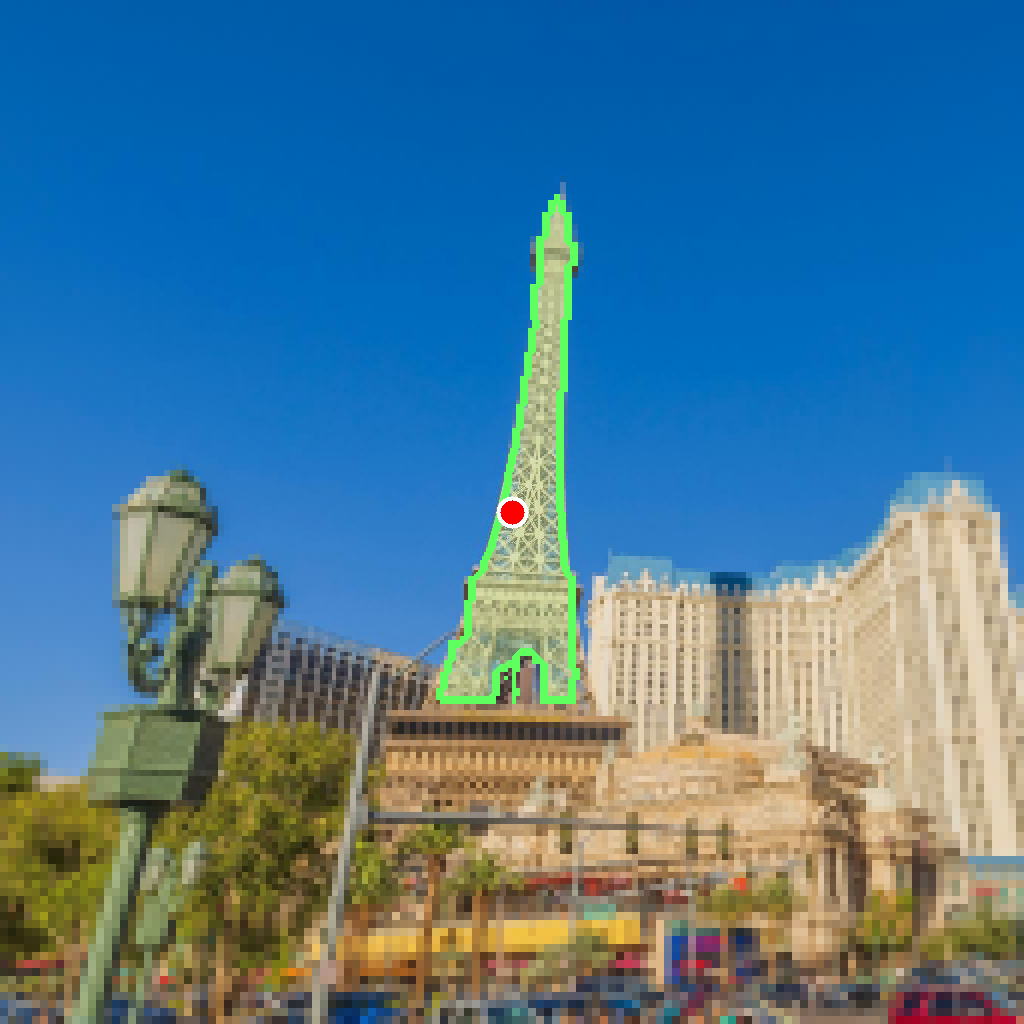}
    \includegraphics[height=0.25\columnwidth]{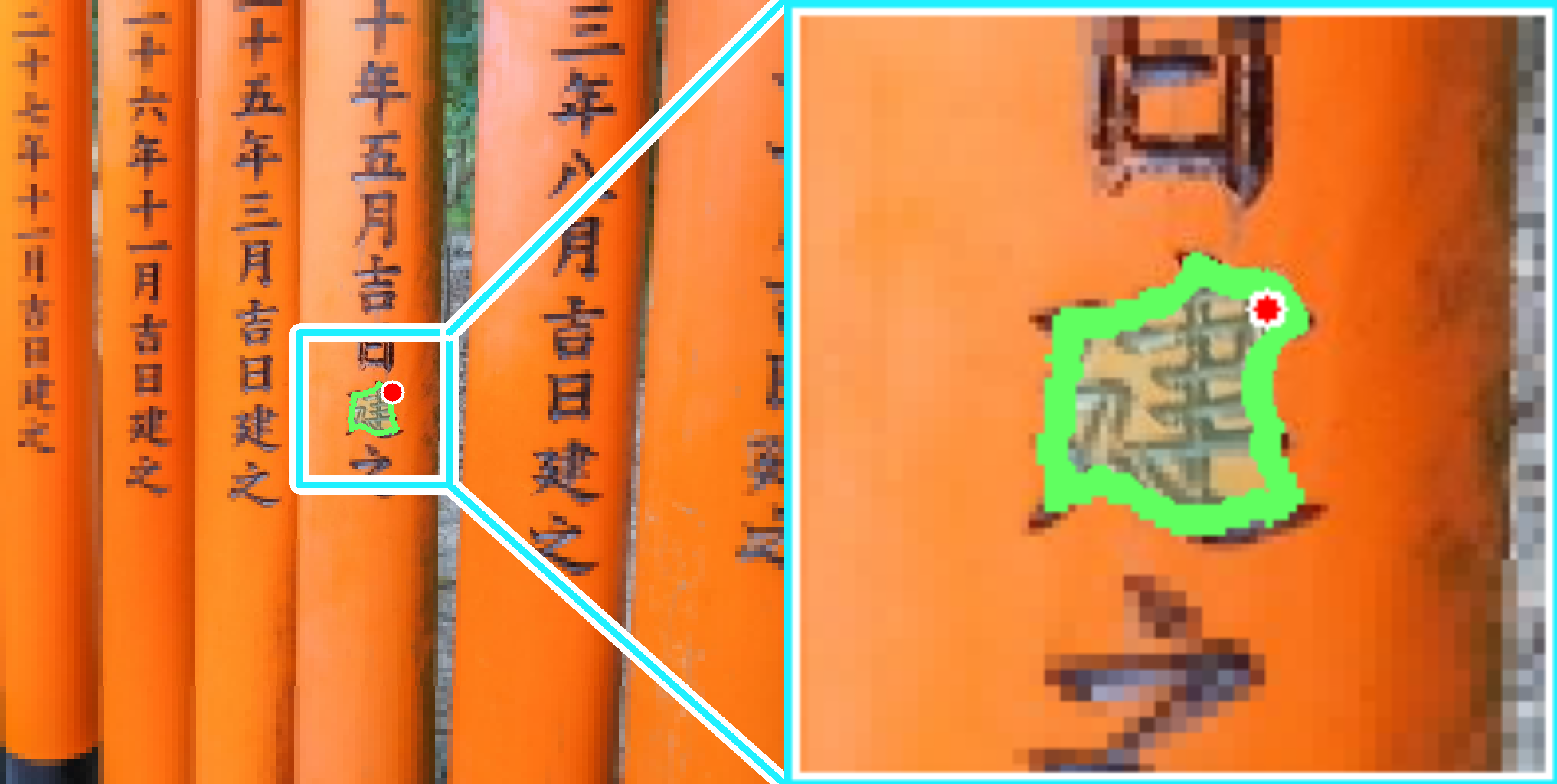}
    \includegraphics[height=0.25\columnwidth]{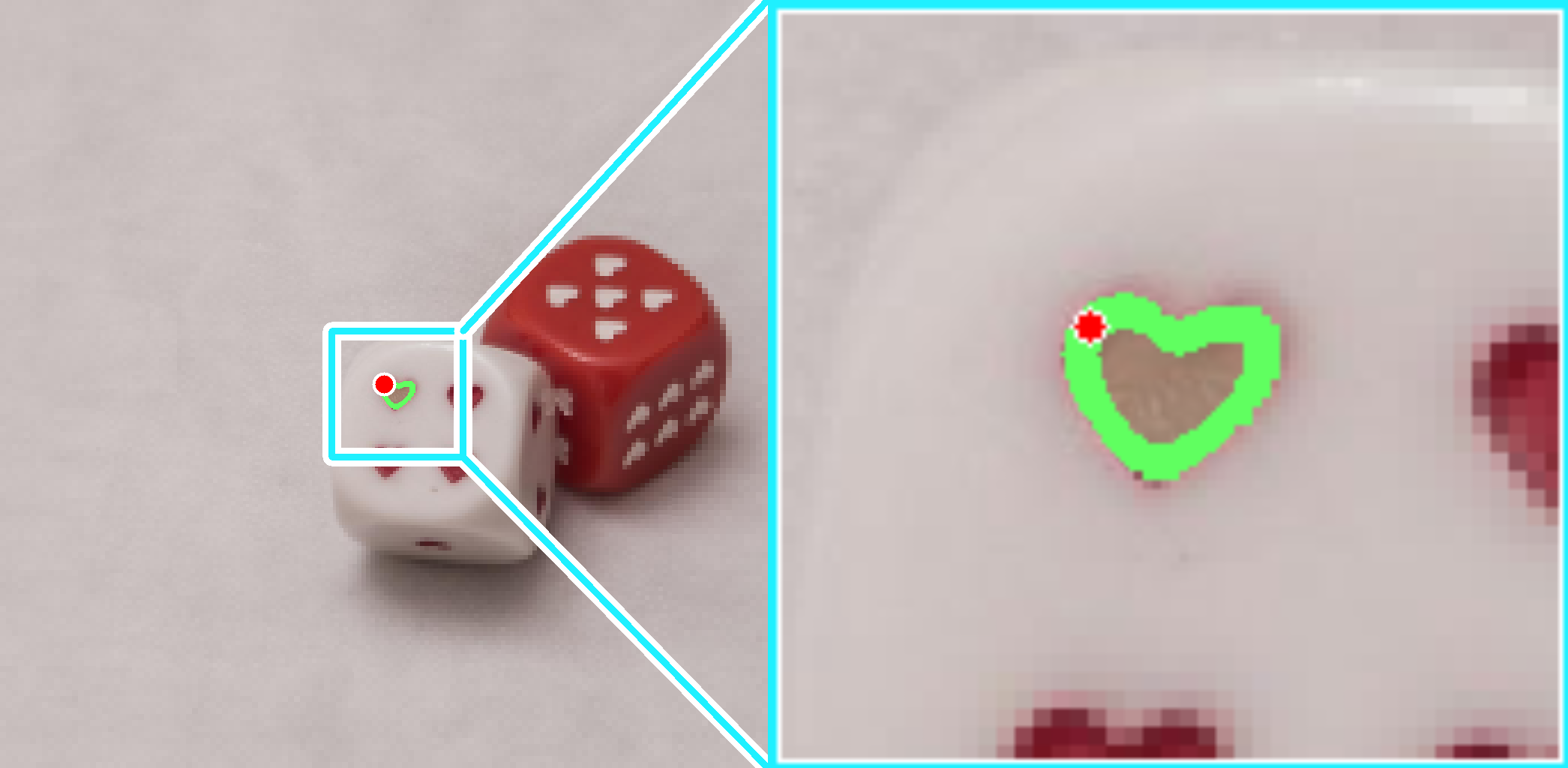} \\
    \includegraphics[height=0.25\columnwidth]{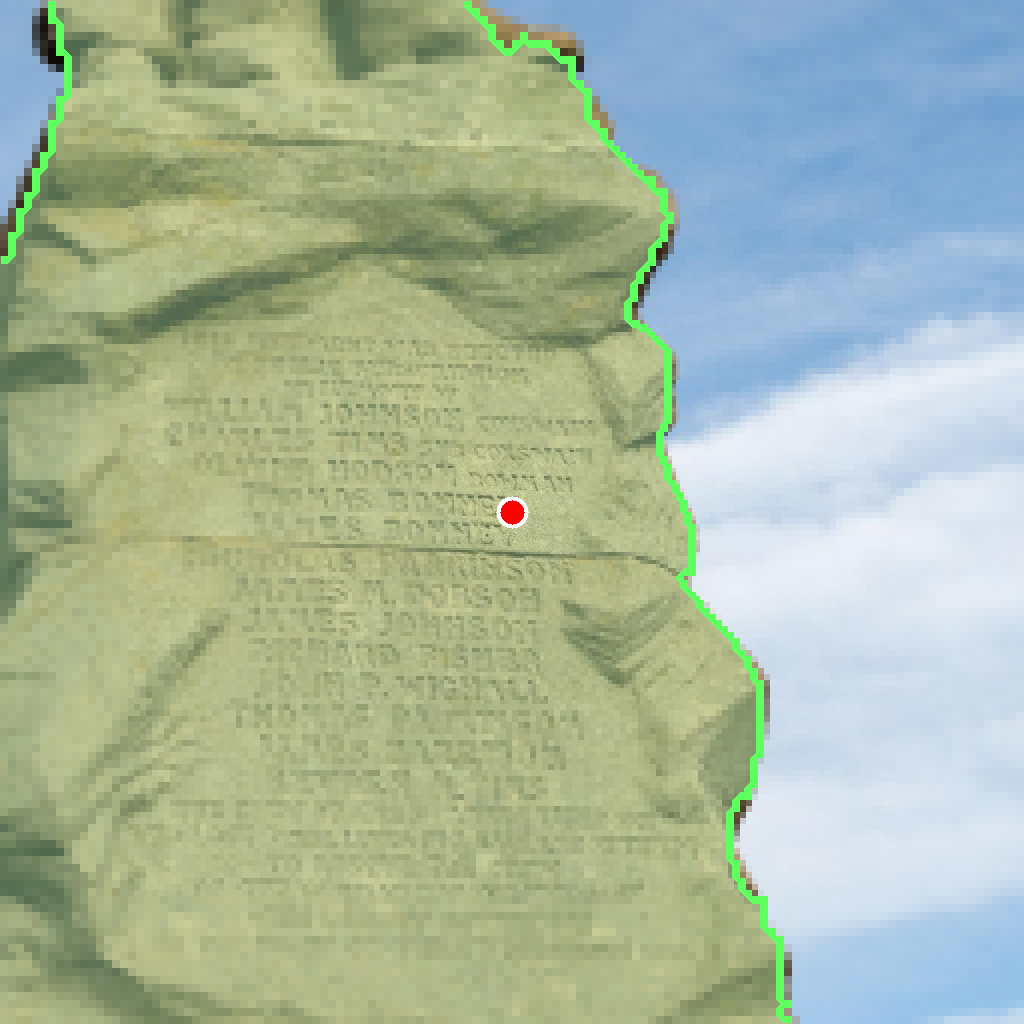}
    \includegraphics[height=0.25\columnwidth]{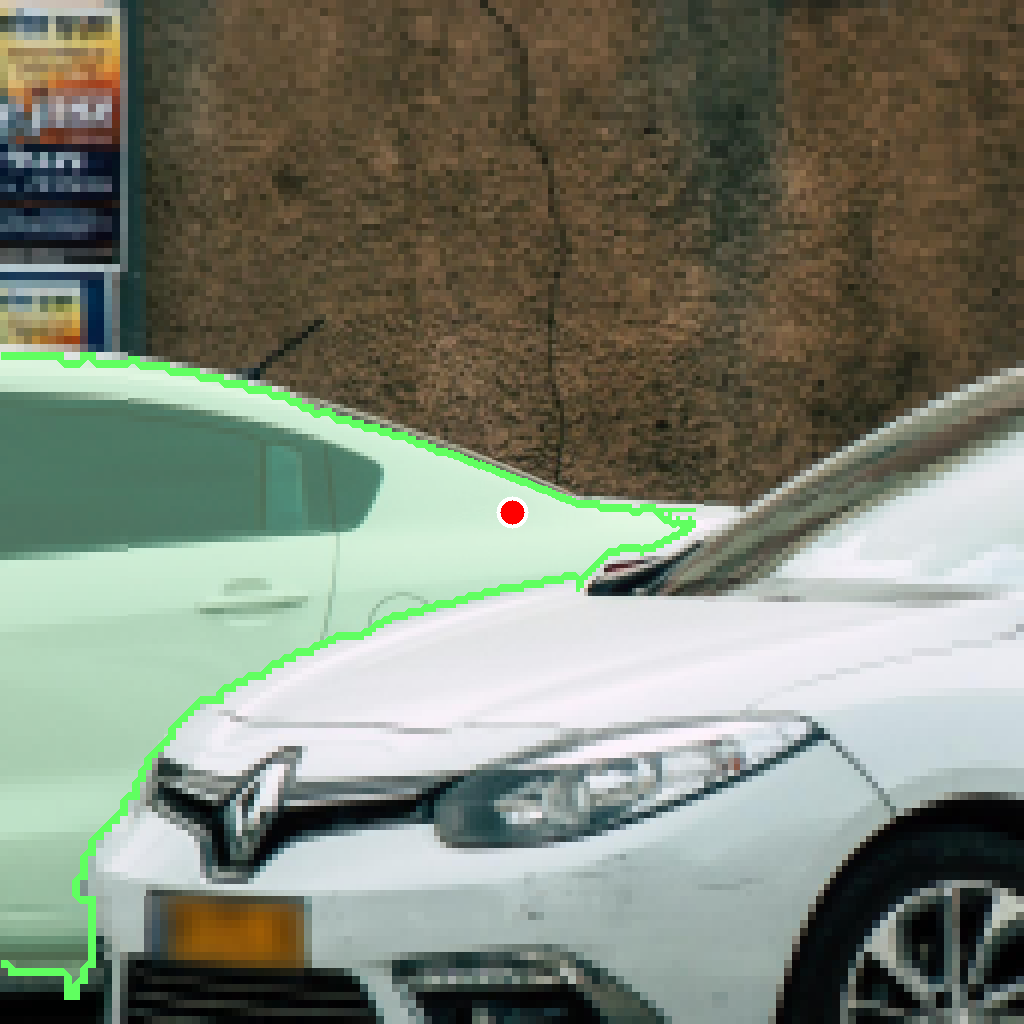} 
    \includegraphics[height=0.25\columnwidth]{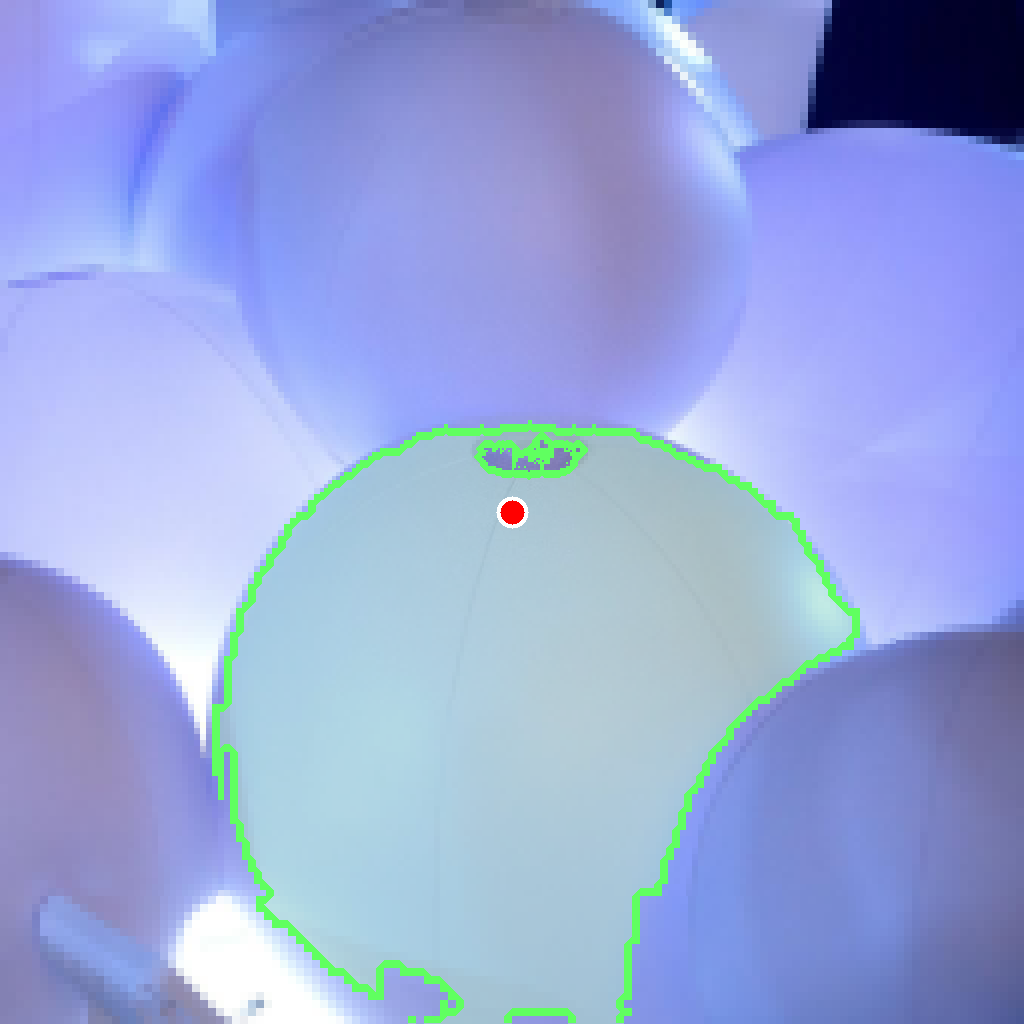}
    \includegraphics[height=0.25\columnwidth]{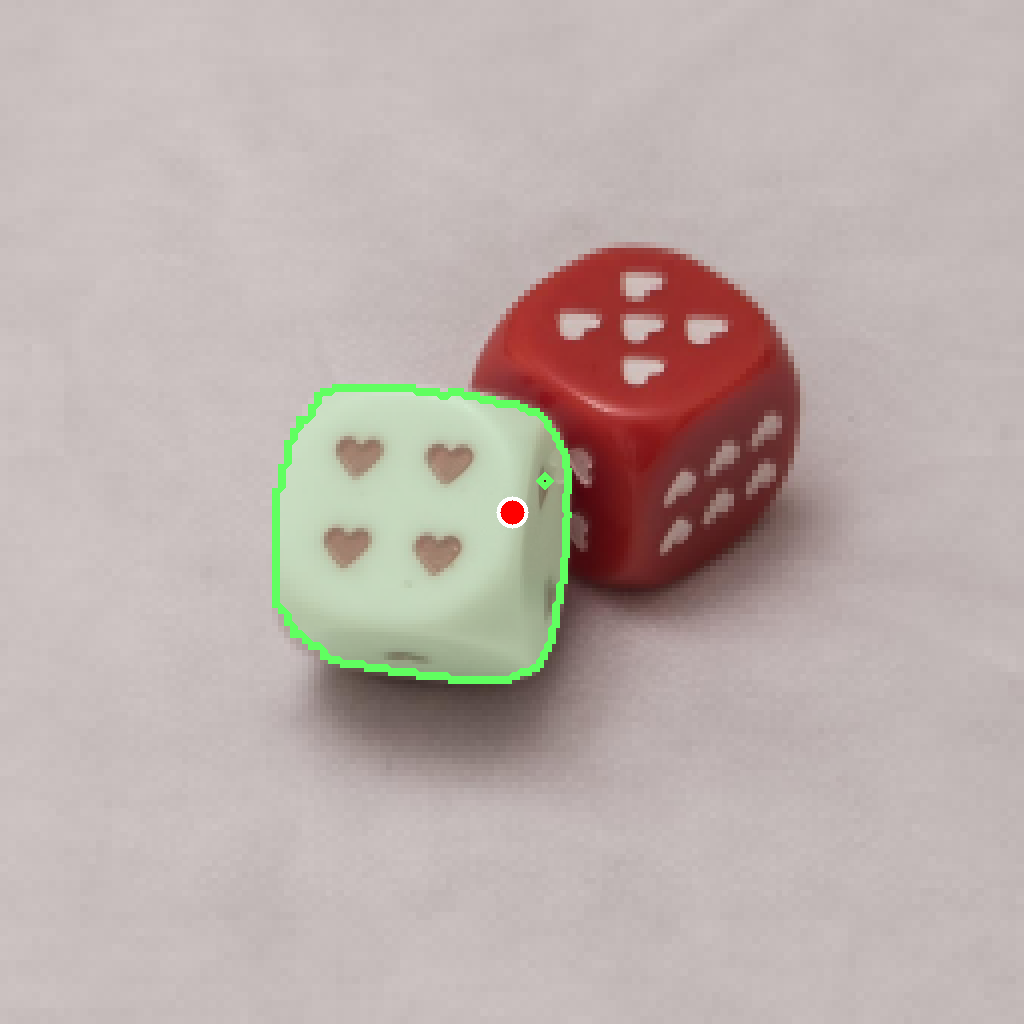}
    \includegraphics[height=0.245\columnwidth]{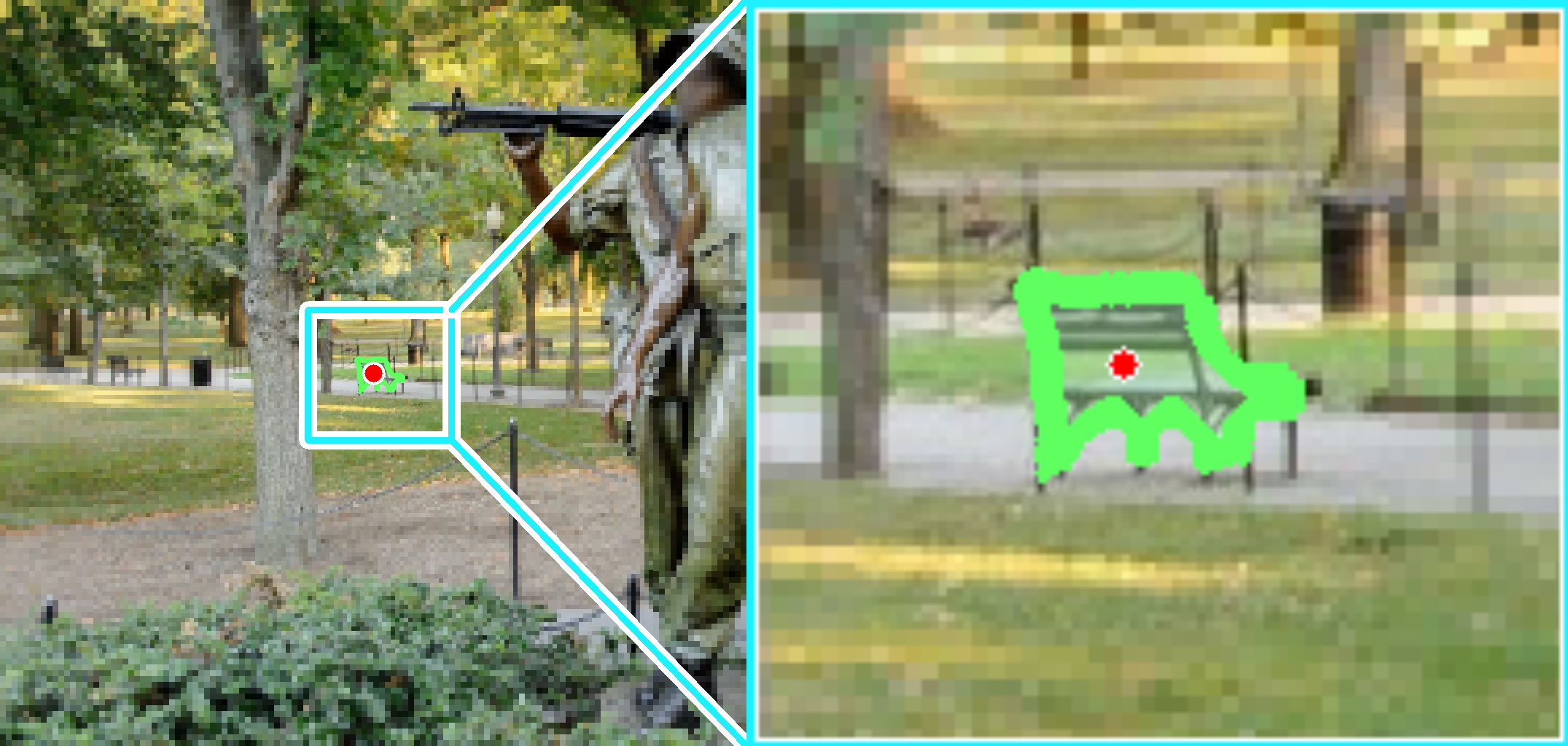}
    \includegraphics[height=0.25\columnwidth]{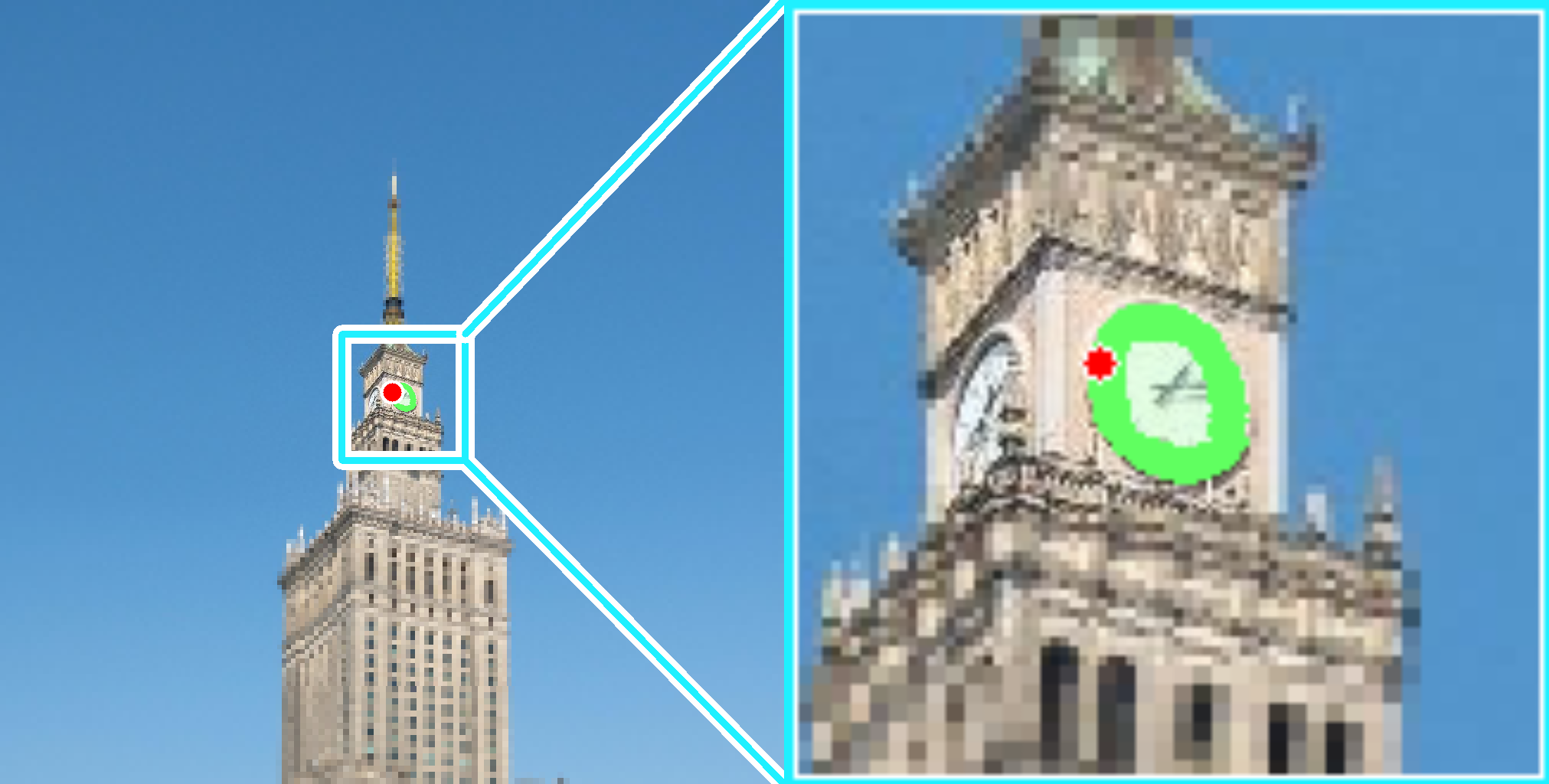}
    \caption{\label{fig:scale_range} Despite the high downsampling rate in the periphery, the Segment This Thing model is capable of segmenting objects with a wide range of shapes and sizes, from those covering most of the receptive field (left) to those covering just a handful of pixels (right). All images are from SA-1B. They have been passed through foveated tokenization and are visualized as described in Section \ref{sec:visualization}. }
\end{figure*}

\subsection{Segment This Thing}
Next we describe the architecture of the Segment This Thing model. 
It is largely based on the three-part SAM model.
However, the prompt encoder has been omitted because the prompt is implicit in the centering of the crop.

\subsubsection{Image Encoder}
Due to the foveation, the input tokens to STT correspond to patches which are not arrayed on a 2D grid and thus the Vision Transformer model used by SAM (with local 2D windowed attention) is not applicable.
However, the complexity of windowed attention was used in SAM to control compute cost, and our drastic cut in the token count removes that imperative.
The image encoder for STT is thus a standard Transformer model.
We simply project the patches (after downsampling to a common patch size) with a single linear layer, add learned position encodings, and pass them through the transformer to obtain a (foveated) feature map.
We also include a register token to allow the model to collect and broadcast information, following Darcet et al. \cite{Darcet:etal:arXiv23}.
As the foveation pattern can extend beyond image boundaries, we also add a token mask to prevent the participation of out-of-bounds tokens in self-attention.

\subsubsection{MaskDecoder}
\label{sec:mask_decoder}
The mask decoder is also modeled on its counterpart in SAM, again with some simplifications.
We do not support iterative segmentation -- the model is always run just once per prompt.
In SAM, the model outputs $N$ (typically $N = 3$) masks when prompted with a single point (which leads to scale ambiguity) and one mask when prompted with multiple points or a bounding box.
Segment This Thing only supports single point prompts, so the decoder always outputs $N$ masks.
The query token set is thus a set of $N + 1$ tokens, one for the intersection over union (IoU) prediction and one for each predicted mask.
The context token set is the output of the encoder (including the register token).
Out-of-bounds context tokens are masked during cross attention with the same token mask used in the encoder.
The decoder otherwise operates in the same way as its counterpart in SAM, applying a two-way cross-attention transformer, followed by learned upsampling of the image tokens, and finally a dot product between the mask tokens (transformed by an MLP) and the image token maps to produce the segmentation map.

In SAM, the segmentation maps are estimated at a quarter of the resolution of the input image by applying two $2 \times 2$ deconvolutions with stride 2 to the token feature map.
In other words, a $4 \times 4$ segmentation label patch is estimated for each $16 \times 16$ token.
If full resolution maps are required, they are computed using bi-linear interpolation.
This design decision was likely made to reduce memory costs during training.

\begin{figure}[b]
    \centering
    \includegraphics[width=0.3\columnwidth]{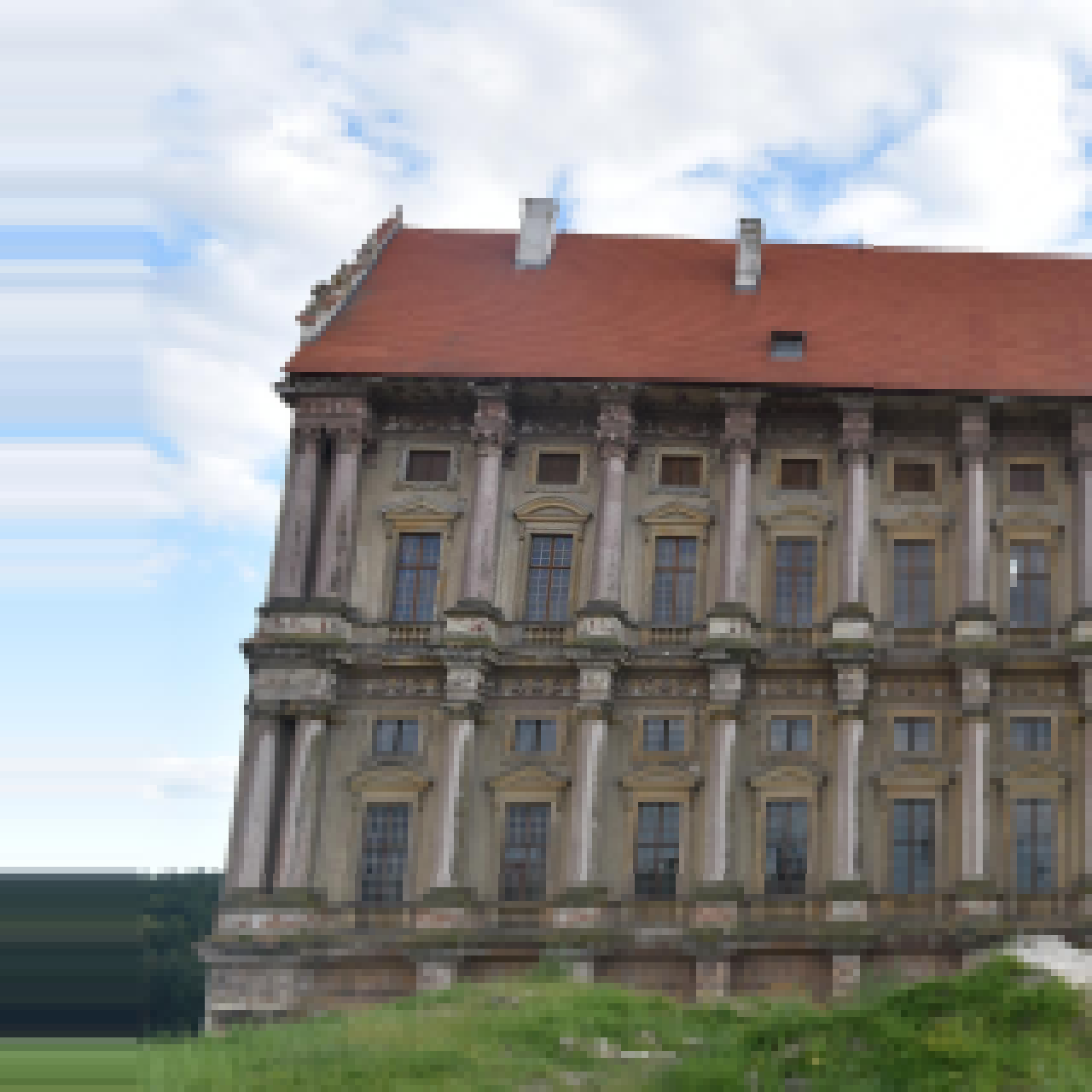}
    \includegraphics[width=0.3\columnwidth]{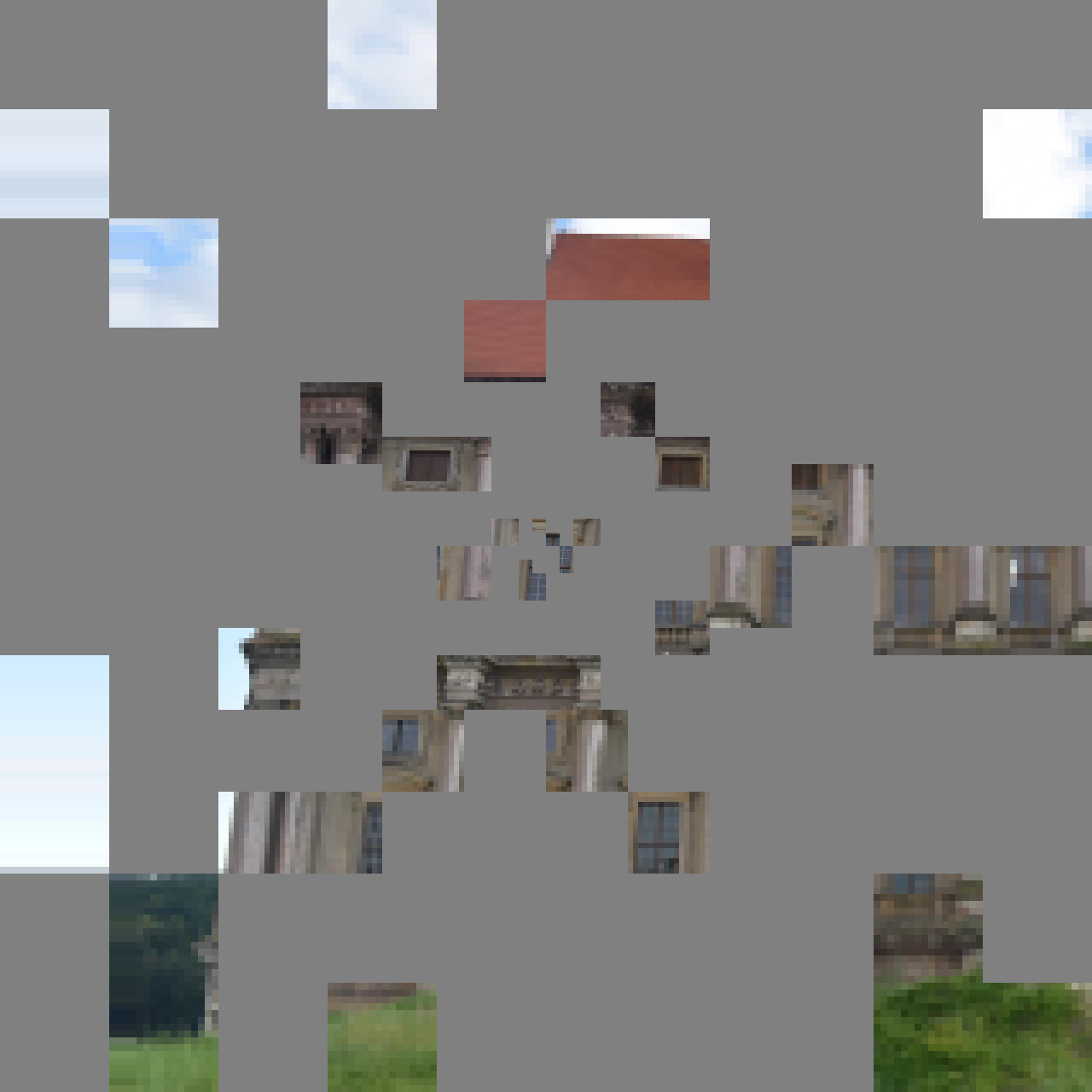}
    \includegraphics[width=0.3\columnwidth]{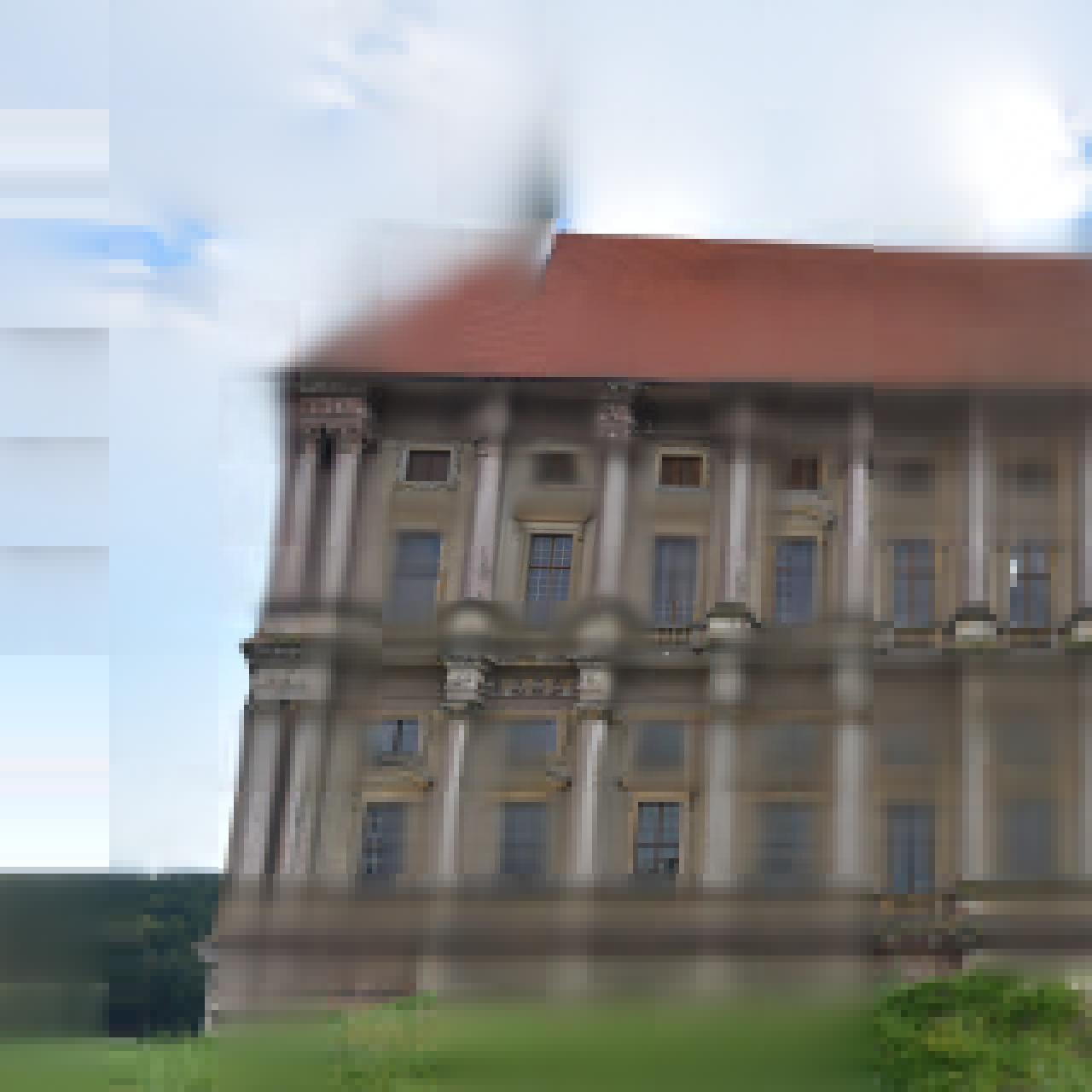}
    \caption{A visualization of a training example for MAE pre-training of the image encoder. \label{fig:mae}\textbf{Left:} the foveated input image (and reconstruction target), visualized as described in \ref{sec:visualization}. \textbf{Center:} a visualization of the token set passed to the encoder. \textbf{Right:} a visualization of the reconstruction produced by the network. Note that the output resolution of the decoder matches the target image pixel-for-pixel, i.e. it is also foveated.}
\end{figure}

Another benefit of our lower token count is that we can afford to produce a full $16 \times 16$ segmentation map per patch by stacking four deconvolutions, each of which double the resolution. 
Thus the labels for the central patches are estimated at full resolution, with the output resolution dropping off towards the periphery at the same rate as the input resolution.
If a full resolution segmentation map is required, it can be computed by interpolating each patch such that it matches its receptive field and then restructuring the patches to form an image.
Note that this means that the output resolution of STT is actually \textit{higher than SAM} in the center of the segmentation map, at parity in the surrounding areas, and only at a lower resolution in the periphery.
We have observed at least some cases where this allows STT to segment extremely small entities that SAM misses (see Figure \ref{fig:tiny_segment}; we suspect that this capability is limited by insufficiently granular labels in SA-1B, caused in part by the use of SAM itself to generate the masks).
STT is also able to segment large objects, including segments that fill most or all of the foveation pattern, albeit at lower resolution.
See Figure \ref{fig:scale_range} for examples of segments estimated by STT across a wide range of scales.

\begin{table}[b]
    \centering
    \begin{tabular}{c|c|c|c}
      \textbf{Model} &  \textbf{Tokens} & \textbf{Pixels} & \textbf{Receptive Field} \\
      \midrule
    SAM \cite{Kirillov:etal:CVPR23}     & 4096 & 1.049M & $1024^2$ \\
    STT (Ours) & 172 & 0.044M & $1280^2$
    \end{tabular}
    \caption{\label{tab:input}A comparison of the input size for SAM and STT.}
\end{table}

%% file: figs_pipeline.pdf_tex
%% Creator: Inkscape 1.4 (e7c3feb1, 2024-10-09), www.inkscape.org
%% PDF/EPS/PS + LaTeX output extension by Johan Engelen, 2010
%% Accompanies image file 'pipeline.pdf' (pdf, eps, ps)
%%
%% To include the image in your LaTeX document, write
%%   \input{<filename>.pdf_tex}
%%  instead of
%%   \includegraphics{<filename>.pdf}
%% To scale the image, write
%%   \def\svgwidth{<desired width>}
%%   \input{<filename>.pdf_tex}
%%  instead of
%%   \includegraphics[width=<desired width>]{<filename>.pdf}
%%
%% Images with a different path to the parent latex file can
%% be accessed with the `import' package (which may need to be
%% installed) using
%%   \usepackage{import}
%% in the preamble, and then including the image with
%%   \import{<path to file>}{<filename>.pdf_tex}
%% Alternatively, one can specify
%%   \graphicspath{{<path to file>/}}
%% 
%% For more information, please see info/svg-inkscape on CTAN:
%%   http://tug.ctan.org/tex-archive/info/svg-inkscape
%%
\begingroup%
  \makeatletter%
  \providecommand\color[2][]{%
    \errmessage{(Inkscape) Color is used for the text in Inkscape, but the package 'color.sty' is not loaded}%
    \renewcommand\color[2][]{}%
  }%
  \providecommand\transparent[1]{%
    \errmessage{(Inkscape) Transparency is used (non-zero) for the text in Inkscape, but the package 'transparent.sty' is not loaded}%
    \renewcommand\transparent[1]{}%
  }%
  \providecommand\rotatebox[2]{#2}%
  \newcommand*\fsize{\dimexpr\f@size pt\relax}%
  \newcommand*\lineheight[1]{\fontsize{\fsize}{#1\fsize}\selectfont}%
  \ifx\svgwidth\undefined%
    \setlength{\unitlength}{426.49700015bp}%
    \ifx\svgscale\undefined%
      \relax%
    \else%
      \setlength{\unitlength}{\unitlength * \real{\svgscale}}%
    \fi%
  \else%
    \setlength{\unitlength}{\svgwidth}%
  \fi%
  \global\let\svgwidth\undefined%
  \global\let\svgscale\undefined%
  \makeatother%
  \begin{picture}(1,0.19940005)%
    \lineheight{1}%
    \setlength\tabcolsep{0pt}%
    \put(0,0){\includegraphics[width=\unitlength,page=1]{figs_pipeline.pdf}}%
    \put(0.2812951,0.14224963){\color[rgb]{0,0,0}\makebox(0,0)[lt]{\lineheight{1.25}\smash{\begin{tabular}[t]{l}Crop and \\Patchify\end{tabular}}}}%
    \put(0.13022441,0.05788299){\color[rgb]{0,0,0}\makebox(0,0)[lt]{\lineheight{1.25}\smash{\begin{tabular}[t]{l}Prompt\end{tabular}}}}%
    \put(0.59102654,0.11903398){\color[rgb]{0,0,0}\makebox(0,0)[lt]{\lineheight{1.25}\smash{\begin{tabular}[t]{l}Resample\end{tabular}}}}%
    \put(0.85612096,0.11926295){\color[rgb]{0,0,0}\makebox(0,0)[lt]{\lineheight{1.25}\smash{\begin{tabular}[t]{l}Concatenate\end{tabular}}}}%
    \put(0,0){\includegraphics[width=\unitlength,page=2]{figs_pipeline.pdf}}%
  \end{picture}%
\endgroup%

%% file: sec_4_experiments.tex
\section{Experiments}
\label{sec:experiments}

\begin{figure*}
    \begin{center}
        \input{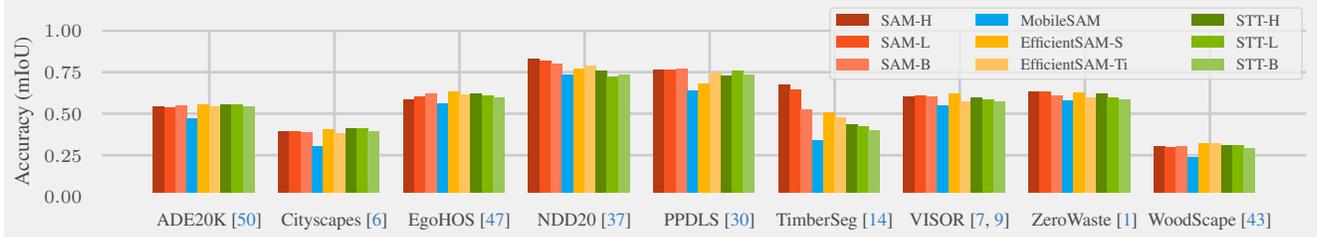}
    \end{center}
    \caption{\label{fig:main_results} 
    Results of evaluating our models and baselines on publicly available segmentation datasets. Models of the same family are grouped by hue, with larger models in darker shades. Results vary widely by dataset, but the trend is consistent and shows that STT outperforms MobileSAM, which runs with similar latency, and is competitive with more expensive models. See Table \ref{tab:full_results} in the supplementary for a full tabular listing of the results.
    }
\end{figure*}

In this section we describe how we train the Segment This Thing model, and present evaluation results.

\subsection{Training}
We trained our STT models using a foveation pattern with 172 tokens, a total size of 1280 pixels, and a stride of 8 pixels in the outer ring.
See the supplementary materials for details.
In Table \ref{tab:input} we compare the input sizes for SAM and our STT model based on this pattern.

Following SAM, we trained our Segment This Thing model exclusively on the SA-1B dataset \cite{Kirillov:etal:CVPR23}.
However, because of the novel tokenization we cannot rely on an existing pre-trained image encoder as do Kirillov \etal.
Thus the first step in training STT is masked autoencoder (MAE) \cite{He:etal:CVPR22} pre-training of the image encoder.
Other than the novel tokenization, we follow the standard MAE pre-training procedure.
We center foveations at points randomly selected from SA-1B images that are more than a threshold distance from the image boundary.
The reconstruction loss is applied to the in-bounds and downsampled tokens --- we do not ask the network to predict the full-resolution source image.
An MAE pre-training example is visualized in Figure \ref{fig:mae}.

After pre-training, the decoder is discarded and the encoder weights are used to initialize the STT encoder.
We then use SA-1B to train the mask decoder and fine-tune the encoder weights for the segmentation task.
For each image, we select a set of segments uniformly at random, and for each segment select a foveation center uniformly within the segment.
We use the same combination of Focal, Dice, and IoU prediction losses as SAM. 
However, we made two minor changes to handle the variable resolution of the STT output masks.

First, each pixel in the STT output may represent a classification of many pixels in the binary segmentation target.
Rather than upsampling the masks to the full image resolution as described in section \ref{sec:mask_decoder}, we map the ground truth to the foveated output space of STT.
This is done by converting the masks to floating point and applying the same downsampling approach as used for input pixels described in section \ref{sec:foveated_tokenization}.
The result is a real-valued foveated segmentation map.
Each value in the map has a variably-sized receptive field and represents the probability that a pixel selected uniformly at random from within that receptive field has a positive label in the full-resolution ground truth segmentation map.
After downsampling there is a 1:1 correspondence between pixels in target and estimated masks, and we can apply the Dice and Focal losses, both of which generalize to continuous target values.

The second change is to the targets for IoU prediction. 
We found that thresholding the target and estimated masks and computing IoU of the binarized masks led to unstable loss values when significant portions of either mask are near a value of 0.5.
We thus use an \textit{expected} IoU of the real-valued segmentation maps.
Given an estimated probability $p_i$ that pixel $i$ is part of the segment and probability $q_i$ that a pixel selected uniformly at random from its receptive field has a positive label, the expected IoU is given by:
\begin{equation}
    \frac{p_i q_i}{1 - (1 - p_i)(1 - p_q)}
\end{equation}
See the supplementary materials for further training details, including hyperparameter settings.

\begin{table}
  \centering
  \begin{tabular}{c c|c c}
    \toprule
    \multicolumn{2}{c|}{\textbf{Method}} & \textbf{Latency (ms)} & \textbf{GFLOPs} \\
    \toprule
    \multirow{3}{*}{SAM \cite{Kirillov:etal:CVPR23}} & H & 572.7 & 6533.7 \\
    & L & 347.6 & 3244.5 \\
    & B & 153.9 & 1027.0 \\
    \midrule
    \multirow{2}{*}{EfficientSAM \cite{Xiong:etal:CVPR24}} & S & 78.6 & 489.4 \\
    & Ti & 39.8 & 201.2 \\
    \midrule
    MobileSAM \cite{Zhang:etal:arXiv23} & & 20.7 & 124.4 \\
    \midrule
    \multirow{2}{*}{STT (Ours)} & H & 26.2 & 223.2 \\
    & L & 13.7 & 108.0 \\
    & B & 7.3 & 30.9 \\
    \bottomrule
  \end{tabular}
  \caption{Comparing the efficiency of evaluated segmentation models. Latency is measured as the time to process a single image with a single prompt on an nVidia RTX 3080 GPU, and excludes preprocessing time for all models. See supplementary materials for details on calculating GFLOPs. Note that latency is highly dependent on hardware (and, to a lesser extent, software) and that there are competing ways to compute FLOPs; we were careful to run all models on the same hardware and used the same method of computing FLOPs, and can see that FLOPs and latency are roughly linearly correlated as expected. }
  \label{tab:efficiency}
\end{table}

\begin{figure*}
    \centering
    \small
    \begin{tabular}{c c c c c}
        SAM & EfficientSAM & MobileSAM & STT (Ours) & Foveated Input \\
        \includegraphics[width=0.18\textwidth]{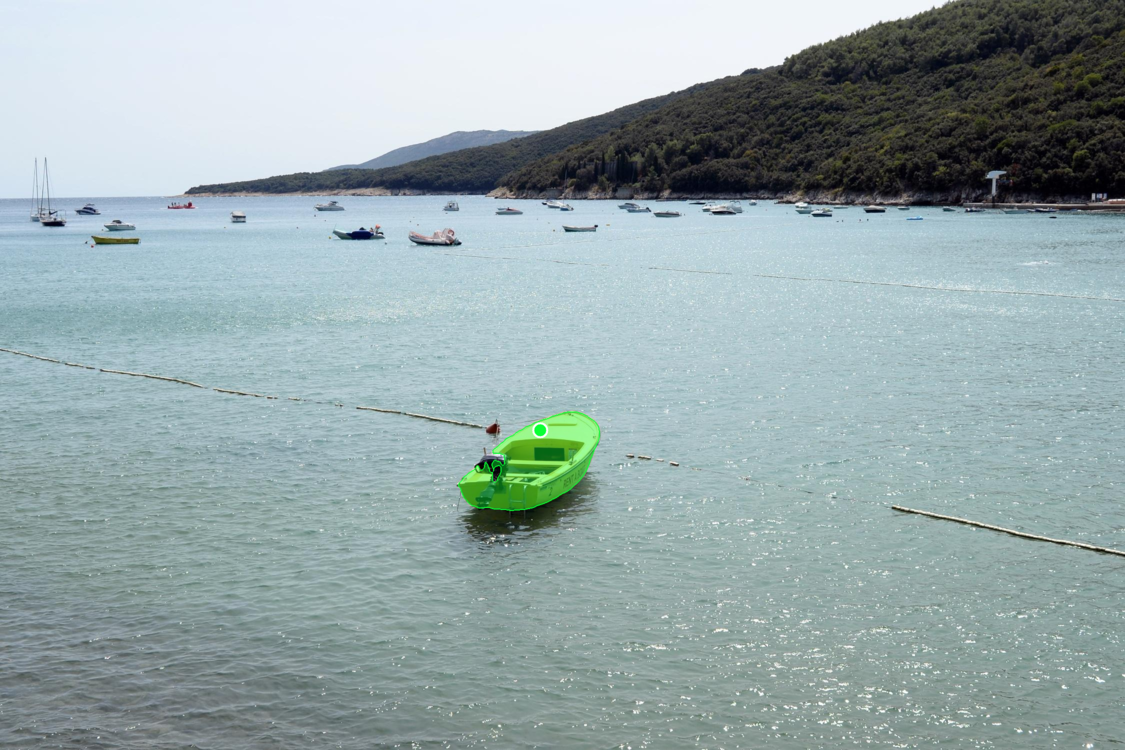} & 
        \includegraphics[width=0.18\textwidth]{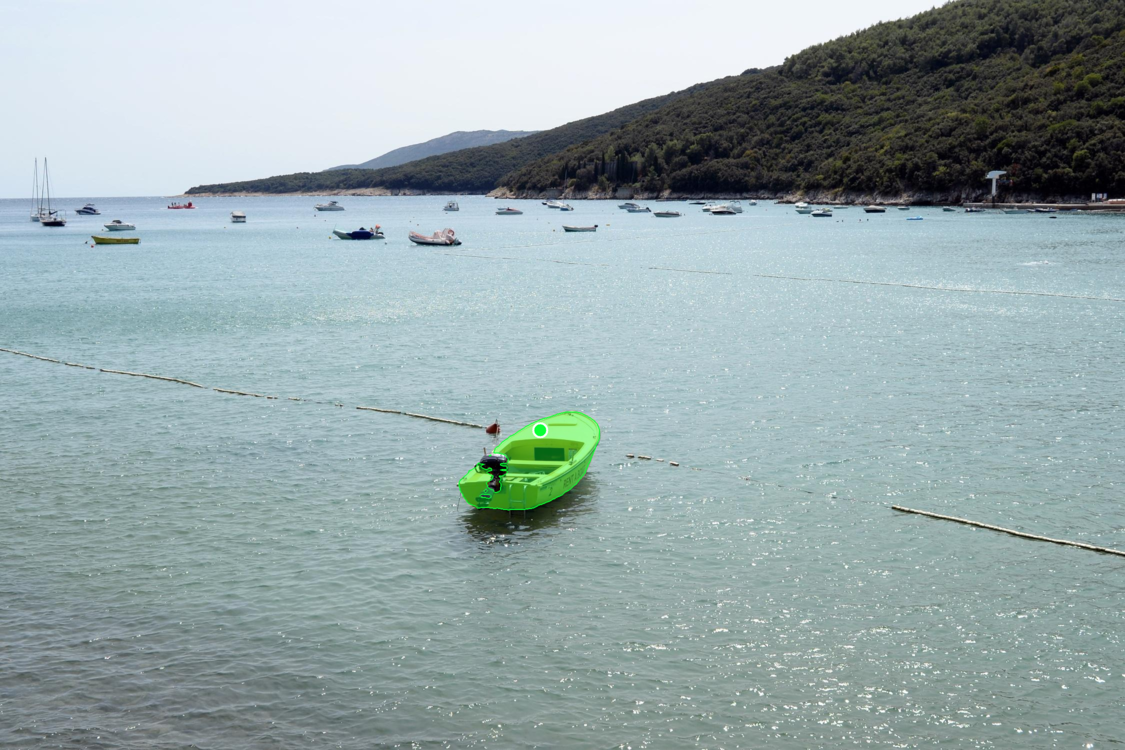} &
        \includegraphics[width=0.18\textwidth]{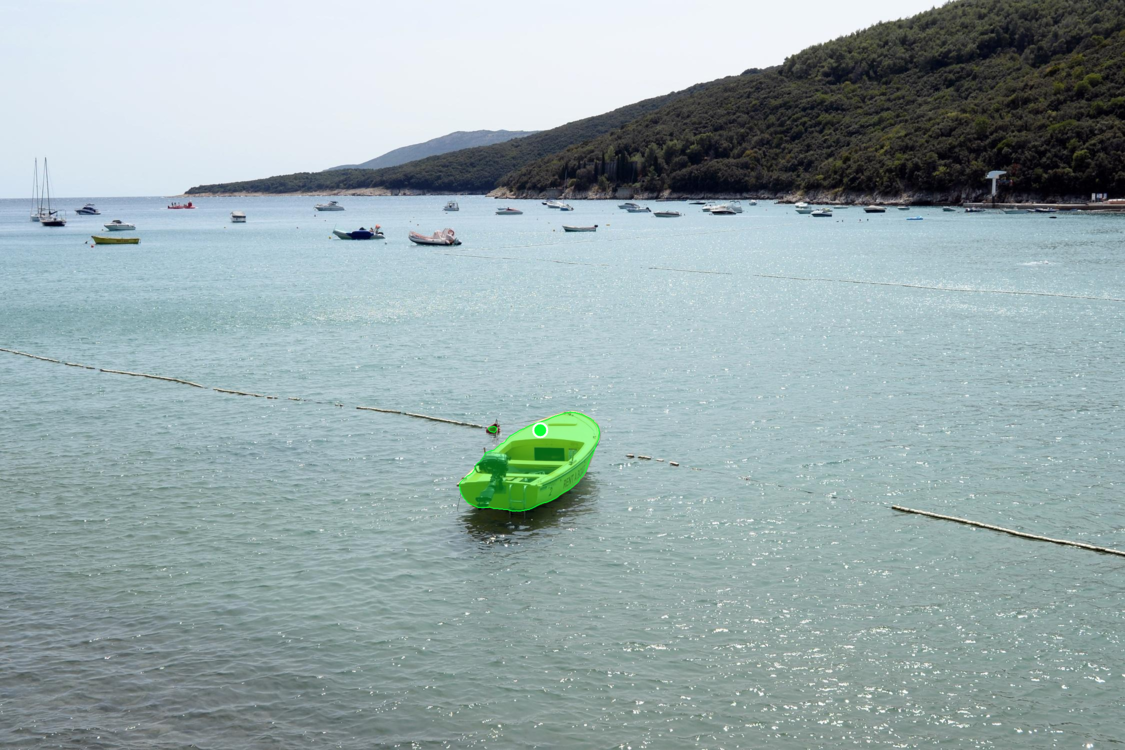} &
        \includegraphics[width=0.18\textwidth]{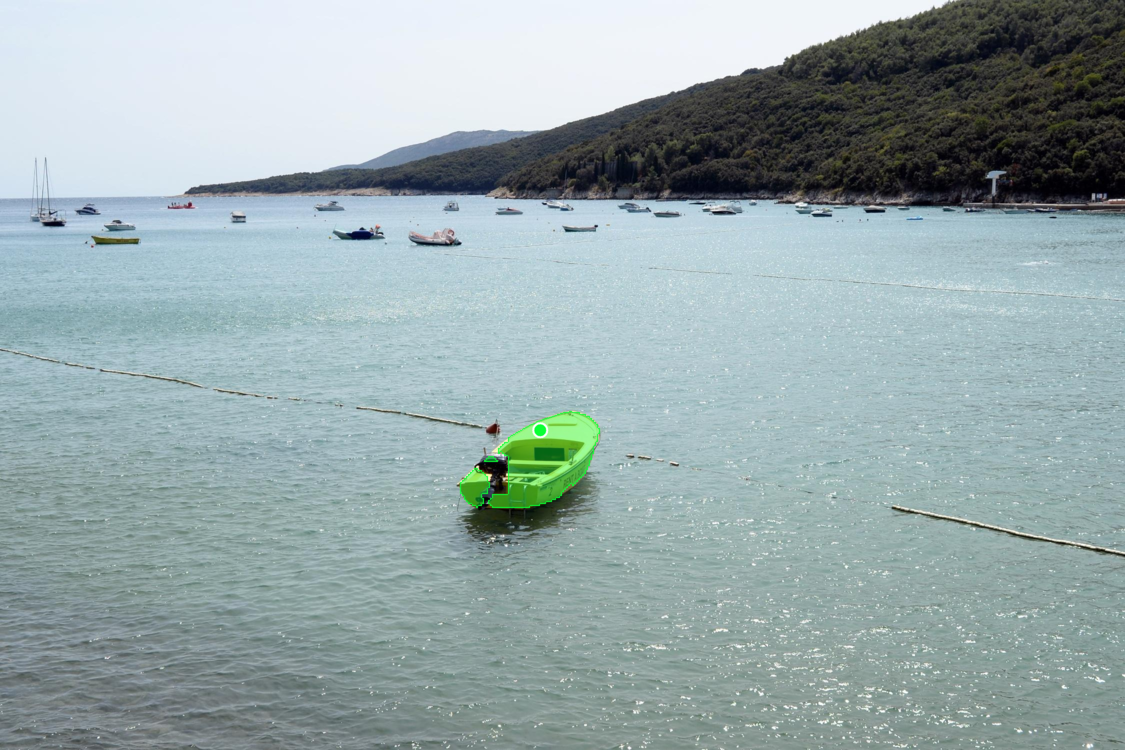} & 
        \includegraphics[width=0.12\textwidth]{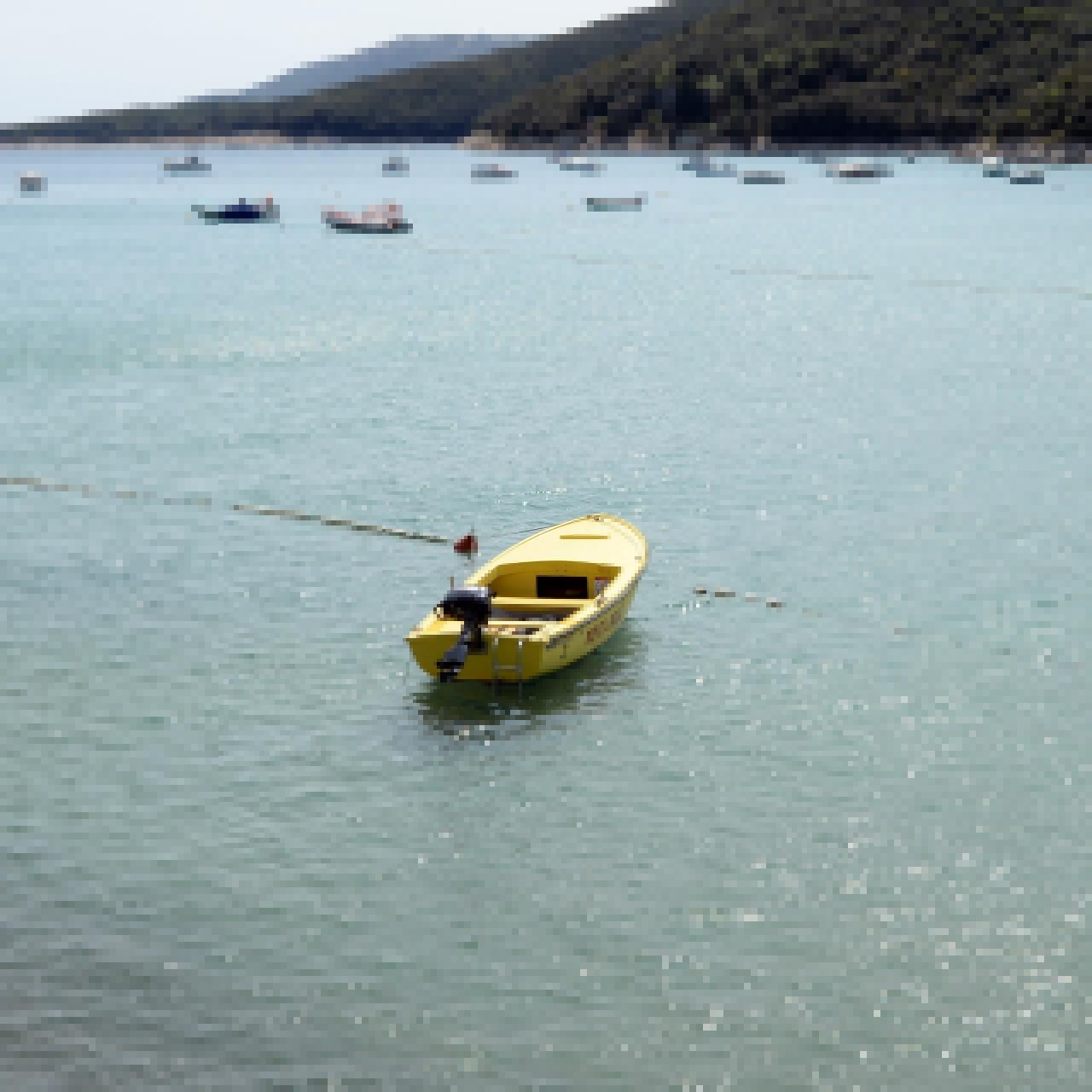} \\
        
        \includegraphics[width=0.18\textwidth]{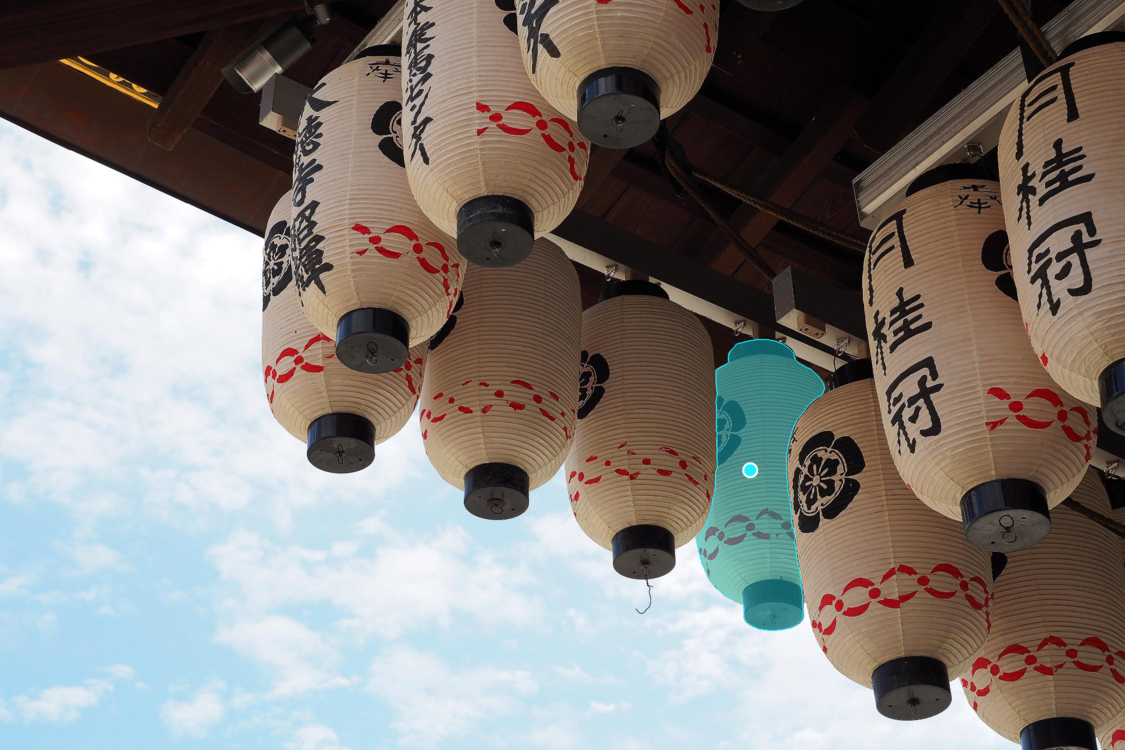} & 
        \includegraphics[width=0.18\textwidth]{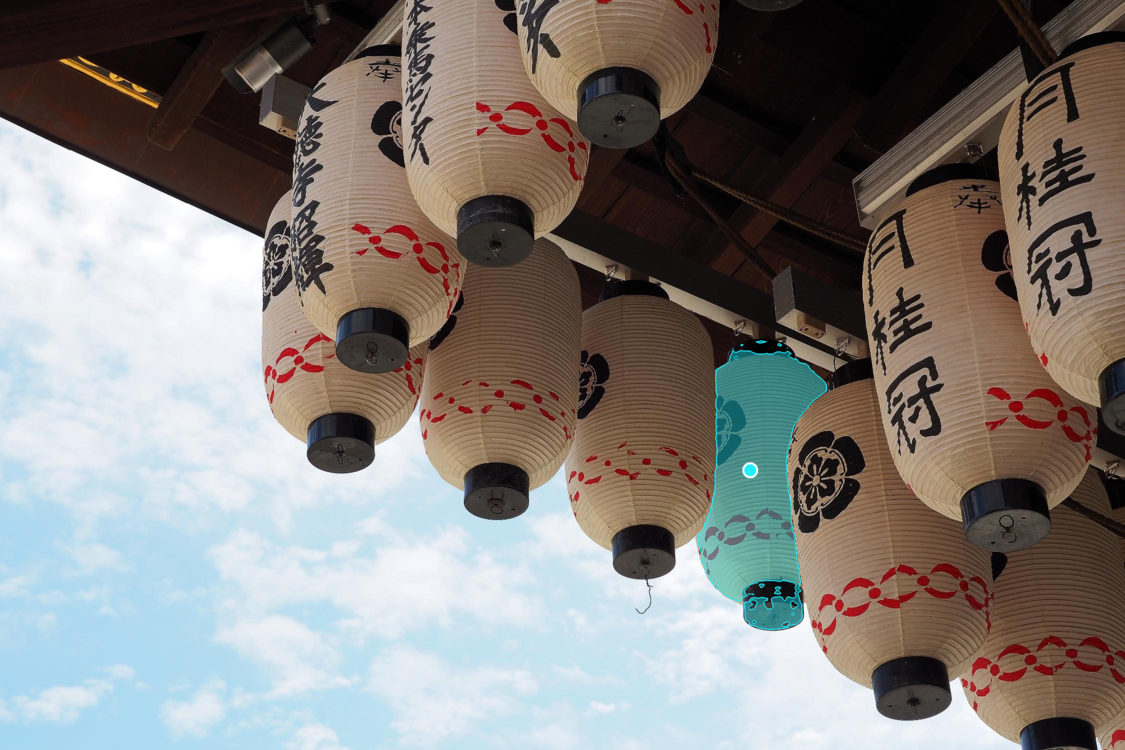} &
        \includegraphics[width=0.18\textwidth]{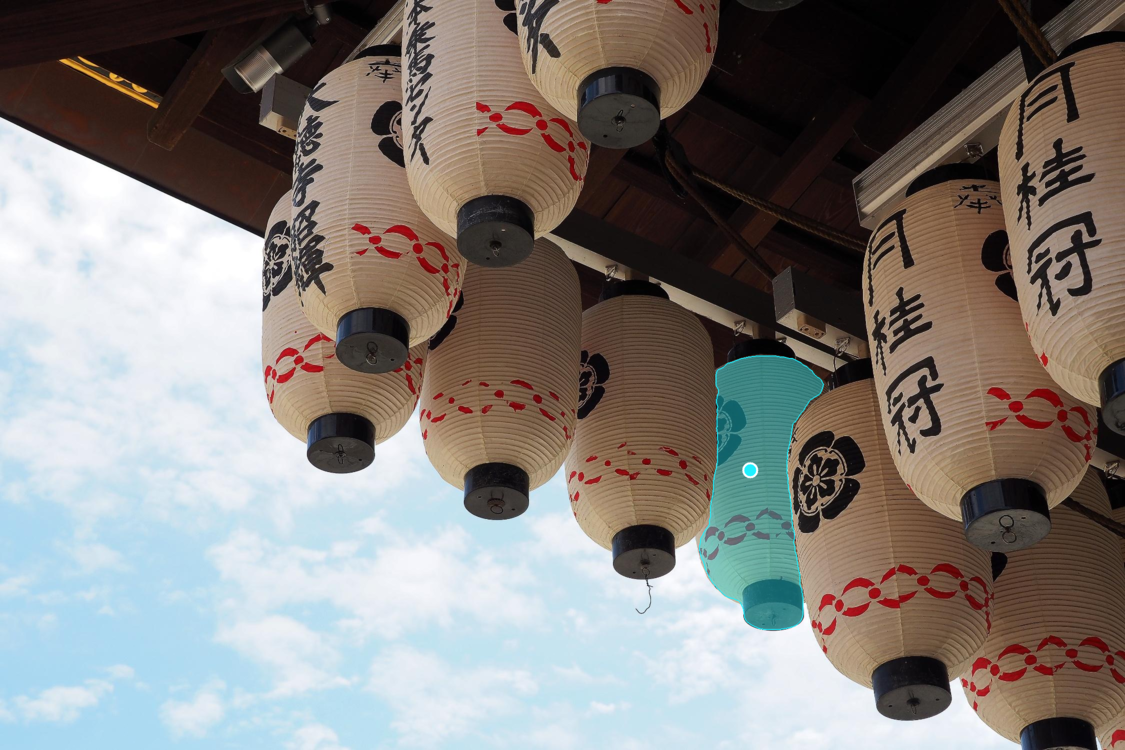} &
        \includegraphics[width=0.18\textwidth]{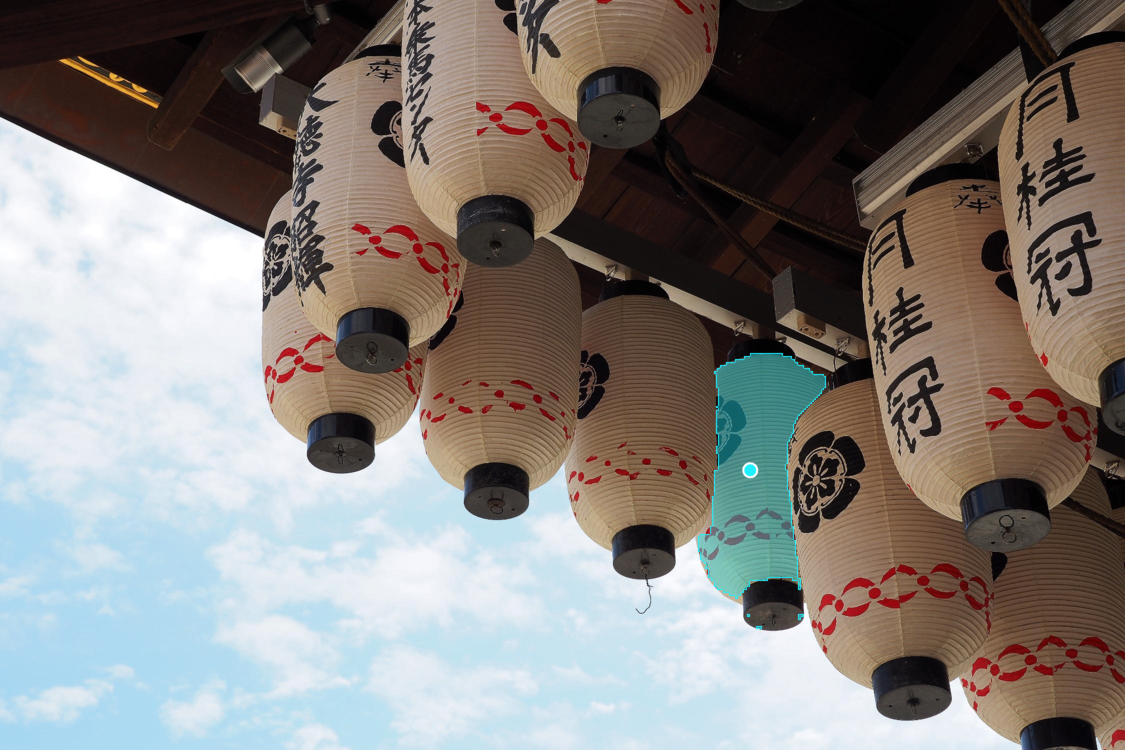} & 
        \includegraphics[width=0.12\textwidth]{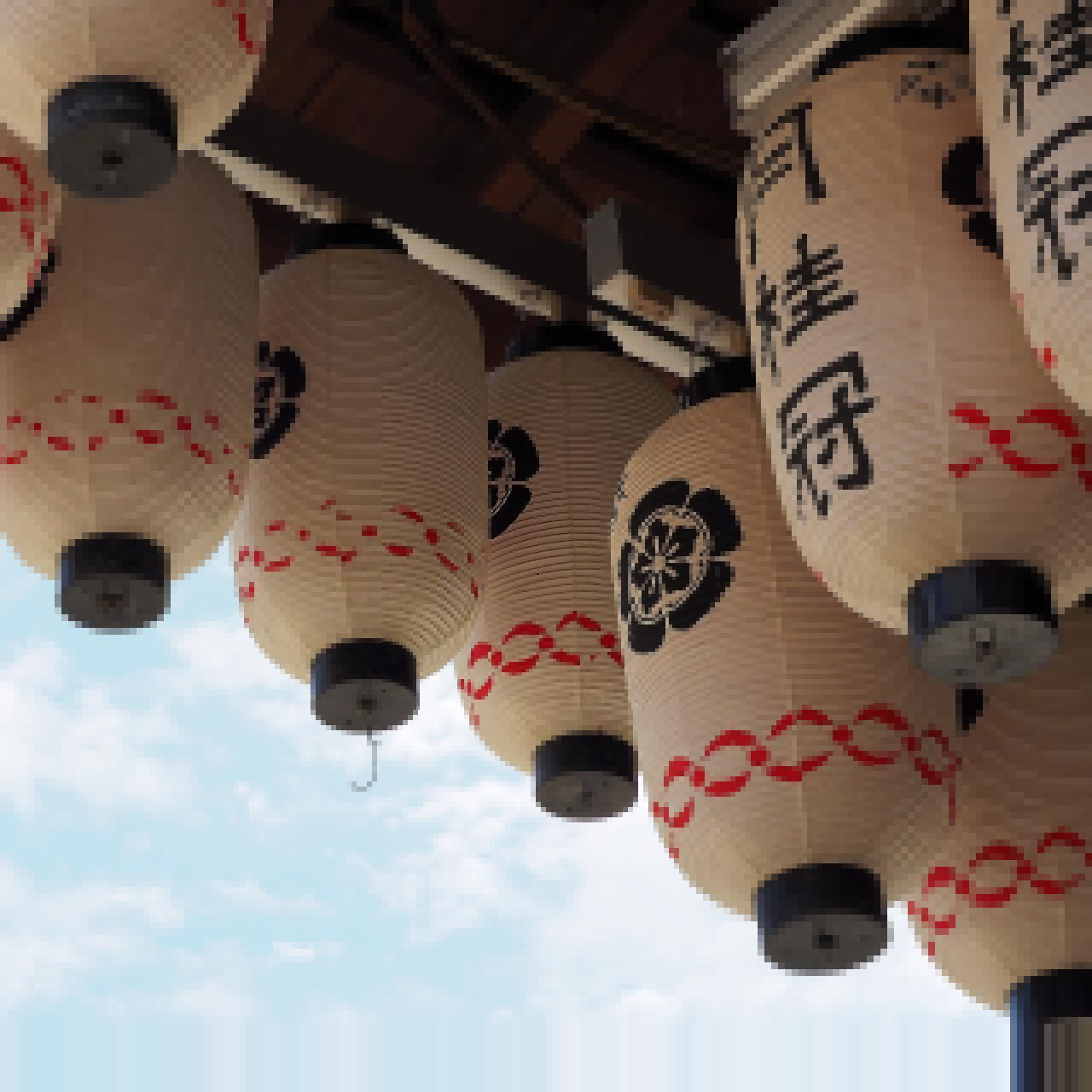} \\

        \includegraphics[width=0.18\textwidth]{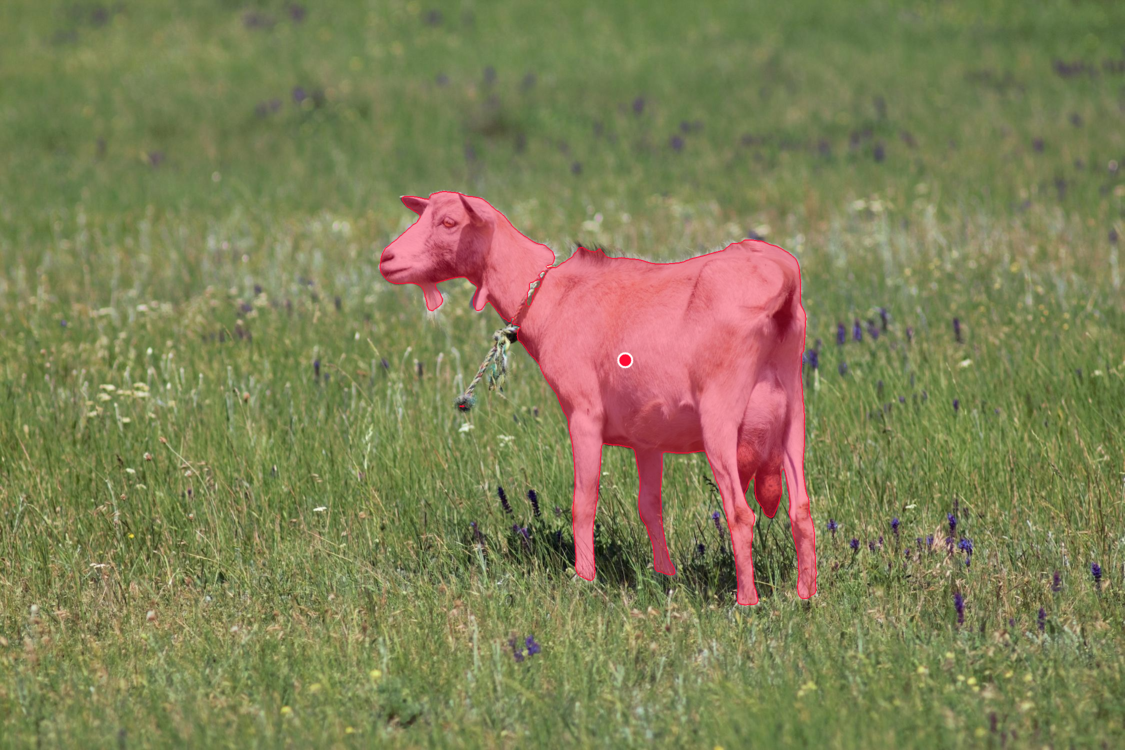} & 
        \includegraphics[width=0.18\textwidth]{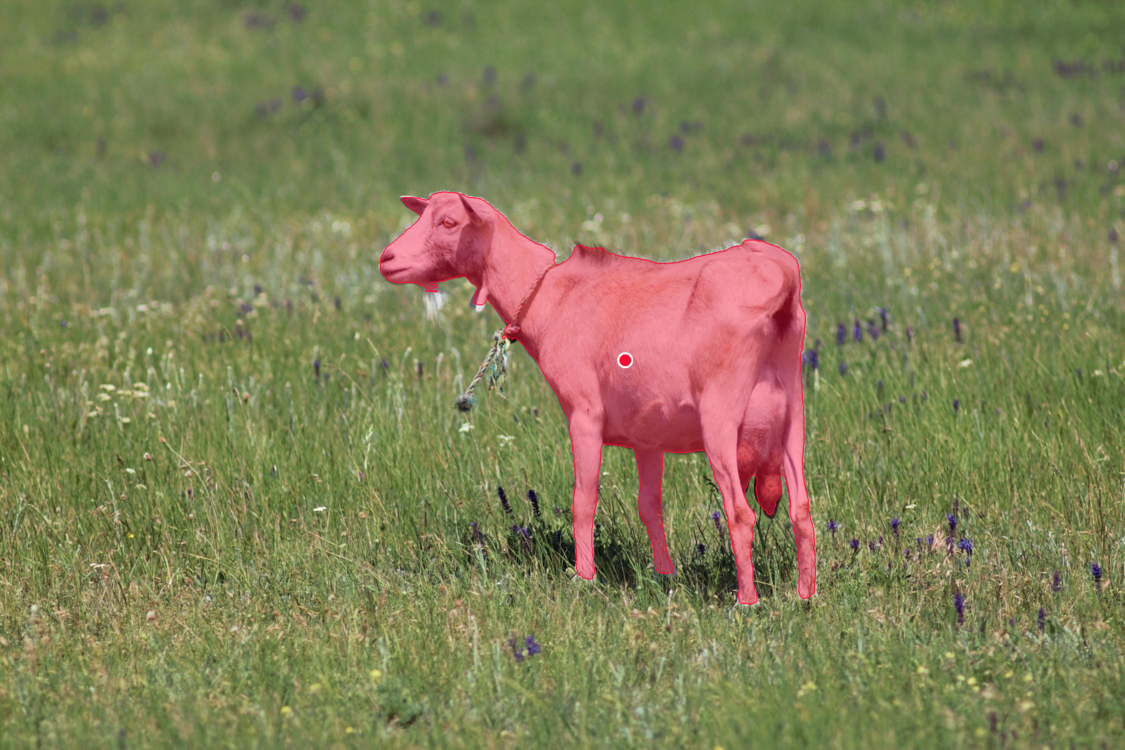} &
        \includegraphics[width=0.18\textwidth]{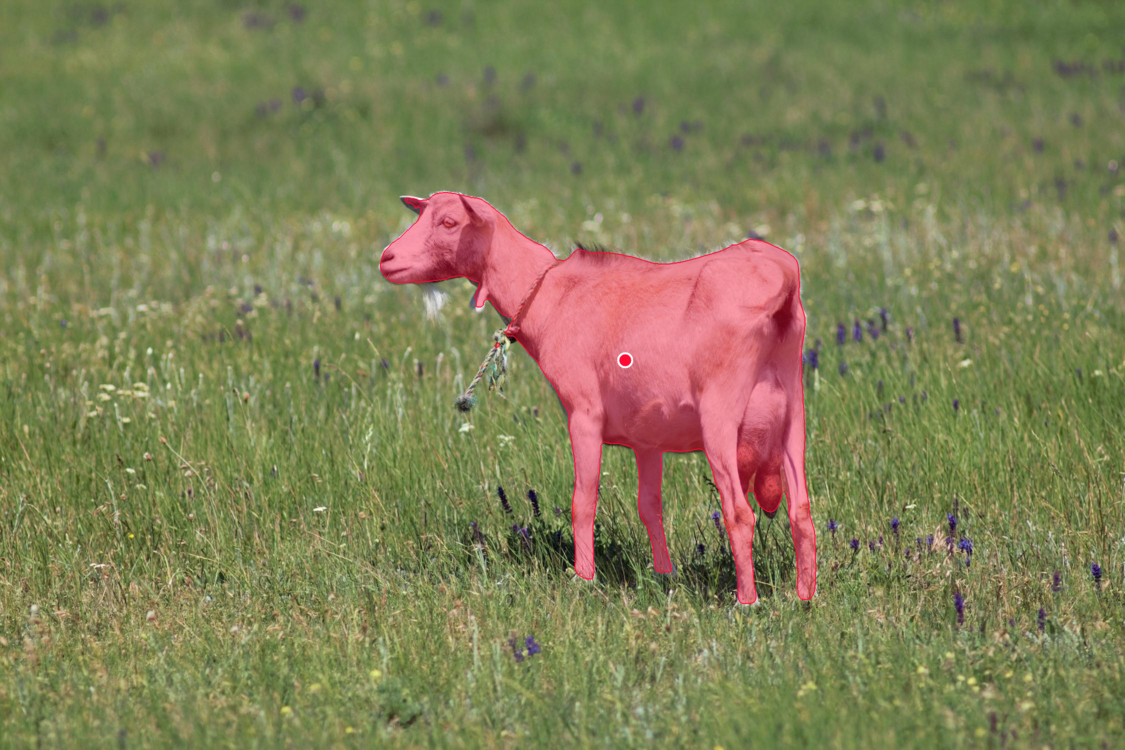} &
        \includegraphics[width=0.18\textwidth]{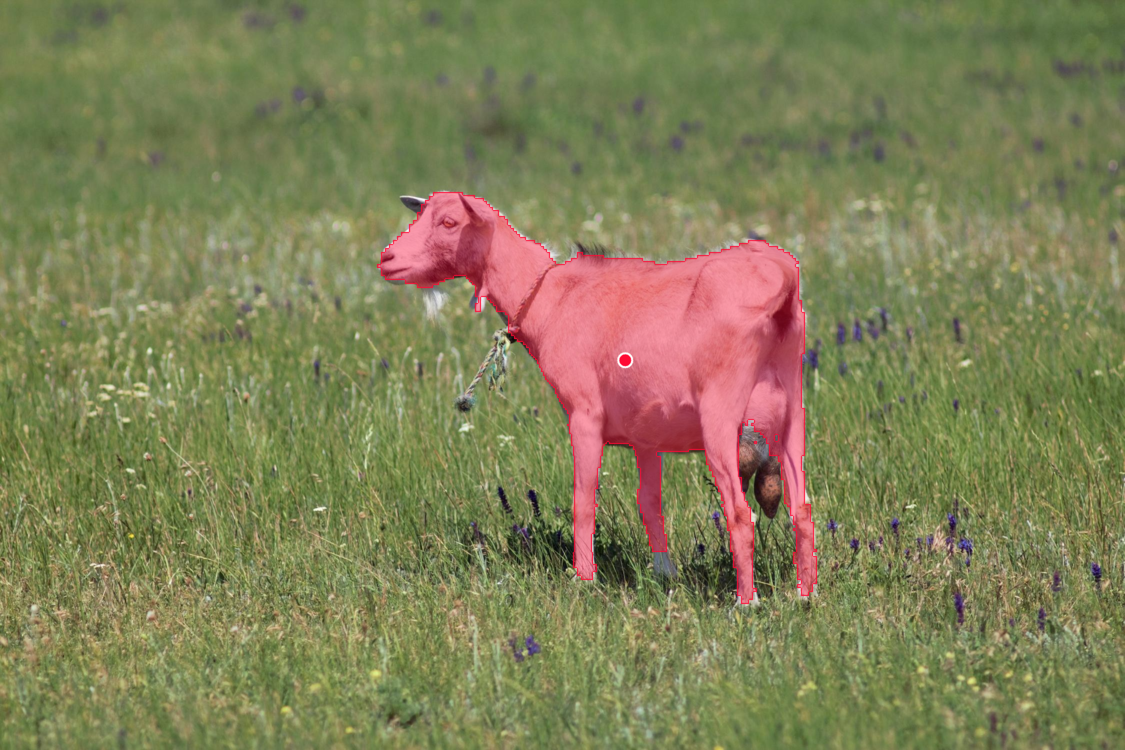} & 
        \includegraphics[width=0.12\textwidth]{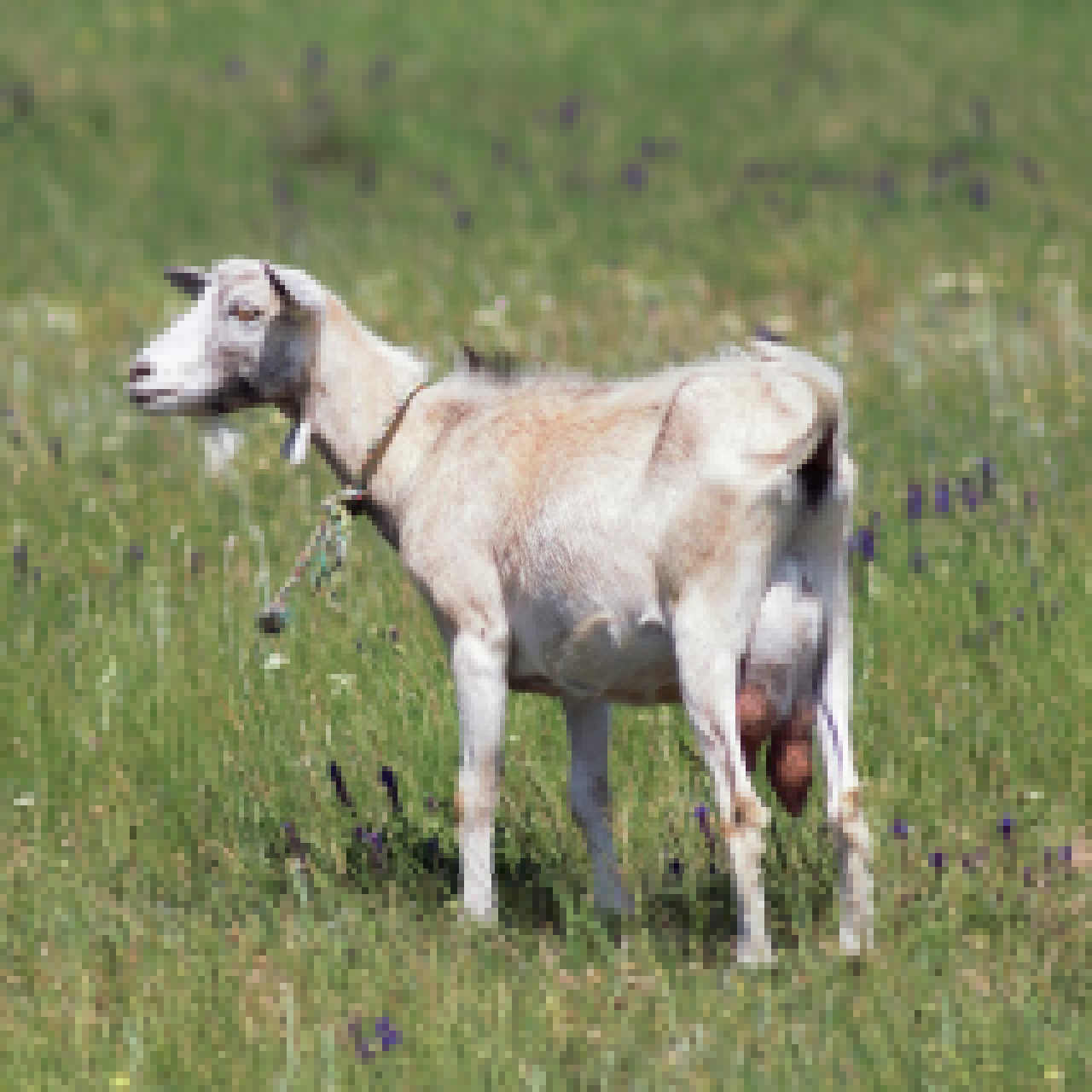}
    \end{tabular}
    \caption{\label{fig:side_by_side} A side-by-side comparison of our Segment This Thing model on a set of images from the SA-1B dataset \cite{Kirillov:etal:CVPR23}. We also visualize the foveated input in the rightmost column as described in Section \ref{sec:visualization}. This figure is best viewed digitally --- zoom in to see the relative coarsity of STT's output for larger segments. For example, the ear of the goat in the third row is missed by STT.} 
\end{figure*}

\subsubsection{Foveation Center Selection}
\label{sec:foveation_centering}
The randomized selection of the foveation center and the resulting variability in the resampling of the image plays much the same role as data augmentation. 
However, rather than altering the data contained in the training set, randomized foveation simply provides many different views of the same data.
While we revisit the same images from SA-1B repeatedly throughout training, the model almost certaintly never receives exactly the same foveated input twice.
We train without drop path as used by Kirillov \etal, but have not observed issues with overfitting.

In some cases, one may wish to prompt a segmentation model with some uncertainty as to whether the prompt is within the intended segment.
In training STT, we can optionally add a small Gaussian-distributed offset vector between the point selected from within a target segment and the center point used for prompting the model.
In this setup, the model is sometimes prompted with points somewhat outside the target segment.
The model then learns to produce not only segments that include the prompt point but also other nearby segments.
The same strategy employed by SAM to handle size ambiguity (i.e. allowing the model to produce multiple masks, and backpropagating only gradients of the loss associated with the best mask) also naturally handles the ambiguity introduced by prompt noise.
If the model is prompted with an image that contains plausible segments both at and near the center, it can simply produce both segments as output.

\subsection{Evaluation}

We evaluate the Segment This Thing model on the ``Zero-Shot Single Point Valid Mask Evaluation'' task defined by Kirillov \etal \cite{Kirillov:etal:CVPR23} in which the model is prompted with a single point in each segment that is furthest from the segment boundary.
We measure accuracy using the mean intersection over union (mIoU) across all segments in each evaluation dataset.

We compare against SAM \cite{Kirillov:etal:CVPR23} and the faster EfficientSAM  \cite{Xiong:etal:CVPR24} and MobileSAM \cite{Zhang:etal:arXiv23} as baselines, all of which involve resizing the image to 1024 x 1024 pixels, encoding a point prompt, and decoding a mask at 256 x 256 pixel resolution.
To evaluate STT, we do cropping and resampling as usual, but map each output mask back to the input image pixel space before evaluating the IoU.
During training, both the height and width of training images are larger than the pattern size, such that crops centered near a corner may need padding on up to two sides.
Some of the evaluation images are significantly smaller, however.
To avoid a domain gap, we upsample images as necessary before cropping such that the crop needs padding on at most two sides.

We analyze the efficiency of our model and the baselines in Table \ref{tab:efficiency}.
Note that STT requires fewer FLOPs and thus achieves lower latency than other efficient variants of SAM while retaining model capacity.

\begin{figure*}
    \centering
    \includegraphics[width=0.28\textwidth]{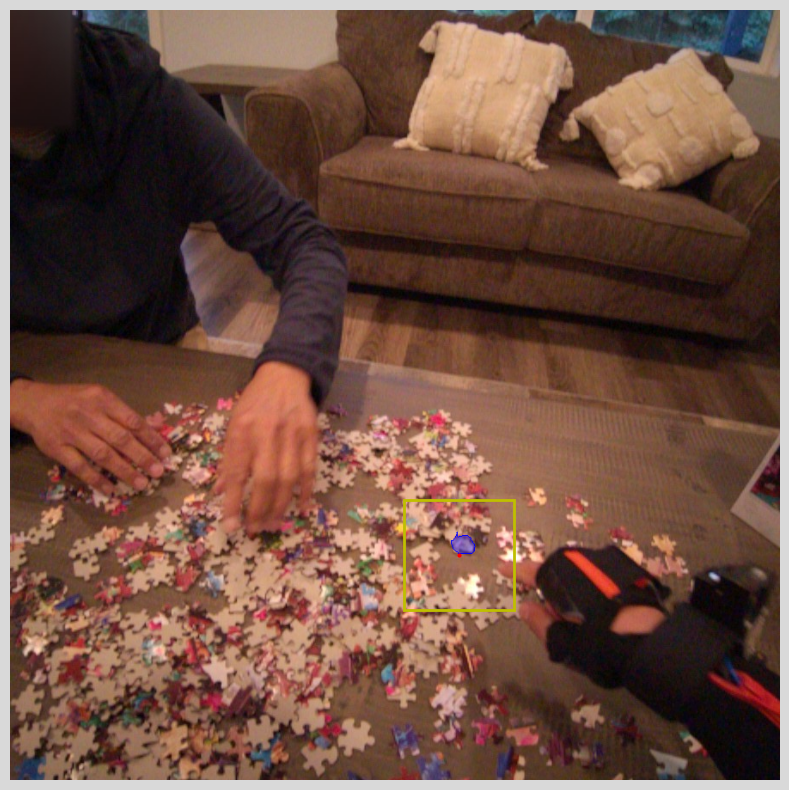}
    \includegraphics[width=0.28\textwidth]{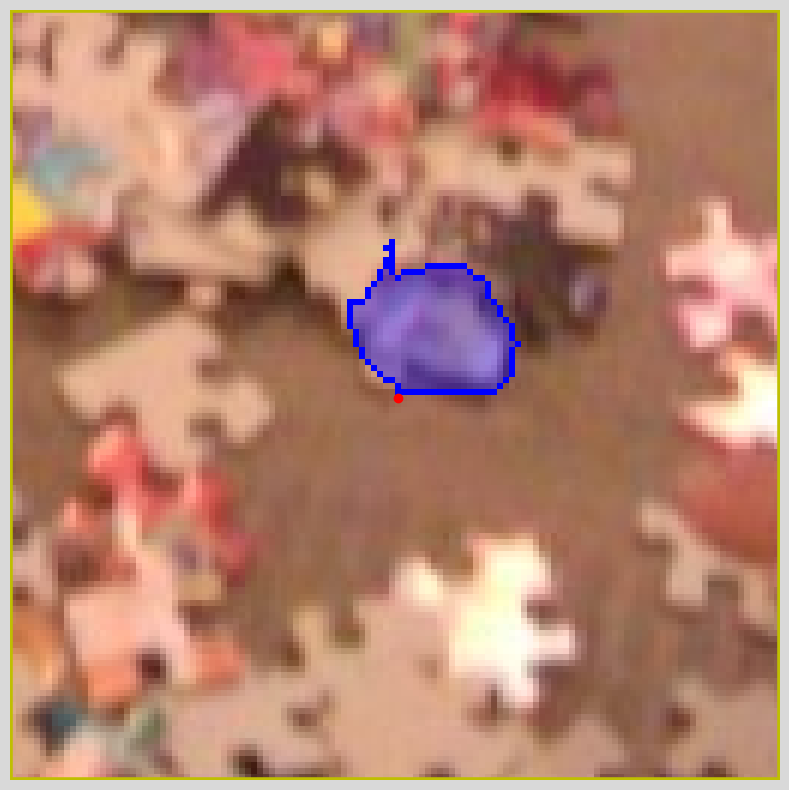}
    \includegraphics[width=0.28\textwidth]{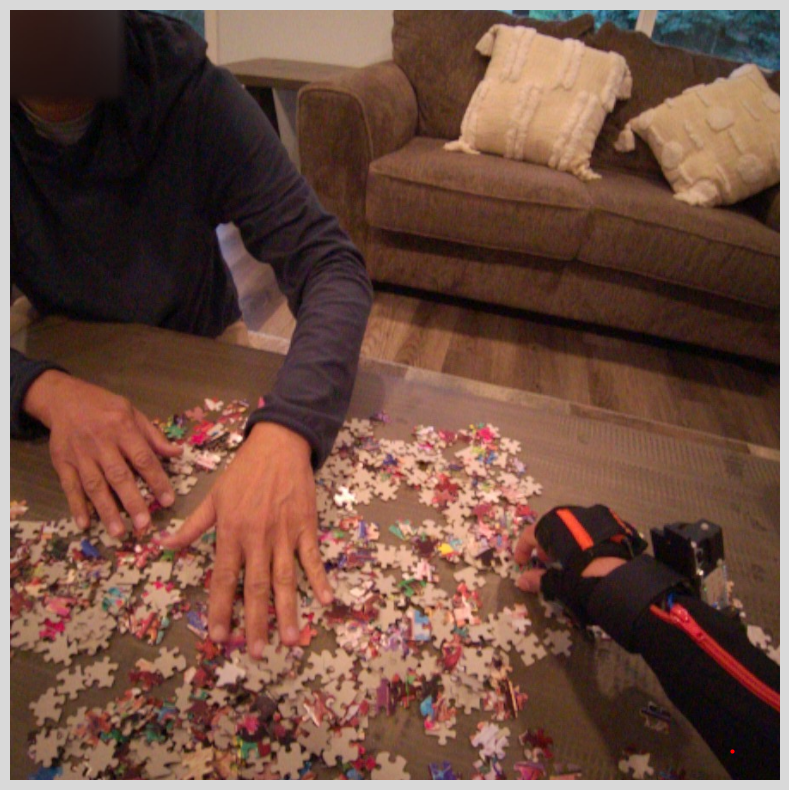}
    \caption{\label{fig:aria} A visualization of prompting Segment This Thing with gaze. In a sequence from the Nymeria dataset \cite{Ma:etal:arXiv24}, a user does a puzzle while wearing Aria glasses \cite{Engel:etal:arXiv23a} equipped with eye-tracking capabilities and a forward-facing camera. The wearer's gaze is inferred and projected into the camera frame, and the resulting point is used to prompt Segment This Thing. A model trained with noise on the foveation center is capable of segmenting objects that the user is looking at even when the gaze is not entirely accurate. \textbf{Left:} the wear is likely looking at a puzzle piece, but the estimate of their gaze projects onto a nearby point on the table (shown in red). \textbf{Center:} in this zoomed-in view, we can see that the most confident segment in STT's estimation is the puzzle piece. \textbf{Right:} a few frames later, the wearer reaches for this exact piece, lending evidence that this indeed had been the object of the wearer's attention.}
\end{figure*}

We use a set of nine datasets to evaluate segmentation accuracy, and plot the results for multiple sizes of each model in Figure \ref{fig:main_results}.
We adopt the standard transformer size denotations used by Vision Transformers \cite{Dosovitskiy:etal:ICLR20}.
Our STT-L model significantly outperforms STT-B, an indication that maintaining model capacity is important to the performance of our model.
However, we find diminishing returns with STT-H.
Our models all significantly outperform MobileSAM, which has similar latency to STT-H.
STT-L is competitive with EfficientSAM-Ti in terms of accuracy,
but despite the larger parameter count is still nearly \textbf{3x} faster.
This is again due to our reduced token count.

A qualitatitve side-by-side comparison of the three models is shown in Figure \ref{fig:side_by_side}. 
Note that some of the performance gap between STT and the more expensive models is due to the relatively coarse segment boundaries caused by foveation (see the supplementary material for further analysis of the performance of STT as a function of segment size).
However, we expect that this will not be critical for many downstream tasks.
If a precise boundary is important, one can always prompt the model again near the boundary once its approximate contours are known.

\subsection{Gaze-based Prompting}

We also show qualitative results in which we prompt the Segment This Thing model with gaze data in Figure \ref{fig:aria} and in the supplementary video.
Specifically, we take sequences from the Nymeria dataset \cite{Ma:etal:arXiv24} in which users wear Aria glasses \cite{Engel:etal:arXiv23a} while performing a variety of activities.
The glasses have forward-facing cameras capturing the wearer's egocentric perspective as well as cameras aimed at the their eyes enabling gaze tracking.
The estimated gaze vectors can then be re-projected into the frame of reference of the forward-facing camera image and used as a prompt for STT.
However, there is some noise in the estimation of the wearer's gaze and therefore some uncertainty in the prompt.
Instead of prompting the model with multiple points drawn from the distribution over the true gaze point, which would incur a cost that scales linearly in the number of prompts, we use a model trained with noise as described in Section \ref{sec:foveation_centering}.
With this model, using a single prompt placed at our best estimate of where the wearer is looking still often results in a segmentation of the object they are actually looking at.
Furthermore, if there is only one obvious segment near the gaze vector, as might be the case when the wearer gazes at an object against a featureless background, this segment will in many cases be the highest scoring segment according to the model's IoU predictions.
The example shown in Figure \ref{fig:aria} is such a case.

%% file: sec_5_conclusion.tex
\section{Conclusion}
\label{sec:conclusion}

In this paper we introduced the Segment This Thing model, a new image segmentation model that builds on the success of SAM by increasing efficiency.
Instead of simply reducing the size of the Transformer-based image encoder, we applied foveated tokenization to reduce the token count by roughly 24x compared to SAM.
This reduces inference latency by roughly 42x compared to SAM-H or 5.6x relative to EfficientSAM.
Foveated tokenization has the additional benefit of reducing the number of pixels required as input to the network, easing up on bandwidth requirements for transmitting image data between sensors and devices; our model requires only about 44K pixels as input (roughly equivalent to a $210 \times 210$ image), but retains the ability to segment small objects at full resolution.

Despite the increased efficiency, Segment This Thing remains competitive on mIoU metrics on multiple public datasets.
By maintaining a relatively large parameter count, our model retains the ability to do more complex computation over the reduced token set compared to smaller models that do simpler computations over more tokens. 
This helps our models cope with the information lost due to foveation.
Furthermore, foveated tokenization concentrates the computational cost where it matters most: near the prompt.

Altogether, the Segment This Thing model is a promising choice for streaming video use cases. 
While we have only evaluated foveated tokenization in the context of image segmentation, we believe the appraoch could be equally applicable to a variety of image and video processing tasks.
We leave this exploration to future work.

%% file: sec_X_suppl.tex
\clearpage
\setcounter{page}{1}
\maketitlesupplementary

\section{Training Details}
\label{sec:supmat_training_details}
\subsection{MAE Pre-training}
We pre-trained our foveated image encoders using MAE pre-training for 500K iterations.
We use an AdamW optimizer with a learning rate of $2^{-13}$ and weight decay of $0.001$.
There is a 10K step linear warm-up of the learning rate, after which it is held constant.
Instead of periodically dropping the learning rate, we double the batch size every 100K iterations.
The initial batch size is 1024 foveations.
For more efficient use of streaming bandwidth, each image is used to generate two foveated views.
We use the standard masking ratio of 0.75 and do not apply loss to the patches provided to the MAE as input.
Fixations are sampled uniformly with a margin of 256 pixels.

\subsection{Segmentation Training}
We trained our full STT models for an additional 250K iterations.
We again use AdamW with a weight decay of $0.001$, this time with a learning rate of $2^{-16}$ and a 5K step linear warm-up.
The batch size starts at 2048 and doubles after every 50K steps.
We sample up to 16 foveateds per image, and again use a sampling margin of 256 pixels.
We use loss weights of 20.0 for the focal loss, 1.0 for the dice loss, and 0.01 for the IoU prediction loss.

\section{Foveation Pattern}
\label{sec:supmat_foveation_pattern}

As described in Section \ref{sec:foveated_tokenization}, our foveation patterns consist of a series of nested rings of patches with a dense grid in the center.
Here we give a more formal definition of the parameterization of such a pattern.

A pattern with $N$ layers must specify a stride $s_i \in \mathcal{Z+}$ and a grid size $g_i \in \mathcal{Z+}$ for each level $i \in [0, N)$.
Generally, $s_0 = 1$ such that the dense grid in the center is full resolution.
We require $s_i > s_{i-1}$ for all $i$, i.e. the layers are defined in order of increasing stride.
Layer $i$ then defines a $g_i \times g_i$ grid of patches with a bounding size of $g_i s_i p$, where $p$ is the patch size (in our experiments $p = 16$ pixels).
In order to ensure nesting, we further require $g_i s_i > g_{i-1} s_{i-1}$ (the grids get larger from layer to layer), and $$s_i \mid \frac{g_i s_i - g_{i-1}s_{i-1}}{2},$$
i.e. the difference in sizes between successive grids is an even multiple of the stride of the higher level such that the lower level grid can be centered and surrounded by patches at the next level, leaving no gaps or overlap.

Successive grids redundantly cover some of the same pixels.
The patch strides are not constrained to be integer multiples of each other, so retaining redundant patches would increase the number of pixels required to represent the image and increase bandwidth requirements.
We therefore drop redundant tokens for efficiency, at each level introducing only those patches that cover regions of the image that are not already covered by lower levels.

\begin{figure}
    \centering
    \footnotesize
    \def\svgwidth{0.99\columnwidth}
    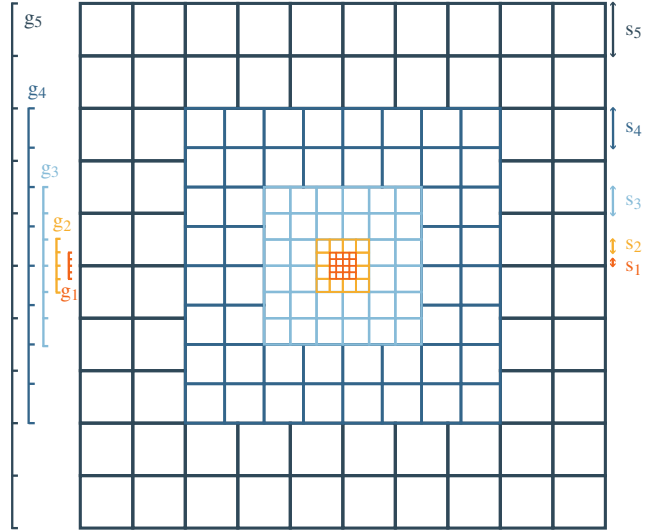
    \caption{\label{fig:foveation_pattern} The foveated tokenization pattern as used in our experiments with the Segment This Thing model. Each square in the image indicates the size and location of the receptive field of a patch. The patch sizes are all integer multiples of the smallest patch size, such that every patch can be downsampled to the same size using a simple box filter. Patches are colored by level with grid sizes indicated at left and strides at right.}
\end{figure}

\begin{table}[h!]
    \centering
    \begin{tabular}{c | c c}
    \textbf{Level} ($i$) & \textbf{Stride} ($s_i$) & \textbf{Grid size} ($g_i$) \\
    \midrule
    1 & 1 & 4 \\
    2 & 2 & 4 \\
    3 & 4 & 6 \\
    4 & 6 & 8 \\
    5 & 8 & 10
    \end{tabular}
    \caption{\label{tab:foveation_definition}The precise definition of our foveation pattern. The interpretation of the parameters is given in Section \ref{sec:supmat_foveation_pattern}. The pattern is depicted in Figure \ref{fig:foveation_pattern}.}
\end{table}

The total foveation pattern size in pixels is $g_{N-1} s_{N-1} p$ pixels.
The total token count can be computed as:
\begin{equation}
    \sum_{i=0}^{N-1} g_i^2 - \sum_{i=1}^{N-1} \frac{s_{i-1}g_{i-1}}{s_i}^2
\end{equation}
where the first term adds all the grid sizes and the second accounts for the removal of redundant tokens.
We trained our model with a five-layer foveation pattern, with parameters listed in Table \ref{tab:foveation_definition}.
The pattern and the geometric interpretation of its parameters are visualizated in Figure \ref{fig:foveation_pattern}.

\begin{figure}
\centering
\includegraphics[width=\columnwidth]{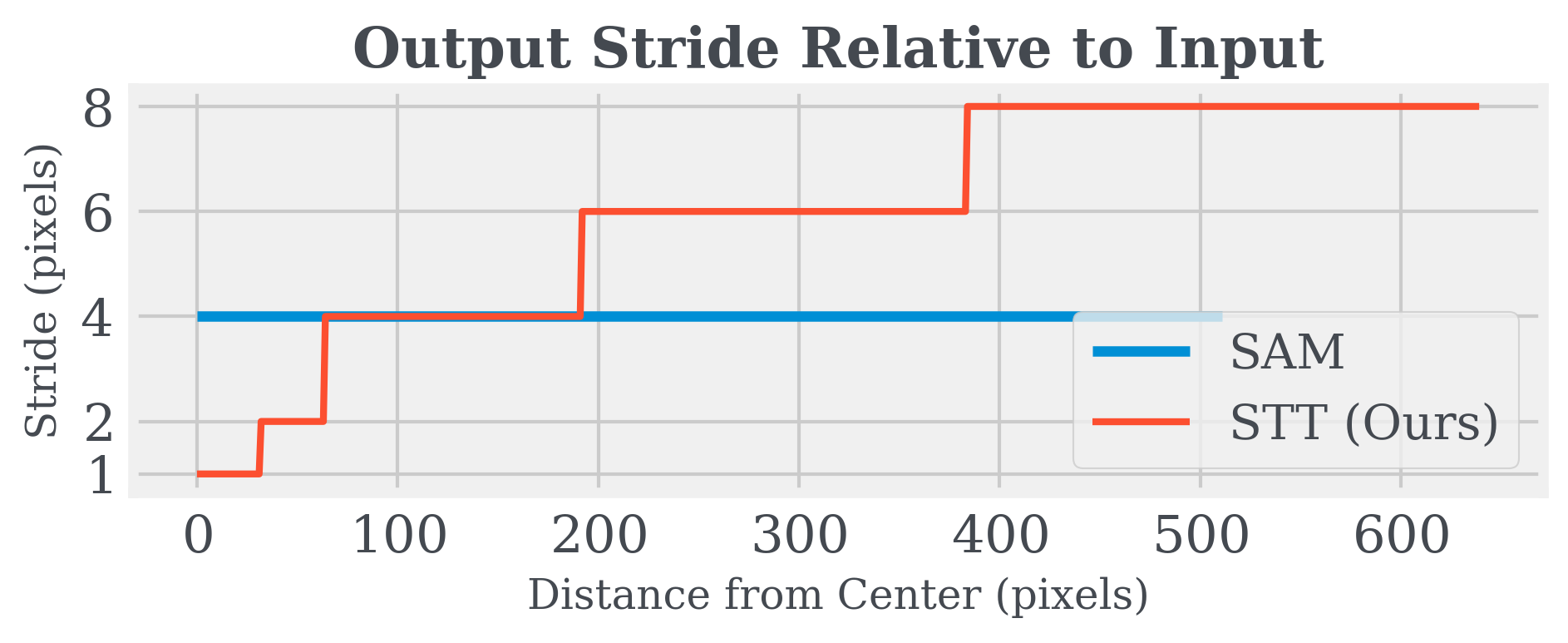}
\caption{\label{fig:stride_vs_distance} The stride of the segmentation maps produced by STT and SAM relative to the input size. SAM outputs segmentation maps at a flat resolution that is one quarter of the input resolution.}
\end{figure}

In Figure \ref{fig:stride_vs_distance} we show the input and output stride of the pattern as a function of horizontal or vertical distance from the center.
We also plot the output stride of SAM, indicating which regions of our output segmentation maps have lower, higher, or equivalent resolution.

\section{MAE Pre-training}
\begin{figure}[hb!]
    \centering
    \includegraphics[width=0.925\columnwidth]{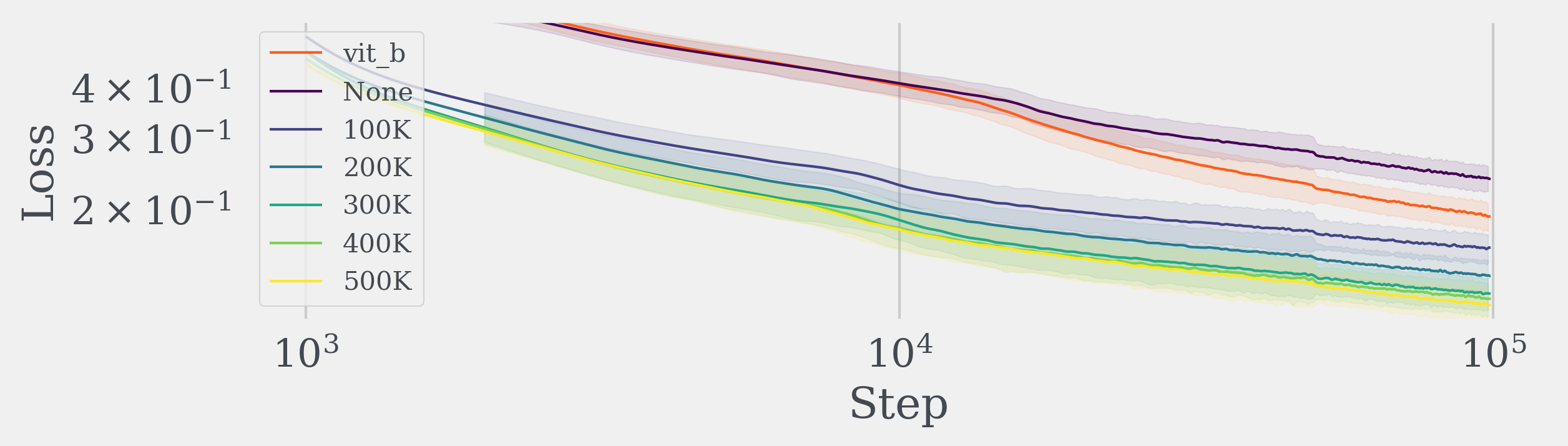}
    \caption{Training curves for STT-B models as a function of initialization, plotted on a log/log scale.}
    \label{fig:mae_plot}
\end{figure}

To evaluate the effectiveness of MAE pretraining on foveated tokenizations, we trained STT-B models for 100K iterations with various initial weights: random initialization, a series of MAE checkpoints, and pre-trained publicly available ViT weights. We plot the training loss curves in Figure \ref{fig:mae_plot}. We see that longer pre-training results in gains that persist through fine-tuning, with eventual diminishing returns. The ViT weights were trained with standard patch tokenization on the regular grid and is initially worse than random initialization. The network is eventually able to repurpose these weights, but after 100K steps of training even a small amount of foveated MAE pre-training yields better results.

\section{Varying the Token Count}
We also ran a small experiment to evaluate the effect of the foveation pattern on segmentation accuracy.
Our foveation patterns exists in a high-dimensional design space, and each new pattern requires its own MAE pre-training.
We thus focused on the token count, performing a somewhat abbreviated training run (300K steps pre-training, 200K steps training) for one pattern with fewer tokens than our standard model and one pattern with more.
The patch size and overall receptive field were kept fixed, and all models are size L.
The segmentation accuracy is presented in Table \ref{tab:pattern_ablation}.
As expected, performance increases with increased token count.

\begin{table}
    \centering
    \begin{tabular}{c | c c c}
    Token Count & Cityscapes & EgoHOS & VISOR \\
    \toprule
    100 (-42\%) & 0.375 & 0.568 & 0.544 \\
    \textbf{172} & 0.400 & 0.597 & 0.571 \\
    268 (+56\%) & 0.405 & 0.612 & 0.586 \\
    \end{tabular}
    \caption{\label{tab:pattern_ablation}Segmentation accuracy of models with more or less tokens than the baseline 172-token model. Note that results for the baseline differ from table \ref{tab:full_results} due to the abbreviated training schedule. }
\end{table}

\section{Full Evaluation Results}
We list the full evaluation results in tabular form in Table \ref{tab:full_results}.

\begin{table*}
  \centering
  \small
  \begin{tabular}{c c| c c c c c c c c c}
  & & \multicolumn{9}{c}{Accuracy (mIoU)} \\
  \multicolumn{2}{c|}{\textbf{Method}} & \textbf{ADE20K} & \textbf{Cityscapes} & \textbf{EgoHOS} & \textbf{NDD20} & \textbf{PPDLS} & \textbf{TimberSeg} & \textbf{VISOR} & \textbf{ZeroWaste} & \textbf{WoodScape} \\
  \toprule
  \multirow{3}{*}{SAM} & H & 0.543 & 0.393 & 0.582 & 0.826 & 0.762 & 0.674 & 0.604 & 0.629 & 0.301 \\
  & L & 0.537 & 0.392 & 0.601 & 0.817 & 0.764 & 0.644 & 0.606 & 0.634 & 0.296 \\
  & B & 0.547 & 0.384 & 0.620 & 0.798 & 0.771 & 0.524 & 0.599 & 0.607 & 0.300 \\
\midrule
\multirow{1}{*}{MobileSAM} &   & 0.471 & 0.302 & 0.557 & 0.733 & 0.640 & 0.338 & 0.549 & 0.579 & 0.234 \\
\midrule
\multirow{2}{*}{EfficientSAM} & S & 0.553 & 0.405 & 0.632 & 0.771 & 0.678 & 0.507 & 0.618 & 0.628 & 0.323 \\
  & Ti & 0.544 & 0.378 & 0.615 & 0.786 & 0.747 & 0.473 & 0.572 & 0.598 & 0.320 \\
\midrule
\multirow{3}{*}{STT (Ours)} & H & 0.552 & 0.410 & 0.620 & 0.754 & 0.730 & 0.434 & 0.596 & 0.620 & 0.308 \\
  & L & 0.553 & 0.412 & 0.607 & 0.719 & 0.758 & 0.421 & 0.582 & 0.595 & 0.310 \\
  & B & 0.541 & 0.393 & 0.596 & 0.732 & 0.735 & 0.398 & 0.571 & 0.583 & 0.291 \\

  \bottomrule
  \end{tabular}
  \caption{\label{tab:full_results} A full listing of all segmentation accuracy results.}
  
\end{table*}

\section{STT Performance Analysis}
\label{sec:supmat_analysis}

In this section we describe two investigations into the performance of STT relative to the baselines to deepen the understanding of the model.

\subsection{Breakdown by Distance from Prompt}
\begin{figure*}
    \centering
    \begin{tabular}{c c c c}
    & \textbf{Cityscapes} & \textbf{EgoHOS} & \textbf{VISOR} \\
    \rotatebox[origin=c]{90}{\textbf{Precision}} &
    \includegraphics[width=0.29\textwidth, valign=c]{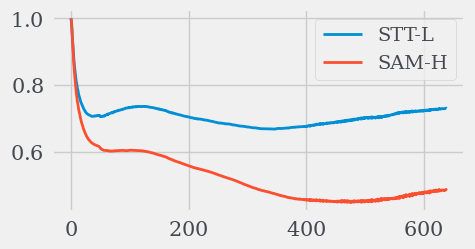} &
    \includegraphics[width=0.29\textwidth, valign=c]{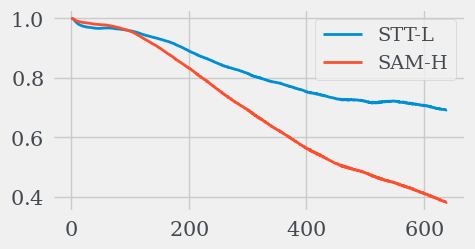} &
    \includegraphics[width=0.29\textwidth, valign=c]{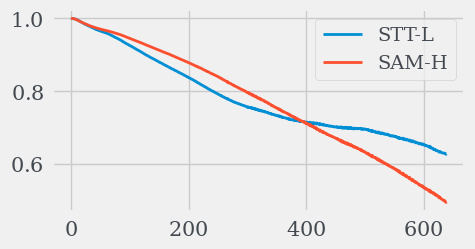} \\
    \rotatebox[origin=c]{90}{\textbf{Recall}} &
    \includegraphics[width=0.29\textwidth, valign=c]{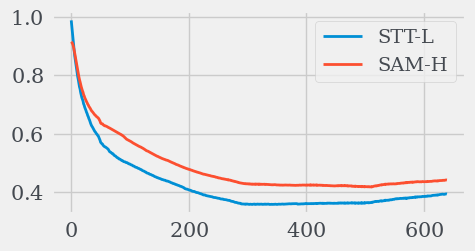} &
    \includegraphics[width=0.29\textwidth, valign=c]{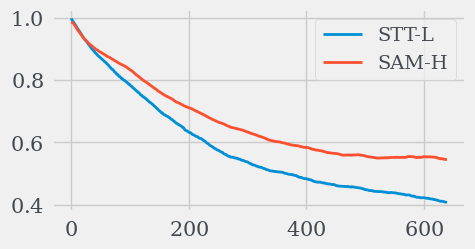} &
    \includegraphics[width=0.29\textwidth, valign=c]{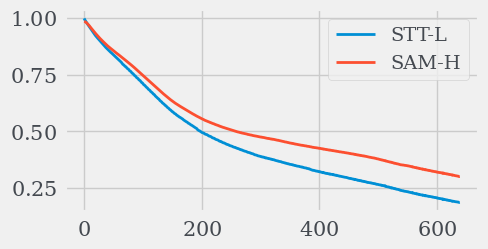} \\    \rotatebox[origin=c]{90}{\textbf{Accuracy}} &
    \includegraphics[width=0.29\textwidth, valign=c]{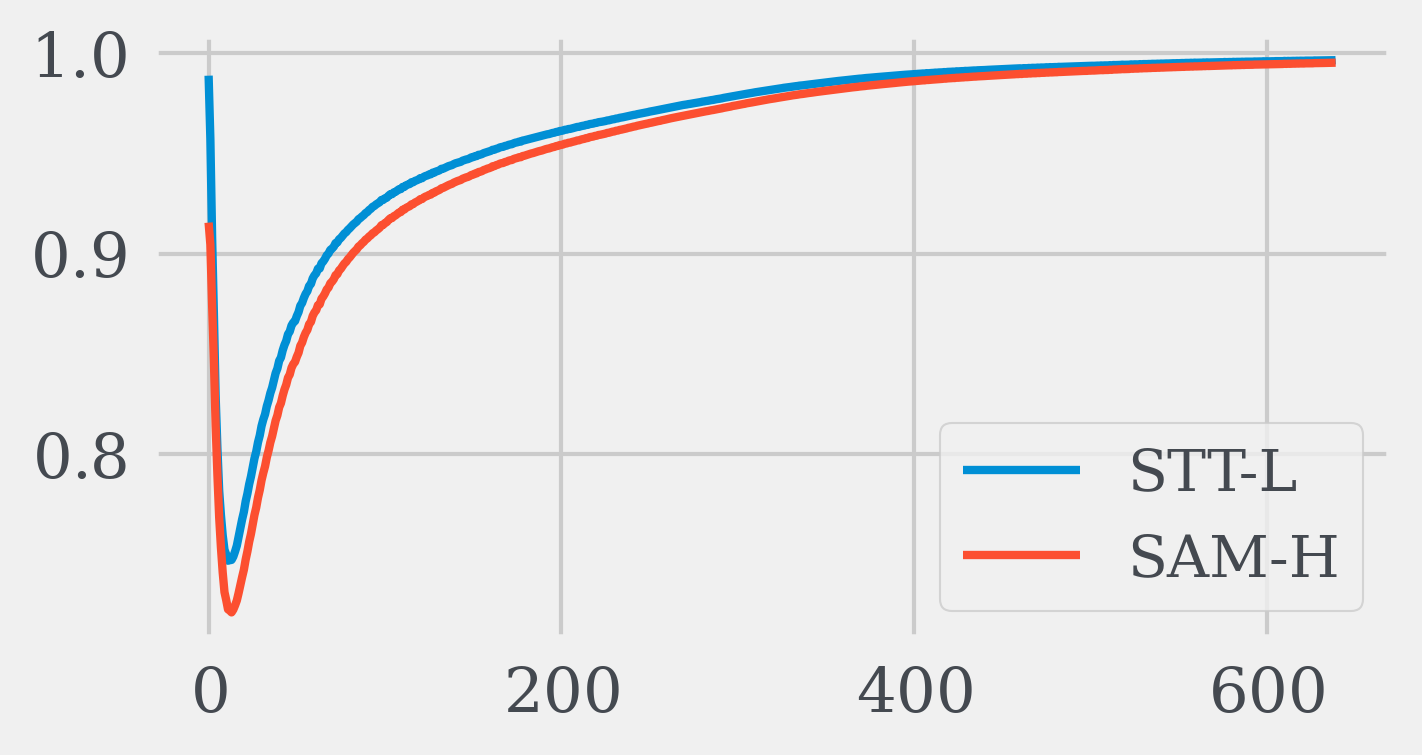} &
    \includegraphics[width=0.29\textwidth, valign=c]{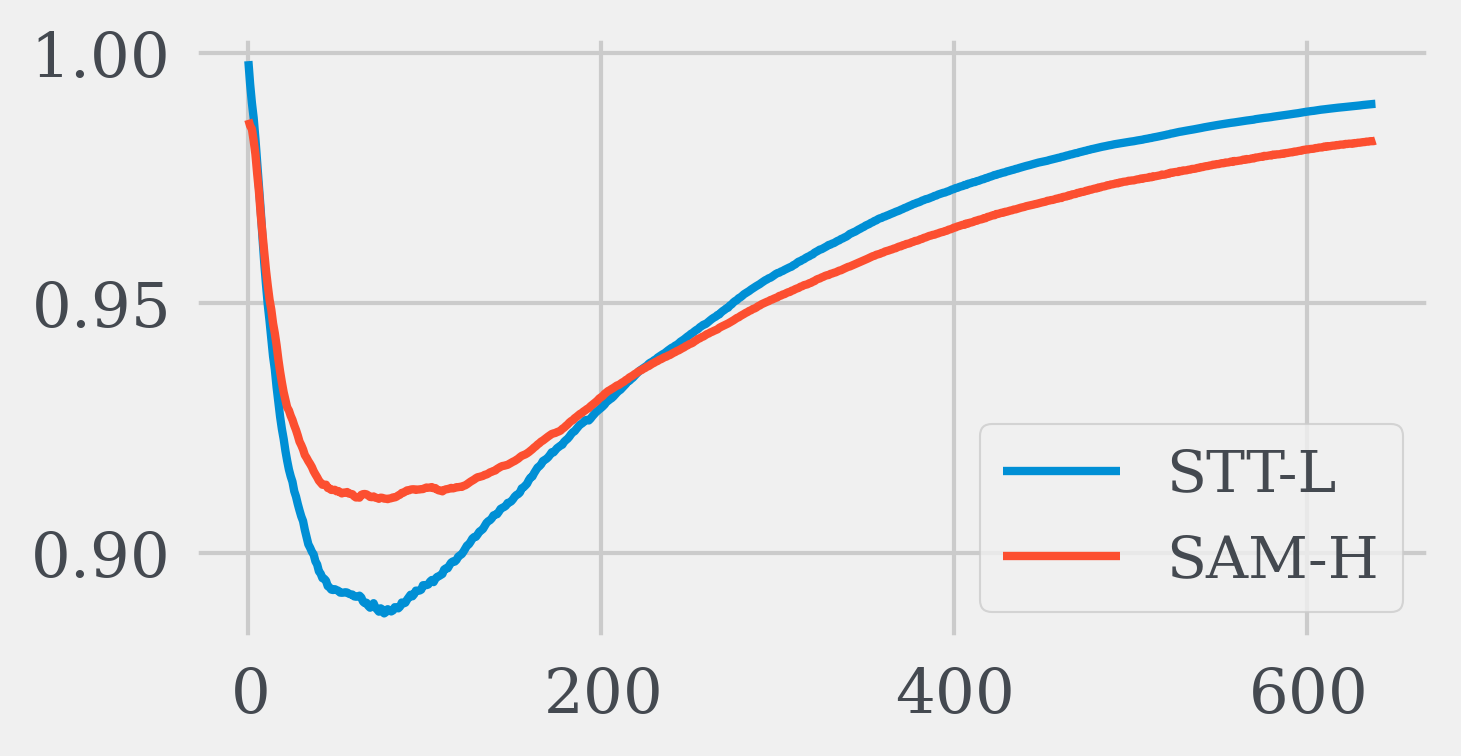} &
    \includegraphics[width=0.29\textwidth, valign=c]{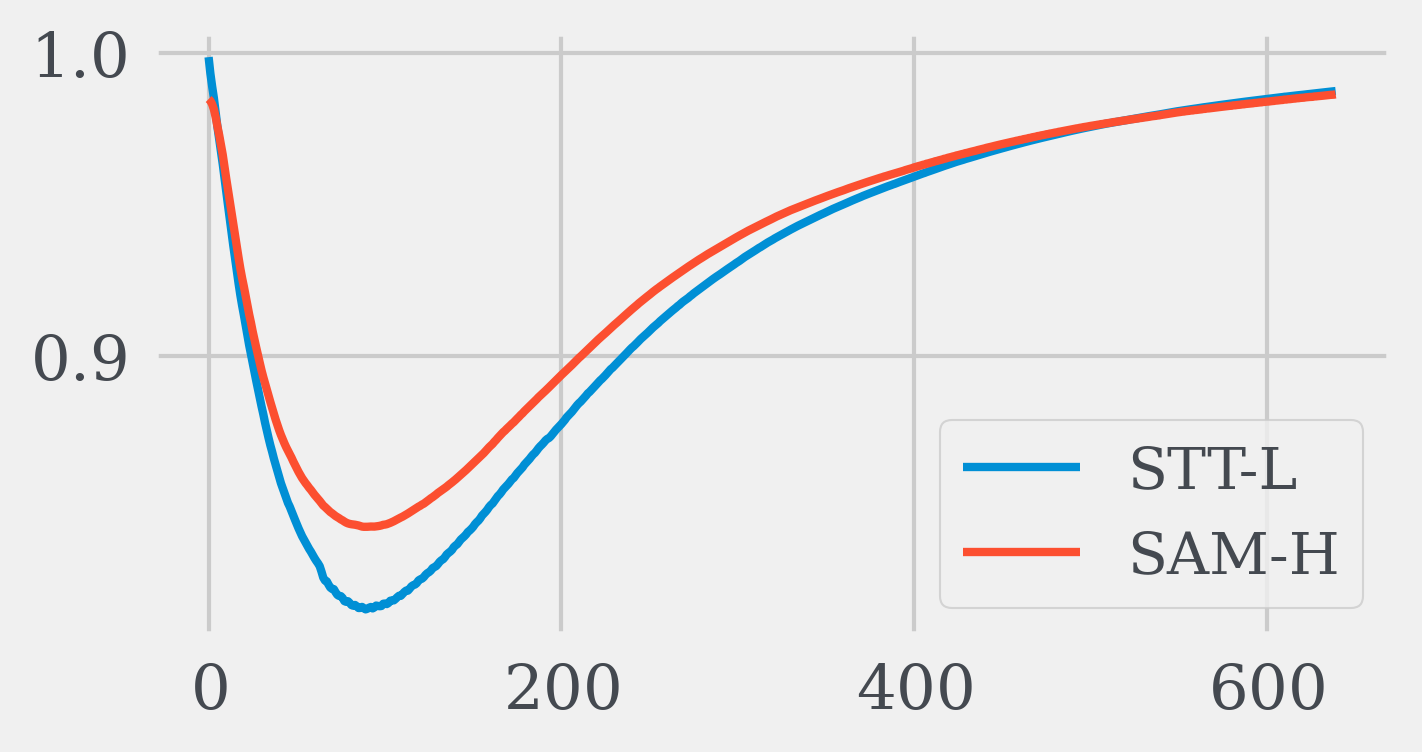} \\
    \rotatebox[origin=c]{90}{\textbf{Pos. Rate}} &
    \includegraphics[width=0.29\textwidth, valign=c]{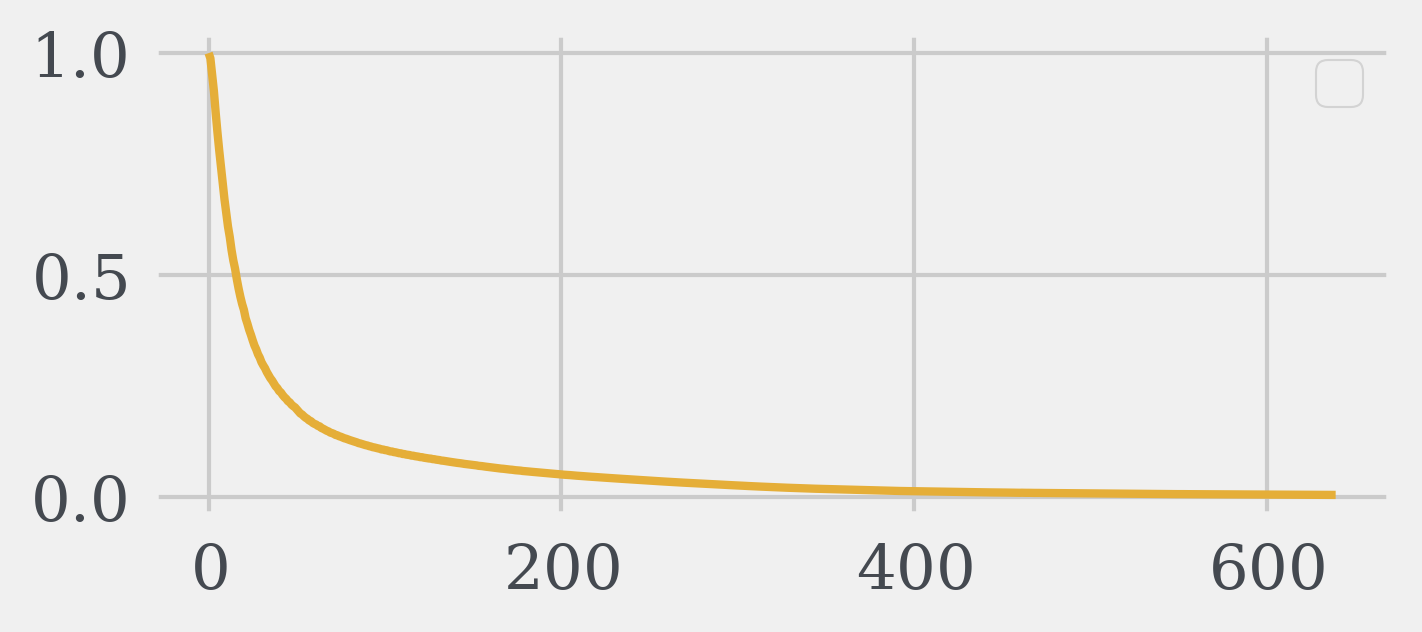} &
    \includegraphics[width=0.29\textwidth, valign=c]{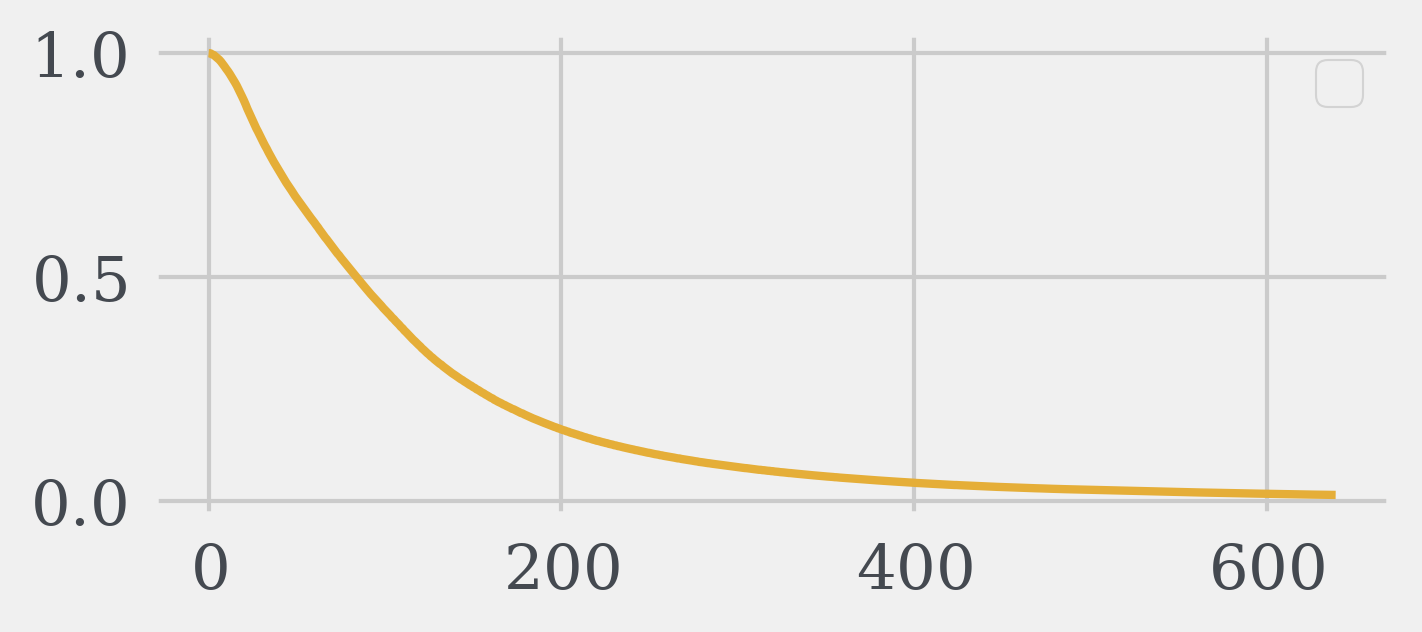} &
    \includegraphics[width=0.29\textwidth, valign=c]{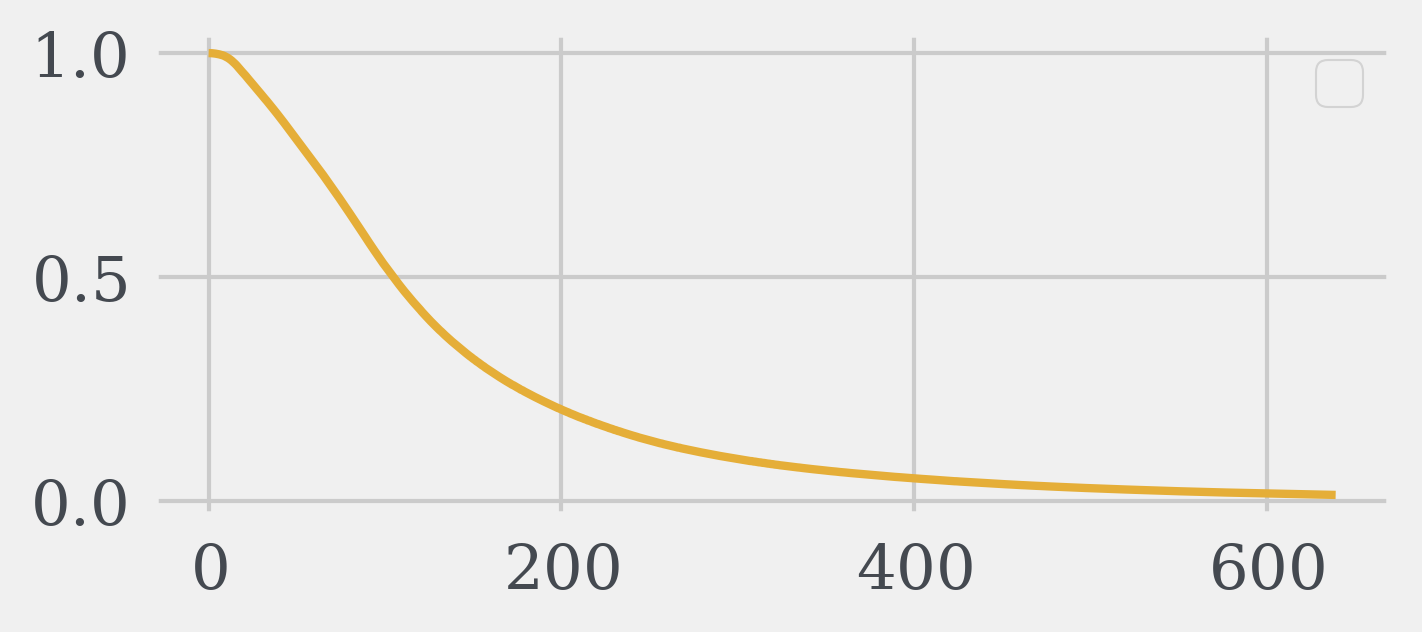} \\
    \end{tabular}
    \caption{\label{fig:pr_by_distance} The top three rows show the pixel-wise precision, recall, and accuracy of SAM and STT, respectively, as a function of the distance of a pixel from the prompt. The last row shows the fraction of pixels with a positive label as a function of distance from the prompt.}
\end{figure*}

To gain further insight into the relative performance of SAM and STT we measured the pixel-wise precision, recall, and overall accuracy as a function of distance from the prompt on a subset of three evaluation datasets.
The results are plotted in Figure \ref{fig:pr_by_distance}.
Both precision and recall decrease for both models with increased distance as expected.
The particular shape of the curve varies significantly by dataset but some trends hold across all three.

The first trend to note is the distinctive "swoosh" shape of the accuracy plots --- accuracy is high in the vicinity of the prompt, drops to its lowest value for both models with a somewhat increased distance, then begins to asymptotically approach 1.0.
The cause can be clearly seen in the plot of the positive label rate by distance.
The nadir of each accuracy curve reliably occurs at the point the positive rate crosses 0.5. 
We note that even in the datasets where SAM is generally more accurate than STT, there is at least a narrow region centered on the prompt with a radius of 4-8 pixels in which STT is more accurate.
This is perhaps due to the higher resolution in this region as shown in Figure \ref{fig:stride_vs_distance} and described in Section \ref{sec:mask_decoder}.

Finally, we note an interesting trend in the precision and recall curves.
As distance increases, STT tends towards higher precision and lower recall compared to SAM.
This could be a function of the inductive bias in the network caused by the centering of the input on the prompt.
STT apparently estimates positive labels for fewer distant pixels, sometimes missing parts of the segment which results in lower recall.
On the other hand, when it does estimate positive labels for pixels far from the prompt, it is correct more often than SAM.

\subsection{Alternative Evaluation Modes}
When evaluating SAM and EfficientSAM, we followed the standard protocol.
Both methods accept the full frame as input and then rescale it to $1024 \times 1024$.
However, STT requires a $1280 \times 1280$ crop centered on the prompt, and we do not rescale the input.
If the ground truth segment extends beyond the crop boundaries, STT therefore simply pays the penalty for failing to include those pixels.

This does mean that STT and the baselines receive input images with differing receptive fields.
To determine the impact of this, we gave EfficientSAM the same $1280 \times 1280$ crops used by STT as input and re-evaluated both methods only on this region (ignoring any labels outside the crop).
The results are given in Table \ref{tab:eval_mode}.
We note that both methods see an increase in performance, due to the restricted evaluation domain and, in the case of EfficientSAM, a potential increase in the input resolution after rescaling.

We further evaluate EfficientSAM on \textit{foveated} versions of the same $1280 \times 1280$ crop.
This is done by passing the crop through the foveation process and then restructuring them into an image as done for visualization (c.f. Section \ref{sec:visualization}).
Evaluating in this mode, both methods receive exactly the same input and EfficientSAM also benefits from the same reduced bandwidth requirements as STT.
However, as can be seen in Table \ref{tab:eval_mode}, the performance of EfficientSAM drops significantly when operating on foveated inputs, indicating that STT can process such reduced-bandwidth data both more efficiently and more effectively.

\begin{table}
    \centering
    \begin{tabular}{c | c c | c c c }
    \multicolumn{1}{c}{} & \multicolumn{2}{c}{\textbf{Eval Mode}} & \multicolumn{3}{c}{\textbf{Accuracy (mIoU)}} \\
    \textbf{Model} & \rotatebox{70}{Cropped} & \rotatebox{70}{Foveated} & \rotatebox{70}{Cityscapes} & \rotatebox{70}{EgoHOS} & \rotatebox{70}{VISOR} \\
    \midrule
    \midrule
    \multirow{2}{*}{EfficientSAM} & \checkmark & & 0.444 & 0.640 & 0.619 \\
     & \checkmark & \checkmark & 0.410 & 0.560 & 0.540 \\
     \midrule
    STT (Ours) & \checkmark & \checkmark & 0.417 & 0.621 & 0.590 \\
    \end{tabular}
    \caption{\label{tab:eval_mode}Evaluating EfficientSAM (size S) and STT (size L) in different evaluation modes.}
\end{table}

\section{Computing FLOPs}
\label{sec:supmat_flops}

We follow Kaplan \etal in computing FLOP counts for transformer architectures (c.f. \cite{Kaplan:etal:arXiv20}, Table 1).
Specifically, we omit non-linearities, biases, normalizations, and other such operations with negligible contributions relative to the FLOPS counts incurred by the remaining operations.
However, Kaplan \etal focus on language models and give counts per token.
We are interested instead in the cost per image.
The expressions we use to calculate FLOPS are given below.
Note that these expressions yield FLOPS counts for a single layer or application of the indicated function, and thus omit the multiplier by layer count given in \cite{Kaplan:etal:arXiv20}.
We also include expressions for linear and convolutional layers as these are used to compute the FLOPS in the mask decoder.

\begin{table}[h!]
    \centering
    \begin{tabular}{ l | c }
    \toprule
    \textbf{Operation} & \textbf{FLOPS} \\
    \midrule
    Attention: QKV & $2 d_\text{model} d_\text{attn} (n_\text{query} + 2 n_\text{key})$ \\
    Attention: QK Logits & $2 n_\text{query} n_\text{key} d_\text{attn}$ \\
    Attention: Softmax & $3 n_\text{query} n_\text{key} n_\text{head}$ \\
    Attention: Reduce V & $2 n_\text{query} n_\text{key} d_\text{attn}$ \\
    Attention: Project V & $2 n_\text{query} d_\text{model} d_\text{attn}$ \\
    Feedforward & $4 n_\text{query} d_\text{model} d_\text{ff}$ \\
    Linear & $2 n_{vals} d_\text{input} d_\text{output}$ \\
    Convolution & $2 d_\text{input} d_\text{output} w_\text{out} h_\text{out} w_\text{kernel} h_\text{kernel}$ \\
    \bottomrule
    \end{tabular}
\end{table}

Here $d_\text{model}$ is the hidden dimension of the transformer model $d_\text{attn}$ is the total dimension over which attention is computed, i.e. the sum of the dimension of each head.
These values are typically the same but need not be.
$d_\text{ff}$ is the inverse bottleneck dimension, which is typically set to $4 d_\text{model}$.
Note that we differentiate between the number of keys and the number of values to include cross-attention, as used for example in the two-way transformer in the mask decoder of all models considered.
In the case of self-attention, $n_\text{key} = n_\text{query}$.

We also have to treat the windowed attention layers in the SAM image encoder specially.
Given a function $f(d_\text{model}, n_\text{tokens}, n_\text{heads})$ that returns the FLOPS count for a transformer encoder layer (with the default settings of $d_\text{attn}$ and $d_\text{ff}$), the global attention layers in the SAM ViT encoder require about $f(d_\text{model}, s^2, n_\text{heads})$ FLOPs for a token map of size $s \times s$. 
Given a window size $w$, the local windowed attention layers require about $\lceil \frac{s}{w} \rceil^2 f(d_\text{model}, w^s, n_\text{heads})$ FLOPS.
For mask decoders, we include the cost of the MLPs, the deconvolutions, and the dot product used to compute the logits and omit all other terms.

%% file: figs_pattern_fig.pdf_tex
%% Creator: Inkscape 1.4 (e7c3feb100, 2024-10-09), www.inkscape.org
%% PDF/EPS/PS + LaTeX output extension by Johan Engelen, 2010
%% Accompanies image file 'pattern_fig_v2.pdf' (pdf, eps, ps)
%%
%% To include the image in your LaTeX document, write
%%   \input{<filename>.pdf_tex}
%%  instead of
%%   \includegraphics{<filename>.pdf}
%% To scale the image, write
%%   \def\svgwidth{<desired width>}
%%   \input{<filename>.pdf_tex}
%%  instead of
%%   \includegraphics[width=<desired width>]{<filename>.pdf}
%%
%% Images with a different path to the parent latex file can
%% be accessed with the `import' package (which may need to be
%% installed) using
%%   \usepackage{import}
%% in the preamble, and then including the image with
%%   \import{<path to file>}{<filename>.pdf_tex}
%% Alternatively, one can specify
%%   \graphicspath{{<path to file>/}}
%% 
%% For more information, please see info/svg-inkscape on CTAN:
%%   http://tug.ctan.org/tex-archive/info/svg-inkscape
%%
\begingroup%
  \makeatletter%
  \providecommand\color[2][]{%
    \errmessage{(Inkscape) Color is used for the text in Inkscape, but the package 'color.sty' is not loaded}%
    \renewcommand\color[2][]{}%
  }%
  \providecommand\transparent[1]{%
    \errmessage{(Inkscape) Transparency is used (non-zero) for the text in Inkscape, but the package 'transparent.sty' is not loaded}%
    \renewcommand\transparent[1]{}%
  }%
  \providecommand\rotatebox[2]{#2}%
  \newcommand*\fsize{\dimexpr\f@size pt\relax}%
  \newcommand*\lineheight[1]{\fontsize{\fsize}{#1\fsize}\selectfont}%
  \ifx\svgwidth\undefined%
    \setlength{\unitlength}{1134.67201996bp}%
    \ifx\svgscale\undefined%
      \relax%
    \else%
      \setlength{\unitlength}{\unitlength * \real{\svgscale}}%
    \fi%
  \else%
    \setlength{\unitlength}{\svgwidth}%
  \fi%
  \global\let\svgwidth\undefined%
  \global\let\svgscale\undefined%
  \makeatother%
  \begin{picture}(1,0.85159256)%
    \lineheight{1}%
    \setlength\tabcolsep{0pt}%
    \put(0,0){\includegraphics[width=\unitlength,page=1]{figs_pattern_fig.pdf}}%
    \put(0.07998018,0.37554417){\color[rgb]{0.94901961,0.39215686,0.09803922}\makebox(0,0)[lt]{\lineheight{1.25}\smash{\begin{tabular}[t]{l}g\textsubscript{1}\end{tabular}}}}%
    \put(0.99006636,0.41690599){\color[rgb]{0.94901961,0.39215686,0.09803922}\makebox(0,0)[lt]{\lineheight{1.25}\smash{\begin{tabular}[t]{l}s\textsubscript{1}\end{tabular}}}}%
    \put(0.06765967,0.48727159){\color[rgb]{0.96470588,0.68235294,0.17647059}\makebox(0,0)[lt]{\lineheight{1.25}\smash{\begin{tabular}[t]{l}g\textsubscript{2}\end{tabular}}}}%
    \put(0.04783015,0.57353){\color[rgb]{0.5254902,0.73333333,0.84705882}\makebox(0,0)[lt]{\lineheight{1.25}\smash{\begin{tabular}[t]{l}g\textsubscript{3}\end{tabular}}}}%
    \put(0.02713351,0.70103116){\color[rgb]{0.2,0.39607843,0.54117647}\makebox(0,0)[lt]{\lineheight{1.25}\smash{\begin{tabular}[t]{l}g\textsubscript{4}\end{tabular}}}}%
    \put(0.02103513,0.81578061){\color[rgb]{0.18431373,0.28235294,0.34509804}\makebox(0,0)[lt]{\lineheight{1.25}\smash{\begin{tabular}[t]{l}g\textsubscript{5}\end{tabular}}}}%
    \put(0.99006636,0.45316311){\color[rgb]{0.96470588,0.68235294,0.17647059}\makebox(0,0)[lt]{\lineheight{1.25}\smash{\begin{tabular}[t]{l}s\textsubscript{2}\end{tabular}}}}%
    \put(0.99006636,0.52289691){\color[rgb]{0.5254902,0.73333333,0.84705882}\makebox(0,0)[lt]{\lineheight{1.25}\smash{\begin{tabular}[t]{l}s\textsubscript{3}\end{tabular}}}}%
    \put(0.99006636,0.6403868){\color[rgb]{0.2,0.39607843,0.54117647}\makebox(0,0)[lt]{\lineheight{1.25}\smash{\begin{tabular}[t]{l}s\textsubscript{4}\end{tabular}}}}%
    \put(0.9900663,0.79819672){\color[rgb]{0.18431373,0.28235294,0.34509804}\makebox(0,0)[lt]{\lineheight{1.25}\smash{\begin{tabular}[t]{l}s\textsubscript{5}\end{tabular}}}}%
    \put(0,0){\includegraphics[width=\unitlength,page=2]{figs_pattern_fig.pdf}}%
  \end{picture}%
\endgroup%

%% file: main.bbl
\begin{thebibliography}{51}
\providecommand{\natexlab}[1]{#1}
\providecommand{\url}[1]{\texttt{#1}}
\expandafter\ifx\csname urlstyle\endcsname\relax
  \providecommand{\doi}[1]{doi: #1}\else
  \providecommand{\doi}{doi: \begingroup \urlstyle{rm}\Url}\fi

\bibitem[Bashkirova et~al.(2022)Bashkirova, Abdelfattah, Zhu, Akl, Alladkani,
  Hu, Ablavsky, Calli, Bargal, and Saenko]{Bashkirova:etal:CVPR22}
Dina Bashkirova, Mohamed Abdelfattah, Ziliang Zhu, James Akl, Fadi Alladkani,
  Ping Hu, Vitaly Ablavsky, Berk Calli, Sarah~Adel Bargal, and Kate Saenko.
\newblock Zerowaste dataset: Towards deformable object segmentation in
  cluttered scenes.
\newblock In \emph{CVPR}, pages 21147--21157, 2022.

\bibitem[Bolya et~al.(2022)Bolya, Fu, Dai, Zhang, Feichtenhofer, and
  Hoffman]{Bolya:etal:arXiv22}
Daniel Bolya, Cheng-Yang Fu, Xiaoliang Dai, Peizhao Zhang, Christoph
  Feichtenhofer, and Judy Hoffman.
\newblock Token merging: Your vit but faster.
\newblock \emph{arXiv preprint arXiv:2210.09461}, 2022.

\bibitem[Carrasco et~al.(1995)Carrasco, Evert, Chang, and
  Katz]{Carrasco:etal:PP95}
Marisa Carrasco, Denise~L Evert, Irene Chang, and Svetlana~M Katz.
\newblock The eccentricity effect: Target eccentricity affects performance on
  conjunction searches.
\newblock \emph{Perception \& psychophysics}, 57:\penalty0 1241--1261, 1995.

\bibitem[Chen et~al.(2024)Chen, Chu, Ren, Zhao, and Huang]{Chen:etal:CVPR24}
Honghao Chen, Xiangxiang Chu, Yongjian Ren, Xin Zhao, and Kaiqi Huang.
\newblock Pelk: Parameter-efficient large kernel convnets with peripheral
  convolution.
\newblock In \emph{CVPR}, pages 5557--5567, 2024.

\bibitem[Chen et~al.(2023)Chen, Shao, Xu, Lin, Zhang, Chao, Ji, Qiao, and
  Luo]{Chen:etal:ICCV23}
Mengzhao Chen, Wenqi Shao, Peng Xu, Mingbao Lin, Kaipeng Zhang, Fei Chao,
  Rongrong Ji, Yu Qiao, and Ping Luo.
\newblock Diffrate: Differentiable compression rate for efficient vision
  transformers.
\newblock In \emph{ICCV}, pages 17164--17174, 2023.

\bibitem[Cordts et~al.(2016)Cordts, Omran, Ramos, Rehfeld, Enzweiler, Benenson,
  Franke, Roth, and Schiele]{Cordts:etal:CVPR16}
Marius Cordts, Mohamed Omran, Sebastian Ramos, Timo Rehfeld, Markus Enzweiler,
  Rodrigo Benenson, Uwe Franke, Stefan Roth, and Bernt Schiele.
\newblock The cityscapes dataset for semantic urban scene understanding.
\newblock In \emph{CVPR}, 2016.

\bibitem[Damen et~al.(2022)Damen, Doughty, Farinella, Furnari, Kazakos, Ma,
  Moltisanti, Munro, Perrett, Price, et~al.]{Damen:etal:IJCV22}
Dima Damen, Hazel Doughty, Giovanni~Maria Farinella, Antonino Furnari,
  Evangelos Kazakos, Jian Ma, Davide Moltisanti, Jonathan Munro, Toby Perrett,
  Will Price, et~al.
\newblock Rescaling egocentric vision: Collection, pipeline and challenges for
  epic-kitchens-100.
\newblock \emph{International Journal of Computer Vision}, pages 1--23, 2022.

\bibitem[Darcet et~al.(2023)Darcet, Oquab, Mairal, and
  Bojanowski]{Darcet:etal:arXiv23}
Timoth{\'e}e Darcet, Maxime Oquab, Julien Mairal, and Piotr Bojanowski.
\newblock Vision transformers need registers.
\newblock \emph{arXiv preprint arXiv:2309.16588}, 2023.

\bibitem[Darkhalil et~al.(2022)Darkhalil, Shan, Zhu, Ma, Kar, Higgins, Fidler,
  Fouhey, and Damen]{Darkhalil:etal:NeurIPS22}
Ahmad Darkhalil, Dandan Shan, Bin Zhu, Jian Ma, Amlan Kar, Richard Higgins,
  Sanja Fidler, David Fouhey, and Dima Damen.
\newblock Epic-kitchens visor benchmark: Video segmentations and object
  relations.
\newblock In \emph{NeurIPS}, 2022.

\bibitem[Deng et~al.(2009)Deng, Dong, Socher, Li, Li, and
  Fei-Fei]{Deng:etal:CVPR09}
Jia Deng, Wei Dong, Richard Socher, Li-Jia Li, Kai Li, and Li Fei-Fei.
\newblock Imagenet: A large-scale hierarchical image database.
\newblock In \emph{CVPR}, pages 248--255. Ieee, 2009.

\bibitem[Dosovitskiy et~al.(2020)Dosovitskiy, Beyer, Kolesnikov, Weissenborn,
  Zhai, Unterthiner, Dehghani, Minderer, Heigold, Gelly,
  et~al.]{Dosovitskiy:etal:ICLR20}
Alexey Dosovitskiy, Lucas Beyer, Alexander Kolesnikov, Dirk Weissenborn,
  Xiaohua Zhai, Thomas Unterthiner, Mostafa Dehghani, Matthias Minderer, Georg
  Heigold, Sylvain Gelly, et~al.
\newblock An image is worth 16x16 words: Transformers for image recognition at
  scale.
\newblock \emph{arXiv preprint arXiv:2010.11929}, 2020.

\bibitem[Engel et~al.(2023)Engel, Somasundaram, Goesele, Sun, Gamino, Turner,
  Talattof, Yuan, Souti, Meredith, et~al.]{Engel:etal:arXiv23a}
Jakob Engel, Kiran Somasundaram, Michael Goesele, Albert Sun, Alexander Gamino,
  Andrew Turner, Arjang Talattof, Arnie Yuan, Bilal Souti, Brighid Meredith,
  et~al.
\newblock Project aria: A new tool for egocentric multi-modal ai research.
\newblock \emph{arXiv preprint arXiv:2308.13561}, 2023.

\bibitem[Fayyaz et~al.(2022)Fayyaz, Koohpayegani, Jafari, Sengupta, Joze,
  Sommerlade, Pirsiavash, and Gall]{Fayyaz:etal:ECCV22}
Mohsen Fayyaz, Soroush~Abbasi Koohpayegani, Farnoush~Rezaei Jafari, Sunando
  Sengupta, Hamid Reza~Vaezi Joze, Eric Sommerlade, Hamed Pirsiavash, and
  J{\"u}rgen Gall.
\newblock Adaptive token sampling for efficient vision transformers.
\newblock In \emph{ECCV}, pages 396--414. Springer, 2022.

\bibitem[Fortin et~al.(2022)Fortin, Gamache, Grondin, Pomerleau, and
  Gigu{\`e}re]{Fortin:etal:IROS22}
Jean-Michel Fortin, Olivier Gamache, Vincent Grondin, Fran{\c{c}}ois Pomerleau,
  and Philippe Gigu{\`e}re.
\newblock Instance segmentation for autonomous log grasping in forestry
  operations.
\newblock In \emph{2022 IEEE/RSJ International Conference on Intelligent Robots
  and Systems (IROS)}, pages 6064--6071. IEEE, 2022.

\bibitem[He et~al.(2016)He, Zhang, Ren, and Sun]{He:etal:ICCV16}
Kaiming He, Xiangyu Zhang, Shaoqing Ren, and Jian Sun.
\newblock Deep residual learning for image recognition.
\newblock In \emph{ICCV}, pages 770--778, 2016.

\bibitem[He et~al.(2022)He, Chen, Xie, Li, Doll{\'a}r, and
  Girshick]{He:etal:CVPR22}
Kaiming He, Xinlei Chen, Saining Xie, Yanghao Li, Piotr Doll{\'a}r, and Ross
  Girshick.
\newblock Masked autoencoders are scalable vision learners.
\newblock In \emph{CVPR}, pages 16000--16009, 2022.

\bibitem[Horton et~al.(2023)Horton, Mehta, Farhadi, and
  Rastegari]{Horton:etal:arXiv23}
Maxwell Horton, Sachin Mehta, Ali Farhadi, and Mohammad Rastegari.
\newblock Bytes are all you need: Transformers operating directly on file
  bytes.
\newblock \emph{arXiv preprint arXiv:2306.00238}, 2023.

\bibitem[Jonnalagadda et~al.(2021)Jonnalagadda, Wang, Manjunath, and
  Eckstein]{Jonnalagadda:etal:arXiv21}
Aditya Jonnalagadda, William~Yang Wang, BS Manjunath, and Miguel~P Eckstein.
\newblock Foveater: Foveated transformer for image classification.
\newblock \emph{arXiv preprint arXiv:2105.14173}, 2021.

\bibitem[Kaplan et~al.(2020)Kaplan, McCandlish, Henighan, Brown, Chess, Child,
  Gray, Radford, Wu, and Amodei]{Kaplan:etal:arXiv20}
Jared Kaplan, Sam McCandlish, Tom Henighan, Tom~B Brown, Benjamin Chess, Rewon
  Child, Scott Gray, Alec Radford, Jeffrey Wu, and Dario Amodei.
\newblock Scaling laws for neural language models.
\newblock \emph{arXiv preprint arXiv:2001.08361}, 2020.

\bibitem[Kirillov et~al.(2023)Kirillov, Mintun, Ravi, Mao, Rolland, Gustafson,
  Xiao, Whitehead, Berg, Lo, et~al.]{Kirillov:etal:CVPR23}
Alexander Kirillov, Eric Mintun, Nikhila Ravi, Hanzi Mao, Chloe Rolland, Laura
  Gustafson, Tete Xiao, Spencer Whitehead, Alexander~C Berg, Wan-Yen Lo, et~al.
\newblock Segment anything.
\newblock In \emph{CVPR}, pages 4015--4026, 2023.

\bibitem[Kong et~al.(2022)Kong, Dong, Ma, Meng, Niu, Sun, Shen, Yuan, Ren,
  Tang, et~al.]{Kong:etal:ECCV22}
Zhenglun Kong, Peiyan Dong, Xiaolong Ma, Xin Meng, Wei Niu, Mengshu Sun, Xuan
  Shen, Geng Yuan, Bin Ren, Hao Tang, et~al.
\newblock Spvit: Enabling faster vision transformers via latency-aware soft
  token pruning.
\newblock In \emph{ECCV}, pages 620--640. Springer, 2022.

\bibitem[Konrad et~al.(2024)Konrad, Padmanaban, Buckmaster, Boyle, and
  Wetzstein]{Konrad:etal:arXiv24}
Robert Konrad, Nitish Padmanaban, J~Gabriel Buckmaster, Kevin~C Boyle, and
  Gordon Wetzstein.
\newblock Gazegpt: Augmenting human capabilities using gaze-contingent
  contextual ai for smart eyewear.
\newblock \emph{arXiv preprint arXiv:2401.17217}, 2024.

\bibitem[Krizhevsky et~al.(2012)Krizhevsky, Sutskever, and
  Hinton]{Krizhevsky:etal:NeurIPS12}
Alex Krizhevsky, Ilya Sutskever, and Geoffrey~E Hinton.
\newblock Imagenet classification with deep convolutional neural networks.
\newblock \emph{NeurIPS}, 25, 2012.

\bibitem[Lin et~al.(2014)Lin, Maire, Belongie, Hays, Perona, Ramanan,
  Doll{\'a}r, and Zitnick]{Lin:etal:ECCV14}
Tsung-Yi Lin, Michael Maire, Serge Belongie, James Hays, Pietro Perona, Deva
  Ramanan, Piotr Doll{\'a}r, and C~Lawrence Zitnick.
\newblock Microsoft coco: Common objects in context.
\newblock In \emph{ECCV}, pages 740--755. Springer, 2014.

\bibitem[Liu et~al.(2023)Liu, Peng, Zheng, Yang, Hu, and Yuan]{Liu:etal:CVPR23}
Xinyu Liu, Houwen Peng, Ningxin Zheng, Yuqing Yang, Han Hu, and Yixuan Yuan.
\newblock Efficientvit: Memory efficient vision transformer with cascaded group
  attention.
\newblock In \emph{CVPR}, pages 14420--14430, 2023.

\bibitem[Ma et~al.(2024)Ma, Ye, Hong, Guzov, Jiang, Postyeni, Pesqueira,
  Gamino, Baiyya, Kim, et~al.]{Ma:etal:arXiv24}
Lingni Ma, Yuting Ye, Fangzhou Hong, Vladimir Guzov, Yifeng Jiang, Rowan
  Postyeni, Luis Pesqueira, Alexander Gamino, Vijay Baiyya, Hyo~Jin Kim, et~al.
\newblock Nymeria: A massive collection of multimodal egocentric daily motion
  in the wild.
\newblock \emph{arXiv preprint arXiv:2406.09905}, 2024.

\bibitem[Marin et~al.(2021)Marin, Chang, Ranjan, Prabhu, Rastegari, and
  Tuzel]{Marin:etal:arXiv21}
Dmitrii Marin, Jen-Hao~Rick Chang, Anurag Ranjan, Anish Prabhu, Mohammad
  Rastegari, and Oncel Tuzel.
\newblock Token pooling in vision transformers.
\newblock \emph{arXiv preprint arXiv:2110.03860}, 2021.

\bibitem[Meng et~al.(2022)Meng, Li, Chen, Lan, Wu, Jiang, and
  Lim]{Meng:etal:CVPR22}
Lingchen Meng, Hengduo Li, Bor-Chun Chen, Shiyi Lan, Zuxuan Wu, Yu-Gang Jiang,
  and Ser-Nam Lim.
\newblock Adavit: Adaptive vision transformers for efficient image recognition.
\newblock In \emph{CVPR}, pages 12309--12318, 2022.

\bibitem[Min et~al.(2022)Min, Zhao, Luo, and Cho]{Min:etal:NeurIPS22}
Juhong Min, Yucheng Zhao, Chong Luo, and Minsu Cho.
\newblock Peripheral vision transformer.
\newblock \emph{NeurIPS}, 35:\penalty0 32097--32111, 2022.

\bibitem[Minervini et~al.(2016)Minervini, Fischbach, Scharr, and
  Tsaftaris]{Minervini:etal:PRL16}
Massimo Minervini, Andreas Fischbach, Hanno Scharr, and Sotirios~A Tsaftaris.
\newblock Finely-grained annotated datasets for image-based plant phenotyping.
\newblock \emph{Pattern recognition letters}, 81:\penalty0 80--89, 2016.

\bibitem[Park and Johnson(2023)]{Park:Johnson:CVPR23}
Jeongsoo Park and Justin Johnson.
\newblock Rgb no more: Minimally-decoded jpeg vision transformers.
\newblock In \emph{CVPR}, pages 22334--22346, 2023.

\bibitem[Rao et~al.(2021)Rao, Zhao, Liu, Lu, Zhou, and
  Hsieh]{Rao:etal:NeurIPS21}
Yongming Rao, Wenliang Zhao, Benlin Liu, Jiwen Lu, Jie Zhou, and Cho-Jui Hsieh.
\newblock Dynamicvit: Efficient vision transformers with dynamic token
  sparsification.
\newblock \emph{NeurIPS}, 34:\penalty0 13937--13949, 2021.

\bibitem[Ravi et~al.(2024)Ravi, Gabeur, Hu, Hu, Ryali, Ma, Khedr, R{\"a}dle,
  Rolland, Gustafson, et~al.]{Ravi:etal:arXiv24}
Nikhila Ravi, Valentin Gabeur, Yuan-Ting Hu, Ronghang Hu, Chaitanya Ryali,
  Tengyu Ma, Haitham Khedr, Roman R{\"a}dle, Chloe Rolland, Laura Gustafson,
  et~al.
\newblock Sam 2: Segment anything in images and videos.
\newblock \emph{arXiv preprint arXiv:2408.00714}, 2024.

\bibitem[Renggli et~al.(2022)Renggli, Pinto, Houlsby, Mustafa, Puigcerver, and
  Riquelme]{Renggli:etal:arXiv22}
Cedric Renggli, Andr{\'e}~Susano Pinto, Neil Houlsby, Basil Mustafa, Joan
  Puigcerver, and Carlos Riquelme.
\newblock Learning to merge tokens in vision transformers.
\newblock \emph{arXiv preprint arXiv:2202.12015}, 2022.

\bibitem[Song et~al.(2022)Song, Xu, He, Jiang, Jing, and
  Liang]{Song:etal:arXiv22}
Zhuoran Song, Yihong Xu, Zhezhi He, Li Jiang, Naifeng Jing, and Xiaoyao Liang.
\newblock Cp-vit: Cascade vision transformer pruning via progressive sparsity
  prediction.
\newblock \emph{arXiv preprint arXiv:2203.04570}, 2022.

\bibitem[Sun et~al.(2024)Sun, Liu, Shen, Zhu, and Hu]{Sun:etal:arXiv254}
Xiaorui Sun, Jun Liu, Heng~Tao Shen, Xiaofeng Zhu, and Ping Hu.
\newblock On efficient variants of segment anything model: A survey.
\newblock \emph{arXiv preprint arXiv:2410.04960}, 2024.

\bibitem[Trotter et~al.(2020)Trotter, Atkinson, Sharpe, Richardson, McGough,
  Wright, Burville, and Berggren]{Trotter:etal:arXiv20}
Cameron Trotter, Georgia Atkinson, Matt Sharpe, Kirsten Richardson, A~Stephen
  McGough, Nick Wright, Ben Burville, and Per Berggren.
\newblock Ndd20: A large-scale few-shot dolphin dataset for coarse and
  fine-grained categorisation.
\newblock \emph{arXiv preprint arXiv:2005.13359}, 2020.

\bibitem[Van Den~Oord et~al.(2017)Van Den~Oord, Vinyals,
  et~al.]{VanDenOord:etal:NeurIPS17}
Aaron Van Den~Oord, Oriol Vinyals, et~al.
\newblock Neural discrete representation learning.
\newblock \emph{NeurIPS}, 30, 2017.

\bibitem[Varadarajan et~al.(2023)Varadarajan, Soran, Iandola, Xiang, Xiong, Wu,
  Zhu, Krishnamoorthi, and Chandra]{Varadarajan:etal:arXiv23}
Balakrishnan Varadarajan, Bilge Soran, Forrest Iandola, Xiaoyu Xiang, Yunyang
  Xiong, Lemeng Wu, Chenchen Zhu, Raghuraman Krishnamoorthi, and Vikas Chandra.
\newblock Squeezesam: User friendly mobile interactive segmentation.
\newblock \emph{arXiv preprint arXiv:2312.06736}, 2023.

\bibitem[Xiong et~al.(2024)Xiong, Varadarajan, Wu, Xiang, Xiao, Zhu, Dai, Wang,
  Sun, Iandola, et~al.]{Xiong:etal:CVPR24}
Yunyang Xiong, Bala Varadarajan, Lemeng Wu, Xiaoyu Xiang, Fanyi Xiao, Chenchen
  Zhu, Xiaoliang Dai, Dilin Wang, Fei Sun, Forrest Iandola, et~al.
\newblock Efficientsam: Leveraged masked image pretraining for efficient
  segment anything.
\newblock In \emph{CVPR}, pages 16111--16121, 2024.

\bibitem[Yan et~al.(2024)Yan, Zaharia, Mnih, Abbeel, Faust, and
  Liu]{Yan:etal:arXiv24}
Wilson Yan, Matei Zaharia, Volodymyr Mnih, Pieter Abbeel, Aleksandra Faust, and
  Hao Liu.
\newblock Elastictok: Adaptive tokenization for image and video.
\newblock \emph{arXiv preprint arXiv:2410.08368}, 2024.

\bibitem[Yin et~al.(2022)Yin, Vahdat, Alvarez, Mallya, Kautz, and
  Molchanov]{Yin:etal:CVPR22}
Hongxu Yin, Arash Vahdat, Jose~M Alvarez, Arun Mallya, Jan Kautz, and Pavlo
  Molchanov.
\newblock A-vit: Adaptive tokens for efficient vision transformer.
\newblock In \emph{CVPR}, pages 10809--10818, 2022.

\bibitem[Yogamani et~al.(2019)Yogamani, Hughes, Horgan, Sistu, Varley, O'Dea,
  Uric{\'a}r, Milz, Simon, Amende, et~al.]{Yogamani:etal:ICCV19}
Senthil Yogamani, Ciar{\'a}n Hughes, Jonathan Horgan, Ganesh Sistu, Padraig
  Varley, Derek O'Dea, Michal Uric{\'a}r, Stefan Milz, Martin Simon, Karl
  Amende, et~al.
\newblock Woodscape: A multi-task, multi-camera fisheye dataset for autonomous
  driving.
\newblock In \emph{ICCV}, pages 9308--9318, 2019.

\bibitem[Yu et~al.(2023)Yu, Lezama, Gundavarapu, Versari, Sohn, Minnen, Cheng,
  Birodkar, Gupta, Gu, et~al.]{Yu:etal:arXiv23}
Lijun Yu, Jos{\'e} Lezama, Nitesh~B Gundavarapu, Luca Versari, Kihyuk Sohn,
  David Minnen, Yong Cheng, Vighnesh Birodkar, Agrim Gupta, Xiuye Gu, et~al.
\newblock Language model beats diffusion--tokenizer is key to visual
  generation.
\newblock \emph{arXiv preprint arXiv:2310.05737}, 2023.

\bibitem[Yu et~al.(2024)Yu, Weber, Deng, Shen, Cremers, and
  Chen]{Yu:etal:arXiv24}
Qihang Yu, Mark Weber, Xueqing Deng, Xiaohui Shen, Daniel Cremers, and
  Liang-Chieh Chen.
\newblock An image is worth 32 tokens for reconstruction and generation.
\newblock \emph{arXiv preprint arXiv:2406.07550}, 2024.

\bibitem[Zhang et~al.(2023)Zhang, Han, Qiao, Kim, Bae, Lee, and
  Hong]{Zhang:etal:arXiv23}
Chaoning Zhang, Dongshen Han, Yu Qiao, Jung~Uk Kim, Sung-Ho Bae, Seungkyu Lee,
  and Choong~Seon Hong.
\newblock Faster segment anything: Towards lightweight sam for mobile
  applications.
\newblock \emph{arXiv preprint arXiv:2306.14289}, 2023.

\bibitem[Zhang et~al.(2022)Zhang, Zhou, Stent, and Shi]{Zhang:etal:ECCV22}
Lingzhi Zhang, Shenghao Zhou, Simon Stent, and Jianbo Shi.
\newblock Fine-grained egocentric hand-object segmentation: Dataset, model, and
  applications.
\newblock In \emph{ECCV}, pages 127--145. Springer, 2022.

\bibitem[Zhang et~al.(2024)Zhang, Cai, and Han]{Zhang:etal:CVPR24}
Zhuoyang Zhang, Han Cai, and Song Han.
\newblock Efficientvit-sam: Accelerated segment anything model without
  performance loss.
\newblock In \emph{CVPR}, pages 7859--7863, 2024.

\bibitem[Zhao et~al.(2023)Zhao, Ding, An, Du, Yu, Li, Tang, and
  Wang]{Zhao:etal:arXiv23}
Xu Zhao, Wenchao Ding, Yongqi An, Yinglong Du, Tao Yu, Min Li, Ming Tang, and
  Jinqiao Wang.
\newblock Fast segment anything.
\newblock \emph{arXiv preprint arXiv:2306.12156}, 2023.

\bibitem[Zhou et~al.(2019)Zhou, Zhao, Puig, Xiao, Fidler, Barriuso, and
  Torralba]{Zhou:etal:IJCV19}
Bolei Zhou, Hang Zhao, Xavier Puig, Tete Xiao, Sanja Fidler, Adela Barriuso,
  and Antonio Torralba.
\newblock Semantic understanding of scenes through the ade20k dataset.
\newblock \emph{IJCV}, 127:\penalty0 302--321, 2019.

\bibitem[Zhou et~al.(2023)Zhou, Li, Loy, and Dai]{Zhou:etal:arXiv23}
Chong Zhou, Xiangtai Li, Chen~Change Loy, and Bo Dai.
\newblock Edgesam: Prompt-in-the-loop distillation for on-device deployment of
  sam.
\newblock \emph{arXiv preprint arXiv:2312.06660}, 2023.

\end{thebibliography}
